\documentclass[10pt,twocolumn,letterpaper]{article}
\usepackage[accsupp]{axessibility}
\usepackage{cvpr}      

\def\cvprPaperID{11304} 
\def\confName{CVPR}
\def\confYear{2023}

\usepackage[utf8]{inputenc} 
\usepackage[T1]{fontenc}    
\usepackage{hyperref}       
\usepackage{url}            
\usepackage{booktabs}       
\usepackage{amsfonts}       
\usepackage{nicefrac}       
\usepackage{microtype}      
\usepackage{xcolor}         

\usepackage{pgfplots}
\usepackage{comment}
\usepackage{amsmath}
\usepackage{hyperref}
\usepackage{gensymb}
\usepackage{booktabs}
\usepackage{multirow}
\usepackage{wrapfig}
\usepackage{tabu}
\usepackage{soul}
\usepackage{floatrow}
\newfloatcommand{capbtabbox}{table}[][\FBwidth]

\newdimen\figrasterwd
\figrasterwd\columnwidth
\newcommand{\revised}[1]{{\color{black}#1}}

\title{Data-efficient Large Scale Place Recognition with Graded Similarity Supervision}

\author{  Mar\'ia Leyva-Vallina\\
  University of Groningen\\
{\tt\small m.leyva.vallina@rug.nl}
\and
 Nicola Strisciuglio \\
University of Twente \\
{\tt\small n.strisciuglio@utwente.nl}
\and
   Nicolai Petkov \\
   University of Groningen \\
{\tt\small n.petkov@rug.nl}
}

\begin{document}

\maketitle

\begin{abstract}
Visual place recognition (VPR) is a fundamental task of computer vision for visual localization. Existing methods are trained using image pairs that either depict the same place or not. Such a binary indication does not consider continuous relations of similarity between images of the same place taken from different positions, determined by the continuous nature of camera pose. The binary similarity induces a noisy supervision signal into the training of VPR methods, which stall in local minima and require expensive hard mining algorithms to guarantee convergence. Motivated by the fact that two images of the same place only partially share visual cues due to camera pose differences, we deploy an automatic re-annotation strategy to re-label VPR datasets. We compute graded similarity labels for image pairs based on available localization metadata. Furthermore, we propose a new Generalized Contrastive Loss (GCL) that uses graded similarity labels for training contrastive networks. We demonstrate that the use of the new labels and GCL allow to dispense from hard-pair mining, and to train image descriptors that perform better in VPR by nearest neighbor search, obtaining superior or comparable results than methods that require expensive hard-pair mining and re-ranking techniques.

\end{abstract}

\section{Introduction}

Visual place recognition (VPR) is  an important task of computer vision, and a fundamental building block of navigation systems for autonomous vehicles~\cite{Lowry2016,Zaffar2021}. It is approached either with structure-based methods, namely Structure-from-Motion~\cite{colmap} and SLAM~\cite{Milford2012}, or with image retrieval~\cite{Arandjelovic2017,hausler2021patch,radenovic2018fine,radenovic2018revisiting,leyvavallina2019caip,weyand2020google}. The former focus on precise relative camera pose estimation~\cite{Sattler2012,Sattler2019}. 
The latter aim at learning image descriptors for effective retrieval of similar images to a given query in a nearest search approach~\cite{benchmarkingir3DV2020}. The goal of descriptor learning is to ensure images of the same place to be projected onto close-by points in a latent space, and images of different places to be projected onto distant points~\cite{milford15metric,Liu2019metric,Chen2022surveyretrieval}. Contrastive~\cite{radenovic2018fine,leyvavallina2019access} and triplet~\cite{Arandjelovic2017, Lopez-Antequera2017,liu2019stochastic,Peng2021} loss were used for this goal and resulted in state-of-the-art performance on several VPR benchmarks. 

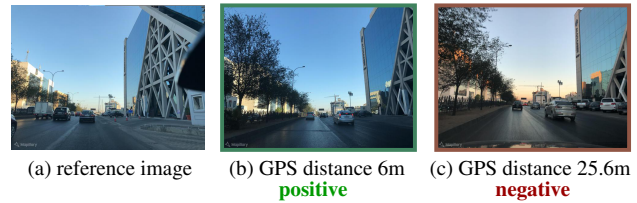
\begin{figure}[t!]
    \centering
    {\fontsize{10pt}{12pt}
    \def\svgwidth{\textwidth}
\begingroup%
  \makeatletter%
  \providecommand\color[2][]{%
    \errmessage{(Inkscape) Color is used for the text in Inkscape, but the package 'color.sty' is not loaded}%
    \renewcommand\color[2][]{}%
  }%
  \providecommand\transparent[1]{%
    \errmessage{(Inkscape) Transparency is used (non-zero) for the text in Inkscape, but the package 'transparent.sty' is not loaded}%
    \renewcommand\transparent[1]{}%
  }%
  \providecommand\rotatebox[2]{#2}%
  \newcommand*\fsize{\dimexpr\f@size pt\relax}%
  \newcommand*\lineheight[1]{\fontsize{\fsize}{#1\fsize}\selectfont}%
  \ifx\svgwidth\undefined%
    \setlength{\unitlength}{319.23899253bp}%
    \ifx\svgscale\undefined%
      \relax%
    \else%
      \setlength{\unitlength}{\unitlength * \real{\svgscale}}%
    \fi%
  \else%
    \setlength{\unitlength}{\svgwidth}%
  \fi%
  \global\let\svgwidth\undefined%
  \global\let\svgscale\undefined%
  \makeatother%
  \footnotesize
  \begin{picture}(1,0.33080496)%
    \lineheight{1}%
    \setlength\tabcolsep{0pt}%
    \put(0,0){\includegraphics[width=\unitlength]{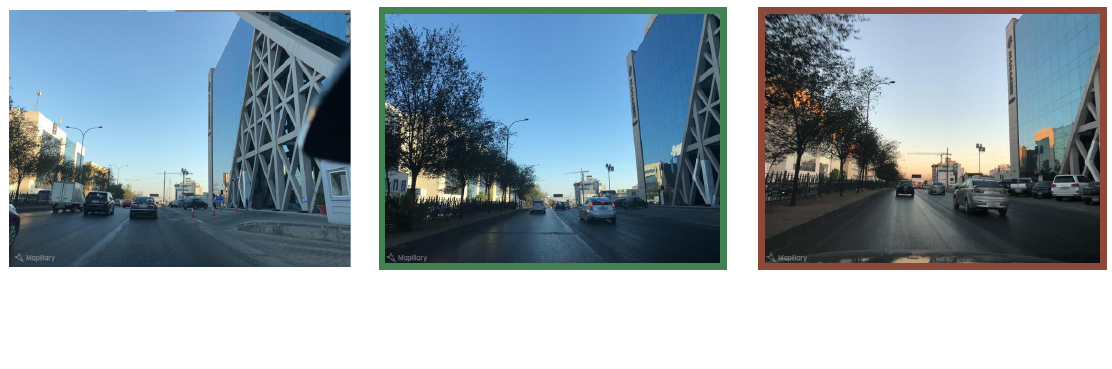}}%
    \put(0.034,0.051){\color[rgb]{0,0,0}\makebox(0,0)[lt]{\lineheight{1.25}\smash{\begin{tabular}[t]{l}(a) reference image\end{tabular}}}}%
    \put(0.355,0.051){\color[rgb]{0,0,0}\makebox(0,0)[lt]{\lineheight{1.25}\smash{\begin{tabular}[t]{l}(b) GPS distance 6m\end{tabular}}}}%
    \put(0.434,0.017){\color[rgb]{0,0.6,0}\makebox(0,0)[lt]{\lineheight{1.25}\smash{\begin{tabular}[t]{l}\textbf{positive}\end{tabular}}}}%
    \put(0.675,0.051){\color[rgb]{0,0,0}\makebox(0,0)[lt]{\lineheight{1.25}\smash{\begin{tabular}[t]{l}(c) GPS distance  25.6m\end{tabular}}}}%
    \put(0.78,0.017){\color[rgb]{0.6,0,0}\makebox(0,0)[lt]{\lineheight{1.25}\smash{\begin{tabular}[t]{l}\textbf{negative}\end{tabular}}}}%
  \end{picture}%
\endgroup%
}
      \vspace{-6mm}
    \caption{(a) A place in the city of Amman. (b) An image taken 6m away is labeled as positive (same place), while (c) an image taken 25.6m away is labeled as negative (not the same place) despite sharing a lot of visual cues.}
    \label{fig:similarity-example}
\end{figure}

VPR methods are normally trained using image pairs labelled to indicate they either depict the same place or not, in a binary fashion. In practice, images of a certain place can be taken from different positions, i.e. with a different camera pose, and thus share only a part of their visual cues (or surface in 3D). In existing datasets, two images are usually labeled to be of the same place (positive) if they are taken within a predefined range (usually $25m$) computed using e.g. GPS metadata. This creates ambiguous cases. For instance, Figure~\ref{fig:similarity-example} shows a reference image (a) of a place and two other pictures taken $6m$ (b) and $25.6m$ (c) away from its position. The images are respectively labeled as positive and negative match, although they share many visual cues (e.g. the building on the right).
Binary labels are thus noisy and interfere with the training of VPR networks, that usually stall in local minima. To address this, resource- and time-costly hard pair mining strategies are used to compose the training batches. For example, training NetVLAD~\cite{Arandjelovic2017} on the Mapillary Street Level Sequences (MSLS) dataset~\cite{msls} can take more than 20 days on an Nvidia v100 gpu due to the complexity of pair mining.
We instead build on the observation that two images depict the same place only to a certain degree of shared cues, namely a degree of similarity, and propose to embed this information in new continuous labels for existing datasets that can be used to reduce the effect of noise in the training of effective VPR methods. %


In this paper we exploit camera pose metadata or 3D information associated to image pairs as a proxy to estimate an approximate degree of similarity (hereinafter, graded similarity) between images of the same place, and use it to relabel popular VPR datasets. Graded similarity labels can be used to pick easy- and hard-pairs and compose training batches without complex pair-mining, thus speeding-up the training of VPR networks and enabling an efficient use of data.
Furthermore, we embed the graded similarity into a Generalized Contrastive Loss (GCL) function that we use to train a VPR pipeline. The intuition behind this choice is that the update of network weights should not be equal for all training pairs, but rather be influenced by their similarity. The representations of image pairs with larger graded similarity should be pushed together in the latent space more strongly than those of images with a lower graded similarity. The distance in the latent space is thus expected to be a better measure of ranking images according to their similarity, avoiding the use of expensive re-ranking to improve retrieval results. 
We validate the proposed approaches on several VPR benchmark datasets. 
To the best of our knowledge, this work is the first to use graded similarity for large-scale place recognition, and paying attention to data-efficient training. 


We summarize the contributions of this work as: 
\begin{itemize}
    \item new labels for VPR datasets indicating the graded similarity of image pairs. We computed the labels with automatic methods that use camera pose metadata included with the images or 3D surface information;
    \item a generalized contrastive loss (GCL) that exploits graded similarity of image pairs to learn effective descriptors for VPR;
    \item an efficient VPR pipeline trained without hard-pair mining, and that does not require re-ranking. Training our pipeline with a VGG-16 backbone converges $\sim100x$ faster than NetVLAD with the same backbone, achieving higher VPR results on several benchmarks. The efficiency of our scheme enables training larger backbones in a short time.
\end{itemize}





\section{Related works}
 
\noindent\textbf{Place recognition as image retrieval.} 
Visual place recognition is widely addressed as a metric learning problem, in which the descriptors of images of a place are learned to be close together in a latent space~\cite{milford15metric}. Existing methods optimize ranking loss functions, such as contrastive, triplet or average precision~\cite{radenovic2018fine,Gordo2017,revaud2019learning}. An extensive  benchmark of different approaches is in~\cite{Berton2022Bench}.
NetVLAD~\cite{Arandjelovic2017} is milestone of VPR and builds on a triplet network with an end-to-end trainable VLAD layer. It requires a computationally- and memory-expensive hard-pair mining to compose proper batches and guarantee convergence. 
SARE~\cite{liu2019stochastic} uses a NetVLAD backbone trained with a probabilistic attractive and repulsive mechanism, also making use of hard-pair mining. 
Hard-pair mining addresses issues of the training stalling in local minima due to noisy binary labels, and is used to compose the training batches so that hard pairs are selected for the training~\cite{Arandjelovic2017}. We instead use image metadata (e.g. camera pose as GPS and compass) to a-priori estimate the graded similarity of image pairs, and subsequently use it to balance hard- and easy-pairs in the training batches. This allows to train VPR models using the graded similarity of images and avoiding hard-pair mining.

Training with noisy binary labels produces image descriptors with drawbacks in nearest neighbor search retrieval, and re-ranking algorithms are necessary to post-process the retrieved results and increase VPR performance~\cite{delg,sarlin20superglue}. Patch-NetVLAD~\cite{hausler2021patch} builds on a NetVLAD backbone and performs multi-scale aggregation of NetVLAD descriptors to re-rank retrieval results. A transformer architecture named TransVPR was trained using a triplet loss function and hard-pair mining in~\cite{wang2022transvpr}. The retrieval step is combined with a costly re-ranking strategy to improve the retrieval results. We instead focus on using  more informative and robust image pair labels to avoid noisy training and obtain more effective image descriptors for nearest neighbor search, with no necessity of performing re-ranking.


\noindent\textbf{Image graded similarity. } 
Soft assignment to positive and negative classes of image pairs was investigated in~\cite{thoma2020soft}, where weighting of the assignment was based on the Euclidean distance between the GPS coordinates associated to the images. As the GPS distance induced label noise in the training process, hard-negative pair mining was still necessary to train VPR networks.
In~\cite{SFRS}, image region similarity was coupled with the GPS weak labels in a self-supervised framework to mine hard positive samples. In~\cite{berton22cosplace}, the authors formulated the VPR metric learning as a classification problem, splitting image training into classes based on similar GPS locations  to facilitate large-scale city-wide recognition.
Camera pose was used in~\cite{balntas2018relocnet} to estimate the camera frustum overlap and regress descriptors for camera (re-)localization in small-scale (indoor) environments. 
In~\cite{kim2021embedding}, a weighting scheme for the contrastive loss function is proposed as a function of the distance in the latent space, which requires an extra step of normalization of the distances to avoid a divergent training. In this work, we relabel VPR datasets using camera pose and field of view overlap, or ratio of shared 3D surface as proxies to estimate the graded similarity of training image pairs. We compute the new labels once, and use them to select the training batches and directly in the optimization of the networks to obtain effective descriptors for VPR in a data-efficient manner.

\noindent\textbf{Relation and difference with prior works. } We undertake a different direction than previous works, and propose a simplified way to learn image descriptors for retrieval-based VPR. We use contrastive architectures without hard-pair mining and exploit the graded similarity of image pairs to learn robust descriptors.
Instead of developing algorithmic solutions (e.g. hard-pair mining or re-ranking) to achieve better VPR results by increasing the complexity of the methods,  
we focus on data-efficiency and improve similarity labels to better exploit the training data. This allows to purposely keep the complexity of the architecture simpler (a convolutional backbone and a straightforward pooling strategy) than other methods. We apply prior knowledge and use metadata about the position and orientation of the cameras to estimate a more robust ground truth image similarity that enables to drop expensive hard-mining procedures and train (bigger) networks efficiently. We show that this approach leads to reduced training time and very robust descriptors that perform well in nearest neighbour search with no need of re-ranking.


\section{Generalized Contrastive Learning}
\label{sec:methodology}
\noindent\textbf{Preliminaries.} 
Contrastive approaches for metric learning in visual place recognition consider training a (convolutional) neural network $\hat{f}(x)$ so that the distance of the vector representation of similar (or dissimilar) images in a latent space is minimized (or maximized). In this work, we consider siamese networks optimized using a Contrastive Loss function~\cite{hadsell2006dimensionality}.

Let $x_i$ and $x_j$ be two input images, with $\hat{f}(x_i)$ and $\hat{f}(x_j)$ their descriptors. The distance of the descriptors in the latent space is the $L_2$-distance $d(x_i, x_j) = \left \| \hat{f}(x_i)-\hat{f}(x_j) \right \|_2$. The Contrastive Loss $\mathcal{L}_{CL}$ used to train the networks is defined as:
\begin{equation}
 \mathcal{L}_{CL}(x_i,x_j)=\begin{cases}
     \frac{1}{2}d(x_i, x_j)^2 ,& \text{if } y= 1\\
    \frac{1}{2}\max(\tau-d(x_i, x_j),0)^2,& \text{if } y=0
\end{cases}
\label{eq:contrastive}
\end{equation}
where $\tau$ is the margin, an hyper-parameter that defines a boundary between similar and dissimilar pairs. The ground truth label $y$ is binary: $1$ indicates a pair of similar images, and $0$ a not-similar pair of images.  In practice, however, a binary ground truth for similarity may cause the trained models to provide unreliable predictions.

\noindent\textbf{Generalized Contrastive Loss. }
We reformulate the Contrastive Loss, using a generalized definition of pair similarity as a continuous value $\psi_{i,j} \in \left[0, 1 \right]$. We define the Generalized Contrastive Loss function $ \mathcal{L}_{GCL}$ as:

\begin{align}
  \begin{split} 
 \mathcal{L}_{GCL}(x_i,x_j)= \psi_{i,j}\cdot \frac{1}{2}d(x_i, x_j)^2 + \\ (1-\psi_{i,j}) \cdot \frac{1}{2}\max(\tau-d(x_i, x_j),0)^2
\end{split}
\label{eq:generalized_contrastive_loss}
\end{align}

In contrast to Eq.~\ref{eq:contrastive}, here the similarity $\psi_{i,j}$ is a continuous value ranging from 0 (completely dissimilar) to 1 (identical). 
By minimising the Generalized Contrastive Loss, the distance of image pairs in the latent space is optimized proportionally to the corresponding degree of similarity. 

\noindent\textbf{Gradient of the GCL.}
In the training phase, the loss function is minimized by gradient descent optimization and the weights of the network are updated by backpropagation. In the case of the Constrastive Loss function, the gradient is:
\begin{equation}
     \nabla  \mathcal{L}_{CL}(x_i,x_j)=\begin{cases}
     d(x_i, x_j) ,& \text{if } y= 1\\
    \min(d(x_i, x_j)-\tau,0),& \text{if } y=0
\end{cases}
\label{eq:contrastive_loss_grad}
\end{equation}
The gradient is computed for all positive pairs, and corresponds to a direct minimization of their descriptor distance in the latent space. For negative pairs, the update of the network weights takes place only in the case the distance of the descriptors is within the margin $\tau$. If the latent vectors are already at a distance higher than $\tau$, no update is done. 

The Generalized Contrastive Loss, instead, explicitly accounts for graded similarity $\psi_{i,j}$ of input pairs $(x_i,\!x_j)$ to weight the learning steps, and this reflects into the gradient:

\begin{equation}
     \nabla  \mathcal{L}_{GCL}(x_i,\!x_j)\!=\!\begin{cases}d(x_i,\!x_j)\!+\!\tau(\psi_{i,j}\!-\!1),& \text{if } d(x_i,\!x_j)\!<\!\tau\\
     d(x_i, x_j) \cdot \psi_{i,j},& \text{if } d(x_i,\!x_j)\!\geq\!\tau\\
     \end{cases}
     \label{eq:generalized_contrastive_loss_grad}
\end{equation} 

\noindent The gradient of $\mathcal{L}_{GCL}$ is modulated by the degree of similarity of the input image pairs, $\psi_{i,j}$. This results in an implicit regularization of learned latent space. In the supplementary material, we provide and compare plots of the latent space learned with the $\mathcal{L}_{CL}$ and $\mathcal{L}_{GCL}$ functions.  
At the extremes of the similarity range, for $\psi_{i,j}=0$ (completely dissimilar input images) and $\psi_{i,j}=1$ (same exact input images), the gradient is the same as in Eq.~\ref{eq:contrastive_loss_grad}.



\section{Experimental evaluation}
\subsection{Data}

\noindent\textbf{Mapillary Street Level Sequences. } The Mapillary Street Level Sequences (MSLS) dataset is designed for life-long large-scale visual place recognition. It contains about 1.6M images taken in $30$ cities across the world~\cite{msls}. Images are divided into a training (22 cities, 1.4M images), validation (2 cities, 30K images) and test (6 cities, 66k images) set. The dataset presents strong challenges related to images taken at different times of the day, in different seasons and with strong variations of camera viewpoint. The images are provided with GPS data in UTM format and compass angle. According to the original paper~\cite{msls}, two images are considered similar if they are taken by cameras located within $25m$ of distance, and with less than $40^{\circ}$ of viewpoint variation. We created (and will release) new ground truth labels for the training set of MSLS, with specification of the graded similarity of image pairs (see next Section for details).
We use the MSLS dataset to train our large-scale VPR models, which we test on the validation set, and also the private test set using the available evaluation server.

 

\noindent\textbf{OOD test data for generalization. }  We use out-of-distribution (OOD) test sets, to evaluate the generalization abilities of our models and compare them with existing methods. We thus use the test and validation sets of several other benchmark datasets, namely the Pittsburgh30k~\cite{Arandjelovic2017}, Tokyo24/7~\cite{Torii-PAMI2015}, RobotCar Seasons v2~\cite{Maddern2017,sattler2018benchmarking} and  Extended CMU Seasons~\cite{Badino2011,sattler2018benchmarking} datasets. In the supplementary material, we also report results on Pittsburgh250k and TokyoTM~\cite{Torii-PAMI2015}. 

\noindent\textbf{TB-Places and 7Scenes.} We carried out experiments also using the TB-Places~\cite{leyvavallina2019access} and 7Scenes~\cite{Shotton2013} datasets, which were recorded in small-scale environments. We train models on them and report the results in the supplementary material.
TB-Places was recorded in an outdoor garden over two years and contains challenges related to drastic viewpoint variations, as well as illumination changes, and scenes mostly filled with repetitive texture of green color. Each image has 6DOF pose metadata. The 7Scenes dataset is recorded in seven indoor environments. It contains 6DOF pose metadata for each image and  a 3D pointcloud of each scene.

The different format and type of metadata, namely 6DOF camera pose and 3D pointclouds of the scenes, are of interest to investigate different ways to estimate the ground truth graded similarity of image pairs. In the following, we present techniques to automatically re-label VPR datasets, when 6DOF pose or 3D pointclouds metadata are available.

\subsection{Graded similarity labels}

%


Images of the same place can be taken from different positions, i.e. with different camera pose, and share only part of the visual cues. On the basis of the amount of shared characteristics among images, we indeed tend to perceive images more or less similar~\cite{Desolneux2008}. 
Actual labeling of VPR datasets do not consider this and instead mark two images either similar (depicting the same place) or not (depicting different places). This does not take into account continuous relations between images, which are induced by the continuous nature of camera pose. 

We combine the concept of perceived visual similarity with the continuous nature of camera pose, and design a method to automatically relabel VPR dataset by annotating the graded similarity of image pairs\footnote{We release the labels at \url{https://github.com/marialeyvallina/generalized_contrastive_loss}.}. We approximate the similarity between two images via a proxy, namely measuring the overlap of the field of view of the cameras or 3D information associated to the images. 

\begin{figure}[t!]
  \centering
      \begin{subfigure}{.2\textwidth}
      \def\svgwidth{\textwidth}
\begingroup%
  \makeatletter%
  \providecommand\color[2][]{%
    \errmessage{(Inkscape) Color is used for the text in Inkscape, but the package 'color.sty' is not loaded}%
    \renewcommand\color[2][]{}%
  }%
  \providecommand\transparent[1]{%
    \errmessage{(Inkscape) Transparency is used (non-zero) for the text in Inkscape, but the package 'transparent.sty' is not loaded}%
    \renewcommand\transparent[1]{}%
  }%
  \providecommand\rotatebox[2]{#2}%
  \newcommand*\fsize{\dimexpr\f@size pt\relax}%
  \newcommand*\lineheight[1]{\fontsize{\fsize}{#1\fsize}\selectfont}%
  \ifx\svgwidth\undefined%
    \setlength{\unitlength}{171.00101852bp}%
    \ifx\svgscale\undefined%
      \relax%
    \else%
      \setlength{\unitlength}{\unitlength * \real{\svgscale}}%
    \fi%
  \else%
    \setlength{\unitlength}{\svgwidth}%
  \fi%
  \global\let\svgwidth\undefined%
  \global\let\svgscale\undefined%
  \makeatother%
  \begin{picture}(1,1.25386037)%
    \lineheight{1}%
    \setlength\tabcolsep{0pt}%
    \put(0,0){\includegraphics[width=\unitlength,page=1]{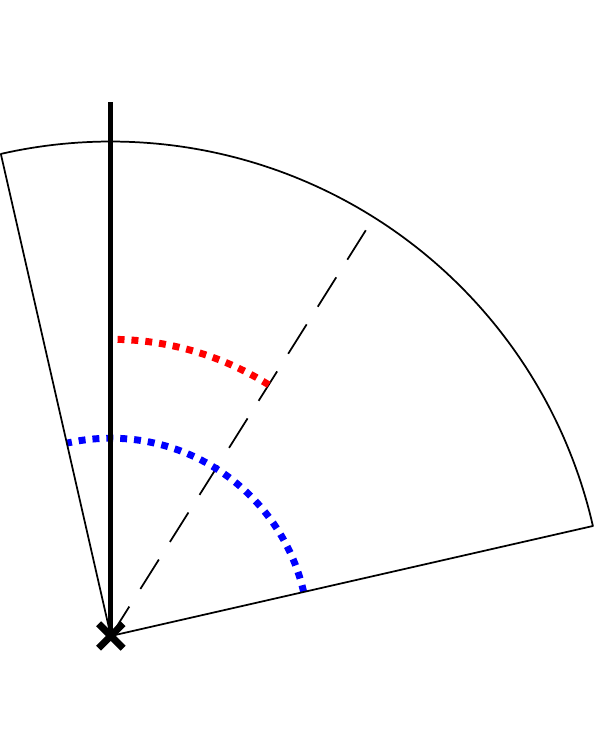}}%
    \put(0.3,0.51){\makebox(0,0)[lt]{\lineheight{1.25}\smash{\begin{tabular}[t]{l}\color[rgb]{0,0,1}{$\theta$}\end{tabular}}}}%
    \put(0,0){\includegraphics[width=\unitlength,page=2]{figures/4_data/twod_fov_tex.pdf}}%
    \put(0.1,0.1){\makebox(0,0)[lt]{\lineheight{1.25}\smash{\begin{tabular}[t]{l}$t$\end{tabular}}}}%
    \put(0.5,0.6){\color[rgb]{0,0,0}\makebox(0,0)[lt]{\lineheight{1.25}\smash{\begin{tabular}[t]{l}$r$\end{tabular}}}}%
    \put(0.2,1.1){\color[rgb]{0,0,0}\makebox(0,0)[lt]{\lineheight{1.25}\smash{\begin{tabular}[t]{l}$N$\end{tabular}}}}%
    \put(0.3,0.7){\color[rgb]{1,0,0}\makebox(0,0)[lt]{\lineheight{1.25}\smash{\begin{tabular}[t]{l}$\alpha$\end{tabular}}}}%
  \end{picture}%
\endgroup%

      \caption{}
      \label{fig:fov_graph}
      \end{subfigure}
      \hspace{.3cm}
      \begin{subfigure}{.3\textwidth}
      \includegraphics[width=\textwidth, trim=2.5cm 2cm 2.5cm 3.25cm, clip]{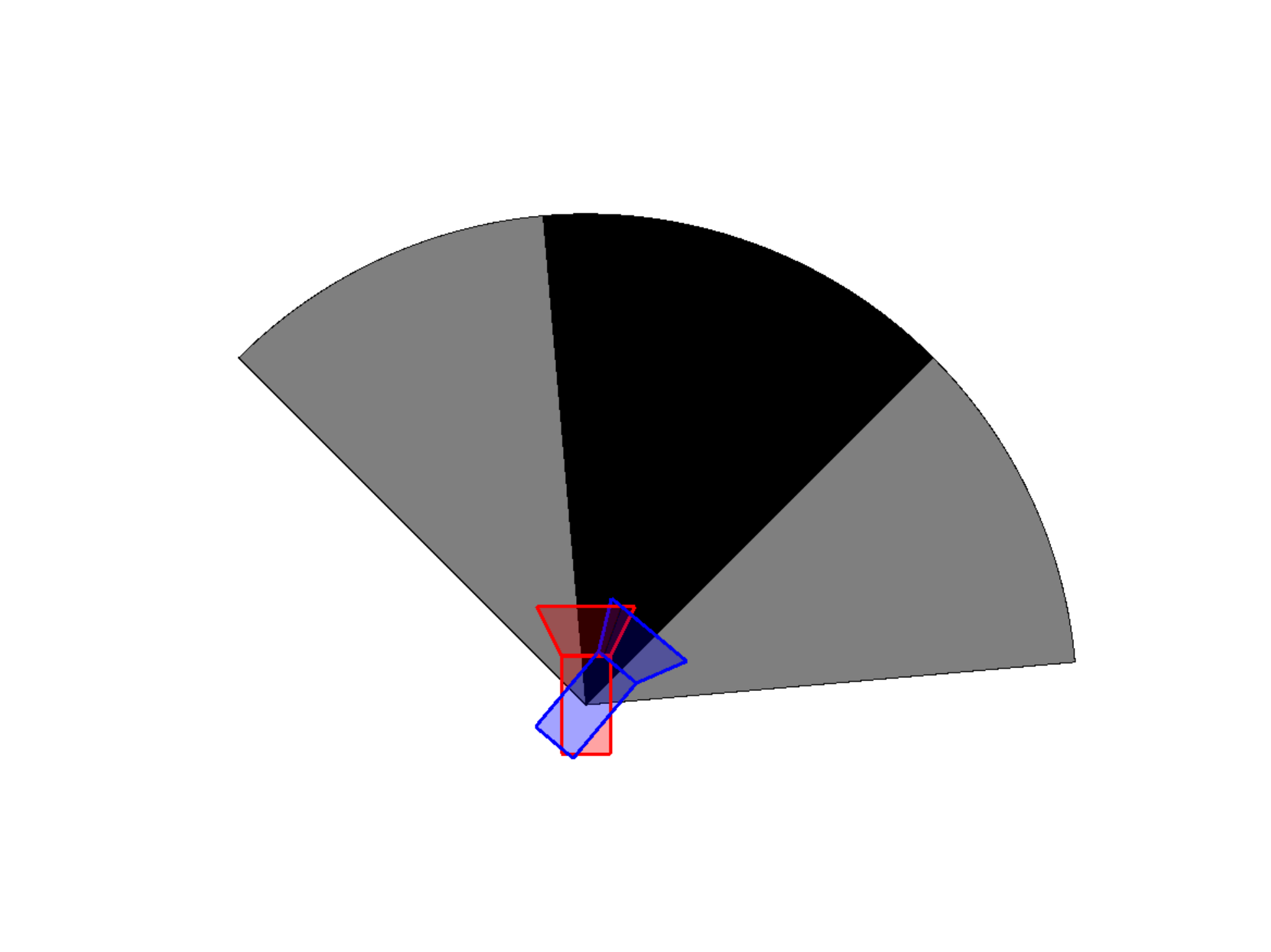}
      \vspace{.8mm}
                \caption{}
            \label{fig:msls_0m_40deg}
          \end{subfigure}
          \hspace{.3cm}
      \begin{subfigure}{0.3\textwidth}
      \includegraphics[width=\textwidth, trim=2.5cm 2cm 2.5cm 3.5cm, clip]{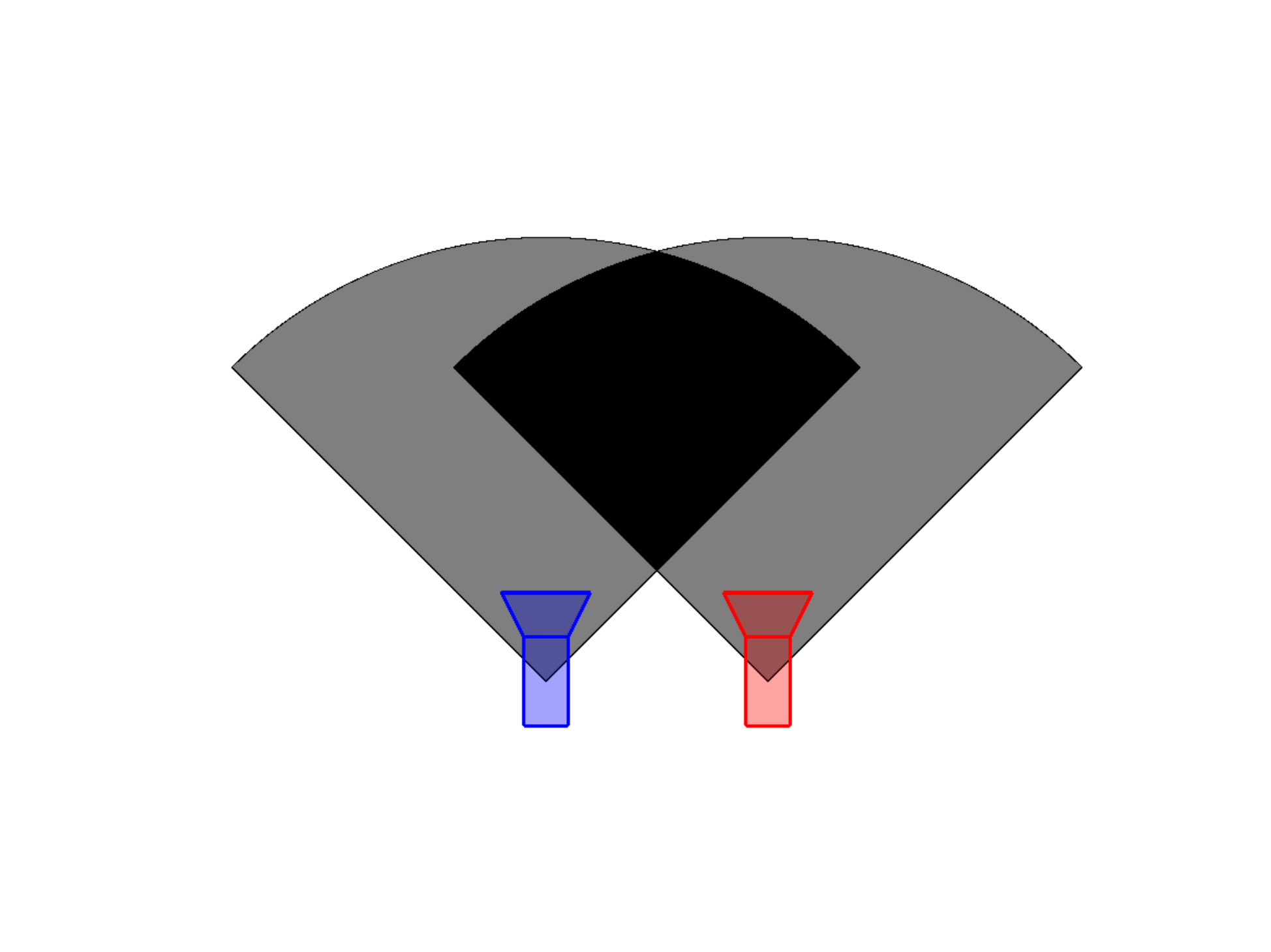}
      \vspace{.8mm}
        \caption{}
        \label{fig:msls_25m_0deg}
        \end{subfigure}
    \label{fig:fov}
    \caption{(a) FoV with angle $\theta$ and radius $r$. The point $\mathbf{t}$ is the camera location in the environment, and $\alpha$ is the camera orientation in the form of a compass angle with respect to the north $N$. (b) An example of FoV overal for two cameras 
    in the same position and with orientations $40 ^{\circ}$ apart. (c) An example of FoV overlap for two cameras 
    located 25m apart but with the same orientation. }
\end{figure}

\begin{figure}[t!]
  \centering
    \begin{subfigure}{0.31\textwidth}
        \includegraphics[height=1.8cm,width=\textwidth]{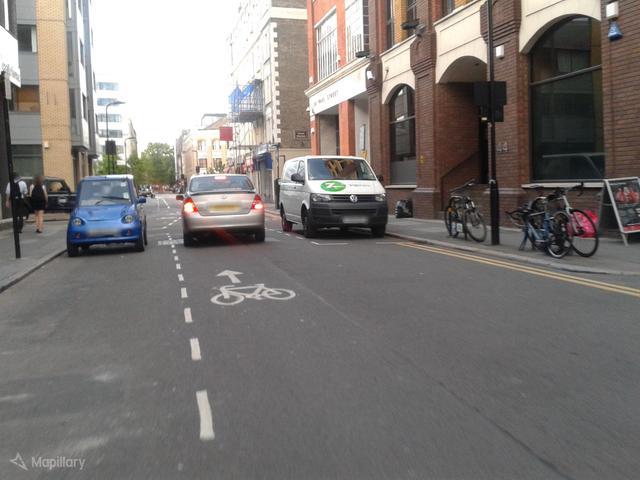}
                  \caption{}
\end{subfigure}
    \begin{subfigure}{0.31\textwidth}
        \includegraphics[height=1.8cm,width=\textwidth]{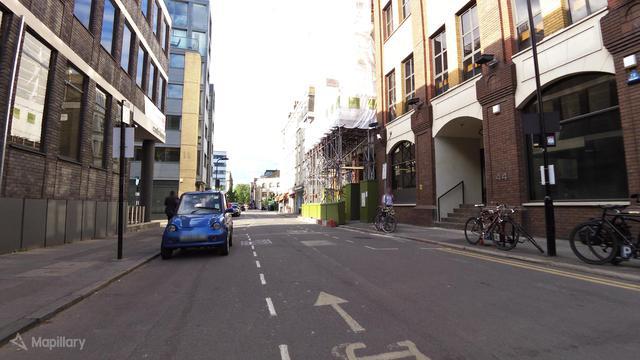}
                      \caption{}

    \end{subfigure}
    \begin{subfigure}{0.31\textwidth}
        \includegraphics[width=\textwidth]{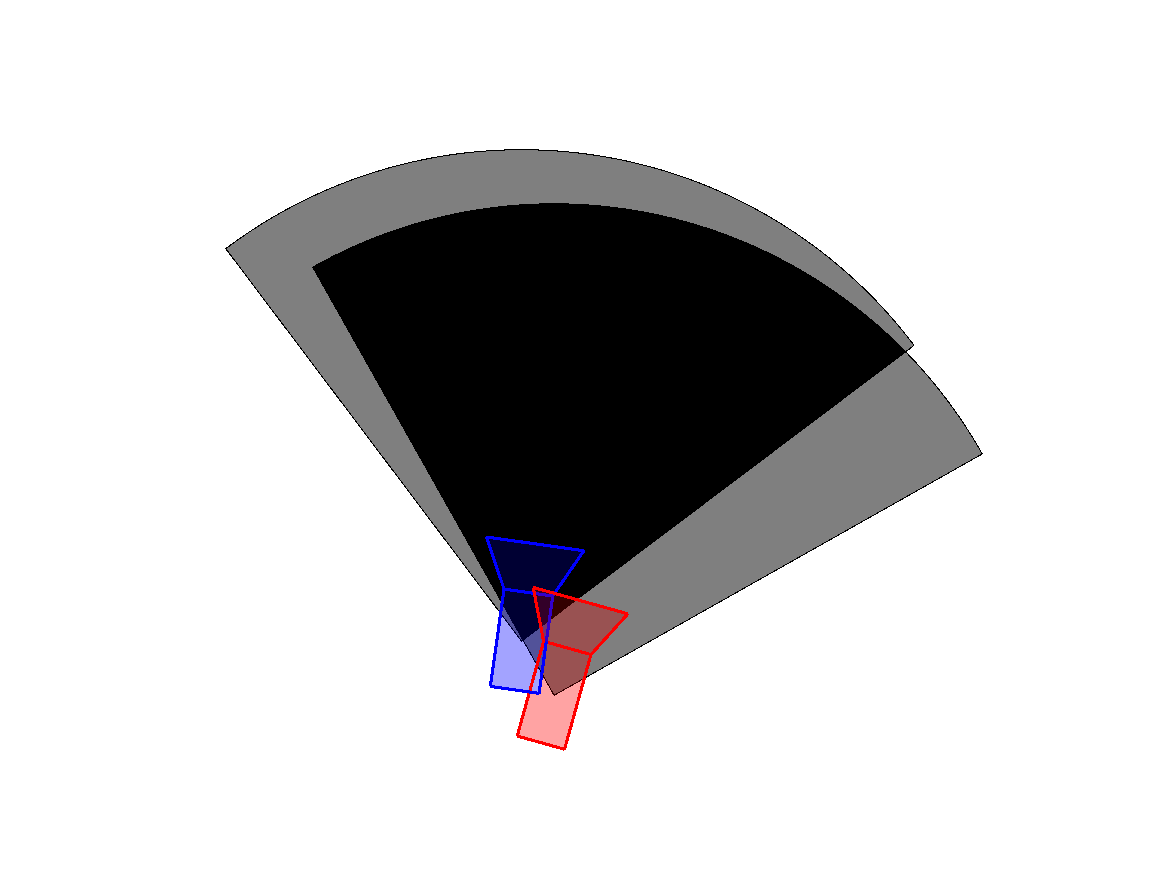}
\caption{FoV overlap 75.5\%}
\end{subfigure}
    
    \begin{subfigure}{0.31\textwidth}

      \includegraphics[width=\textwidth,height=1.8cm]{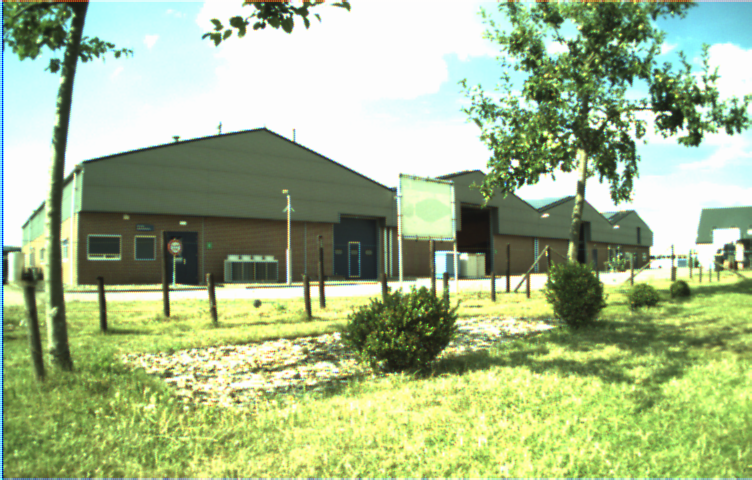}
                    \caption{}
\end{subfigure}
      \begin{subfigure}{0.31\textwidth}

\includegraphics[width=\textwidth,height=1.8cm]{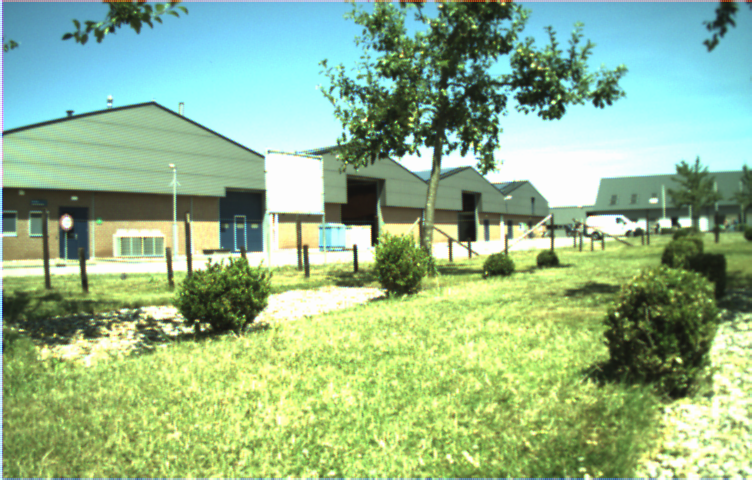}
                    \caption{}
\end{subfigure}
      \begin{subfigure}{0.31\textwidth}

      \includegraphics[width=\columnwidth, trim=1cm 2cm 1cm 2cm, clip]{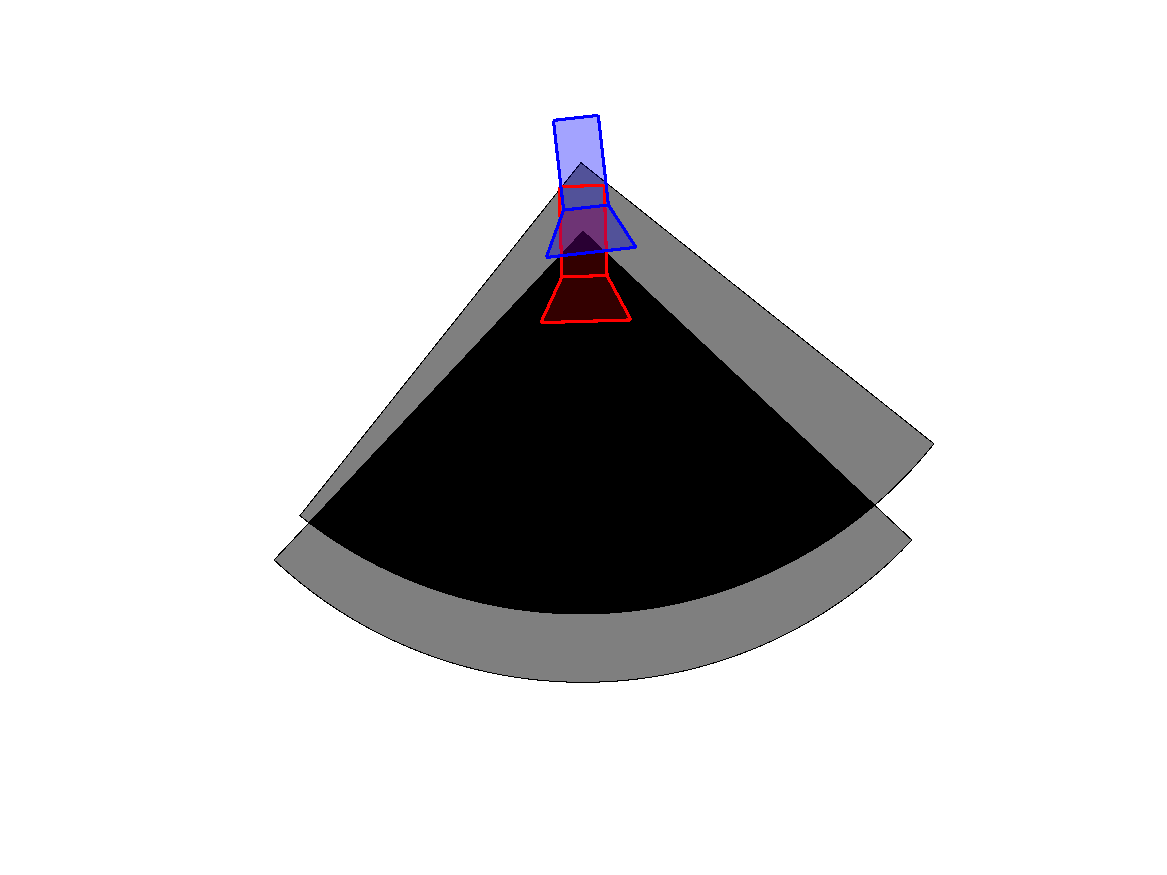}
        \caption{FoV overlap 75\%}
\end{subfigure}
      
      \begin{subfigure}{0.31\textwidth}

      \includegraphics[width=\textwidth,height=1.8cm]{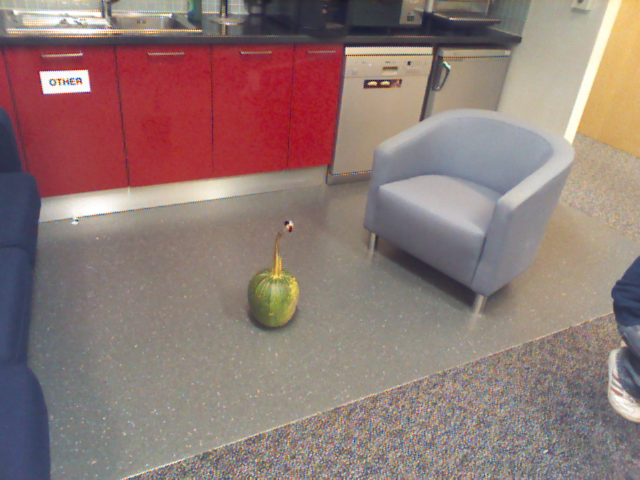}
                  \caption{}
\end{subfigure}
      \begin{subfigure}{0.31\textwidth}

\includegraphics[width=\textwidth,height=1.8cm]{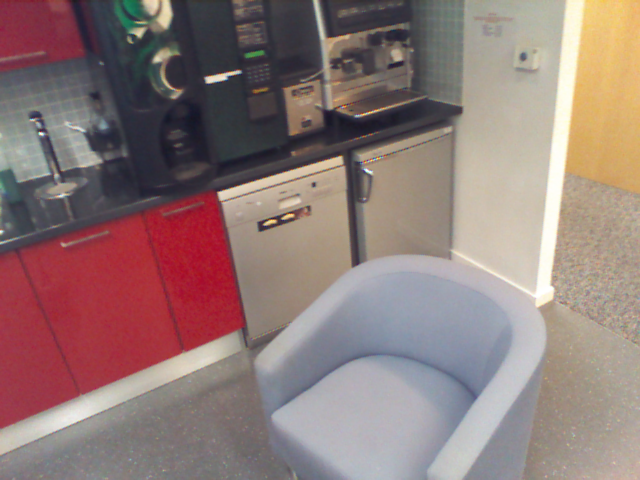}
              \caption{}
\end{subfigure}
      \begin{subfigure}{0.31\textwidth}

\includegraphics[width=\textwidth]{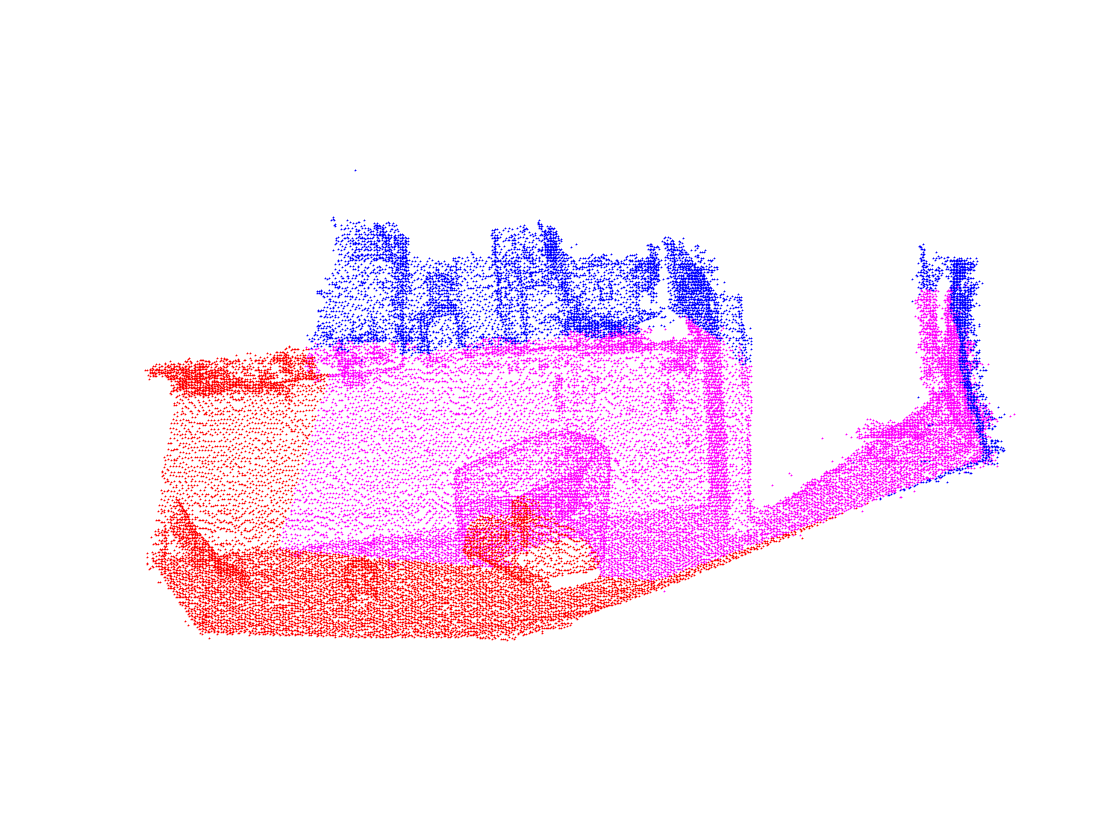}
        \caption{FoV overlap 50\%}

      \end{subfigure}

    \caption{Examples of graded similarity estimated for MSLS (first row) and TB-Places (second row) with the FoV overlap, and 7Scenes (third row) with 3D overlap (magenta color points). }
    \label{fig:examples-gt}
\end{figure}

\noindent\textbf{Graded similarity for MSLS and TB-Places: field of view overlap. } For MSLS and TB-Places dataset, images are provided with camera pose metadata
in the form of a vector $\mathbf{t}$ and orientation $\alpha$. The MSLS has UTM data and compass angle information associated to the images, while in TB-Places the images are provided with precise camera pose recorded with a laser tracker and IMU.

For an image, we build the field of view (FoV) of the camera as the sector of the circle centered at $\mathbf{t}$ with radius $r$, delimited by the angle range $[\alpha - \frac{\theta}{2}, \alpha +\frac{\theta}{2}]$ (see Fig.~\ref{fig:fov_graph}), where $\theta$ is the nominal size of the FoV of the camera concerned. The FoV overlap is the intersection-over-union (IoU) of the FoV of the cameras. In the first and second row of Figure~\ref{fig:examples-gt}, we show examples of the graded similarity estimated for pairs in the MSLS and TB-Places datasets. This approach differs from the camera frusta overlap~\cite{balntas2018relocnet} that needs 3D overlap measures for precise camera pose estimation. 



\noindent\textbf{Graded similarity for 7Scenes: 3D overlap. } The 7Scenes dataset has a 3D pointcloud for each scene, and 6DOF camera pose associated to the images. In this case, we can estimate the similarity overlap differently from the cases above. 
We project a pair of images onto the 3D pointcloud, so that we select the points associated to the two images and measure their intersection-over-union (IoU) as a measure of the image pair similarity. A similar strategy based on maximum inliers was used for hard-pair mining in~\cite{radenovic2018fine}.
In the third row of Figure~\ref{fig:examples-gt}, we show an example of graded similarity estimation for two images in the 7Scenes dataset.

\subsection{Place recognition pipeline }
\noindent\textbf{Embeddings.}  We use a fully convolutional backbone (ResNet~\cite{he2016deep}, VGG16~\cite{simonyan2015very} and ResNeXt~\cite{xie2017aggregated}) with a GeM pooling layer~\cite{radenovic2018fine}, which receives as input an image  $x \in R^{w_n \times h_n \times 3}$, and outputs a  representation $\hat{f}(x)  \in R^{d_m}$, where $d_m$ is the number of kernels of the last convolutional layer. We train $\hat{f}(x)$ using a contrastive learning framework. 

\noindent\textbf{Training batch composition.} Batch composition is an important part of model training. 
For contrastive architectures, the selection of meaningful image tuples is crucial for the correct optimization of the model. If the selected tuples are too challenging, the training might become unstable~\cite{shi2016embedding}. If they are too easy, the learning might stall. This, coupled with binary pairwise labels, makes necessary to use complex descriptor-based mining strategies to ensure model convergence. The \emph{hard-negative mining} strategy needed to train contrastive networks~\cite{Arandjelovic2017,hausler2021patch,liu2019stochastic} periodically computes the descriptor of all training images and their pairwise distance to select certain pairs (tuples) of images to be used for the subsequent training steps. This is a memory- and computation-expensive procedure. 

We do not perform hard-pair mining. We instead compose the training batches taking into account the graded similarity labels that we computed. 
We balance the pairs in the training batches on the basis of their annotated degree of similarity.
For each batch, we make sure to select $50\%$ of positive pairs (similarity higher than $50\%$), $25\%$ of soft negative samples (similarity higher than $0\%$ and lower than $50\%$) and $25\%$ of hard negatives ($0\%$ similarity) -- see Section~\ref{sec:discussion} for results. 

\noindent\textbf{Image retrieval.} Let us consider a set $X$ of reference images with a known camera location, and a set $Y$ of query images taken from unknown positions. In order to localize the camera that took the query images, similar images to the query are to be retrieved from the reference set. We compute the descriptors of the reference images $\hat{f}(x)  \,\forall x \in X$, and of the query images $\hat{f}(y)\, \forall y \in Y$. For a given query descriptor $\hat{f}(y)$, image retrieval is performed by nearest neighbor search within the reference descriptors $\hat{f}(x)\, \forall x \in X$, retrieving $k$ images ranked by the closest descriptor distance.

\subsection{Performance measures}
We apply widely used place recognition evaluation protocols and consider a query as correctly identified if any of the top-k retrieved images are annotated as a positive match~\cite{Sattler2012,Arandjelovic2017,msls}. We computed the following metrics.
For the MSLS, Pittsburgh30k, Tokyo24/7 (Pittsburgh250k, TokyoTM, TrimBot2020 and 7Scences in the supplementary material) we compute the \textbf{Top-k recall (R@k)}. It measures the percentage of queries for which at least a correct map image is present among their $k$ nearest neighbors retrieved. For the RobotCar Seasons v2 and the Extended CMU datasets, we compute the \textbf{percentage of correctly localized queries}. It measures the amount of images that are correctly retrieved for a given translation and rotation threshold. 



\section{Results and discussion}
\label{sec:discussion}

\begin{table}[!t]
\centering
\caption{Effect of graded similarity labels on batch composition and model training.}
\label{tab:msls-labels-effect}
\resizebox{\textwidth}{!}{%
\setlength\tabcolsep{1.5pt}

\begin{tabular}{@{}l@{\hspace{4\tabcolsep}}l@{\hspace{4\tabcolsep}}c@{\hspace{8\tabcolsep}}ccc@{\hspace{8\tabcolsep}}ccc}
 &  &  & \multicolumn{3}{c}{\textbf{MSLS-Val}} & \multicolumn{3}{c}{\textbf{MSLS-Test}} \\

\textbf{Method} & \textbf{Loss} & \textbf{Batch}  & \textbf{R@1} &\textbf{ R@5} & \textbf{R@10} & \textbf{R@1} & \textbf{R@5} & \textbf{R@10}  \\
\toprule
VGG-GeM & CL & binary  &  47.0 & 60.3 & 65.5 & 27.9 & 40.5 & 46.5  \\
VGG-GeM & GCL & binary  &  57.4 & 73.4 & 76.9 & 35.9 & 49.3 & 57.8  \\ \midrule
VGG-GeM & CL & graded  &  45.8 & 60.1 & 65.1 & 28.0 & 40.8 & 47.0  \\
VGG-GeM & GCL & graded  &  65.9 & 77.8 & 81.4 & 41.7 & 55.7 & 60.6  \\
 \bottomrule
\end{tabular}
}
\end{table}

\noindent\textbf{Graded similarity for batch composition and model training.} We carry out a baseline experiment to evaluate the impact of the new graded similarity labels on the effectiveness of the learned descriptors. We analyze their contribution to the composition of the batches, and directly to the training of the network by using them in combination with the proposed GCL. We first consider the traditional binary labels only, and compose batches by balancing positive and negative pairs. Subsequently, we compose the batches by considering the new graded similarity labels, and select $50\%$ of positive pairs (similarity higher than $50\%$), $25\%$ of soft negative samples (similarity higher than $0\%$ and lower than $50\%$) and $25\%$ of hard negatives ($0\%$ similarity). 

In Table~\ref{tab:msls-labels-effect}, we report the results using a VGG16 backbone on the MSLS dataset. These results demonstrate that the proposed graded similarity labels are especially useful for training descriptors that perform better in nearest neighbor search retrieval, and also contribute to form better balanced batches to exploit the data in a more efficient way. In the following, all experiments use the batch composition based on graded similarity.

\begin{table*}[t!]
\centering
\renewcommand{\arraystretch}{1}
\caption{Comparison to state-of-the-art methods on benchmark datasets. All methods are trained on the MSLS training set. Our top results are underlined, while overall best results are in bold. Methods using re-ranking are in the middle part of the table and marked with $^\star$.}
\label{tab:msls-sota}

\resizebox{\textwidth}{!}{%
\setlength\tabcolsep{1.5pt}
\revised{
\begin{tabular}{@{}lcc@{\hspace{4\tabcolsep}}ccc@{\hspace{4\tabcolsep}}ccc@{\hspace{4\tabcolsep}}ccc@{\hspace{4\tabcolsep}}ccc@{\hspace{4\tabcolsep}}ccc@{\hspace{4\tabcolsep}}ccc@{}}
 &  &  & \multicolumn{3}{c}{\textbf{MSLS-Val}} & \multicolumn{3}{c}{\textbf{MSLS-Test}} & \multicolumn{3}{c}{\textbf{Pitts30k}} & \multicolumn{3}{c}{\textbf{Tokyo24/7}} & \multicolumn{3}{c}{\textbf{RobotCar Seasons v2}} & \multicolumn{3}{c}{\textbf{Extended CMU Seasons}} \\ 
\textbf{Method} & \textbf{PCA$_w$} & \textbf{Dim} & \textbf{R@1} & \textbf{R@5} & \textbf{R@10} & \textbf{R@1} & \textbf{R@5} & \textbf{R@10} & \textbf{R@1} & \textbf{R@5} & \textbf{R@10} & \textbf{R@1} & \textbf{R@5 }& \textbf{R@10} & \textbf{.25m/2$\degree$} & \textbf{.5m/5$\degree$ }& \textbf{5m/10$\degree$ }& \textbf{.25m/2$\degree$ }& \textbf{.5m/5$\degree$ }& \textbf{5m/10$\degree$} \\\toprule 
NetVLAD 64  & N & 32768 & 44.6 & 61.1 & 66.4 & 28.8 & 44.0 & 50.7 & 40.4 & 64.5 & 74.2 & 11.4 & 24.1 & 31.4 & 2.0 & 9.2 & 45.5 & 1.3 & 4.5 & 31.9 \\
NetVLAD 64  & Y & 4096 & 70.1 & 80.8 & 84.9 & 45.1 & 58.8 & 63.7 & 68.6 & 84.7 & 88.9 & 34.0 & 47.6 & 57.1 & 4.2 & 18.0 & 68.1 & 3.9 & 12.1 & 58.4 \\
NetVLAD 16  & N & 8192 & 49.5 & 65.0 & 71.8 & 29.3 & 43.5 & 50.4 & 48.7 & 70.6 & 78.9 & 13.0 & 33.0 & 43.8 & 1.8 & 9.2 & 48.4 & 1.7 & 5.5 & 39.1 \\
NetVLAD 16  & Y & 4096 & 70.5 & 81.1 & 84.3 & 39.4 & 53.0 & 57.5 & 70.3 & 84.1 & 89.1 & 37.8 & 53.3 & 61.0 & 4.8 & 17.9 & 65.3 & 4.4 & 13.7 & 61.4 \\
 
TransVPR~\cite{wang2022transvpr} & - & - & 70.8 & 85.1 & 86.9 & 48 & 67.1 & 73.6 & 73.8 & 88.1 & 91.9 & - & - & - & 2.9 & 11.4 & 58.6 & - & - & - \\ \hline
 
SP-SuperGlue$^\star$~\cite{sarlin20superglue} & - & - & 78.1 & 81.9 & 84.3 & 50.6 & 56.9 & 58.3 & 87.2 & 94.8 & 96.4 & 88.2 & 90.2 & 90.2 & 9.5 & \bfseries 35.4 & 85.4 & 9.5 & 30.7 & \textbf{96.7} \\
DELG$^\star$~\cite{delg} & - & - & 83.2 & 90.0 & 91.1 & 52.2 & 61.9 & 65.4 & \bfseries 89.9 & \bfseries 95.4 & \bfseries 96.7 & \bfseries 95.9 & \bfseries 96.8 & \bfseries 97.1 & 2.2 & 8.4 & 76.8 & 5.7 & 21.1 & 93.6 \\

Patch NetVLAD$^\star$~\cite{hausler2021patch} & Y & 4096 & 79.5 & 86.2 & 87.7 & 48.1 & 57.6 & 60.5 & 88.7 & 94.5 & 95.9 & 86.0 & 88.6 & 90.5 & 9.6 & 35.3 & \textbf{90.9} & \textbf{11.8} & \textbf{36.2} & 96.2 \\

TransVPR$^\star$~\cite{wang2022transvpr} & - & - & \textbf{86.8} & \textbf{91.2} & 92.4 & \textbf{63.9} & 74 & 77.5 & 89 & 94.9 & 96.2 & - & - & - & \textbf{9.8} & 34.7 & 80 & - & - & - \\ \midrule
NetVLAD-GCL
 &
  N &
  32768 &
  62.7 &
  75.0 &
  79.1 &
  41.0 &
  55.3 &
  61.7 &
  52.5 &
  74.1 &
  81.7 &
  20.3 &
  45.4 &
  49.5 &
  3.3 &
  14.1 &
  58.2 &
  3.0 &
  9.7 &
  52.3 \\
NetVLAD-GCL
  &
  Y &
  4096 &
  63.2 &
  74.9 &
  78.1 &
  41.5 &
  56.2 &
  61.3 &
  53.5 &
  75.2 &
  82.9 &
  28.3 &
  41.9 &
  54.9 &
  3.4 &
  14.2 &
  58.8 &
  3.1 &
  9.7 &
  52.4 \\ 
VGG-GeM-GCL  & N & 512 & 65.9 & 77.8 & 81.4 & 41.7 & 55.7 & 60.6 & 61.6 & 80.0 & 86.0 & 34.0 & 51.1 & 61.3 & 3.7 & 15.8 & 59.7 & 3.6 & 11.2 & 55.8 \\
VGG-GeM-GCL  & Y & 512 & 72.0 & 83.1 & 85.8 & 47.0 & 60.8 & 65.5 & 73.3 & 85.9 & 89.9 & 47.6 & 61.0 & 69.2 & 5.4 & \underline{21.9} & 69.2 & 5.7 & 17.1 & 66.3 \\
ResNet50-GeM-GCL & N & 2048 & 66.2 & 78.9 & 81.9 & 43.3 & 59.1 & 65.0 & 72.3 & 87.2 & 91.3 & 44.1 & 61.0 & 66.7 & 2.9 & 14.0 & 58.8 & 3.8 & 11.8 & 61.6 \\
ResNet50-GeM-GCL & Y & 1024 & 74.6 & 84.7 & 88.1 & 52.9 & 65.7 & 71.9 & 79.9 & 90.0 & 92.8 & 58.7 & 71.1 & 76.8 & 4.7 & 20.2 & 70.0 & 5.4 & 16.5 & 69.9 \\
ResNet152-GeM-GCL& N & 2048 & 70.3 & 82.0 & 84.9 & 45.7 & 62.3 & 67.9 & 72.6 & 87.9 & 91.6 & 34.0 & 51.8 & 60.6 & 2.9 & 13.1 & 63.5 & 3.6 & 11.3 & 63.1 \\
ResNet152-GeM-GCL & Y & 2048 & 79.5 & 88.1 & 90.1 & 57.9 & 70.7 & 75.7 & \underline{80.7} & \underline{91.5} & \underline{93.9} & \underline{69.5} & \underline{81.0} & \underline{85.1} & \underline{6.0} & 21.6 & 72.5 & 5.3 & 16.1 & 66.4 \\
ResNeXt-GeM-GCL & N & 2048 & 75.5 & 86.1 & 88.5 & 56.0 & 70.8 & 75.1 & 64.0 & 81.2 & 86.6 & 37.8 & 53.6 & 62.9 & 2.7 & 13.4 & 65.2 & 3.5 & 10.5 & 58.8 \\
ResNeXt-GeM-GCL & Y & 1024 & \underline{ 80.9} & \underline{90.7} & \underline{\textbf{92.6}} & \underline{62.3} & \underline{\textbf{76.2}} & \underline{\textbf{81.1}} & 79.2 & 90.4 & 93.2 & 58.1 & 74.3 & 78.1 & 4.7 & 21.0 & \underline{74.7} & \underline{6.1} & \underline{18.2} & \underline{74.9} \\ \bottomrule

\end{tabular}
}}
\end{table*}

\noindent\textbf{Comparison with existing works.}
We compared our results with several place recognition works. We considered methods that use global descriptors like NetVLAD~\cite{Arandjelovic2017} (with 16 and 64 clusters in the VLAD layer) and methods based on two-stages retrieval and re-ranking pipelines, such as Patch-NetVLAD~\cite{hausler2021patch}, DELG~\cite{delg} and SuperGlue~\cite{sarlin20superglue}. We compared also against TransVPR~\cite{wang2022transvpr}, a transformer with and without a re-ranking stage. 
Table~\ref{tab:msls-sota} reports the results of our method in comparison to others. All the methods included in the table are based on backbones trained on the MSLS datasets. 
The results of Patch-NetVLAD and TransVPR are taken from the respective papers, which also contain those of DELG and SuperGlue. 
When trained with VGG16 as backbone, our model (VGG16-GeM-GCL) obtains an absolute improvement of R@5
equal to 11.7\% compared to NetVLAD-64. 
This shows that the proposed graded similarity labels and the GCL function contribute to learn more powerful descriptors for place recognition, while keeping the complexity of the training process lower as hard-pair mining is not used. The result improvement holds also when the descriptors are post-processed with PCA whitening.

The data- and memory-efficiency of our pipeline allows us to easily train more powerful backbones, such as ResNeXt, that is instead tricky to do for other methods due to memory and compute requirements. Our ResNeXt+GCL outperforms the best method on the MSLS test set, namely TransVPR without re-ranking by 8.9\% and with re-ranking by 2.6\% (absolute improvement of R@5). It compares favorably with re-ranking based methods such as Patch-NetVLAD, DELG and SuperGLUE improving the R@5 by 18.6\%, 14.3\% and  18.3\%, respectively. 
We point out that we do not re-rank the retrieved images, and purposely keep the complexity of the steps at the strict necessary to perfom the VPR retrieval task. 
We attribute our high results mainly to the effectiveness of the descriptors learned with the GCL function using the new graded similarity labels.

\begin{table*}[!t]
\centering
\caption{\revised{Ablation study on backbone, Contrastive (CL) vs Generalized Contrastive (GCL) loss, and PCA. All models are trained on MSLS.} }
\label{tab:msls-ablation}
\resizebox{\textwidth}{!}{%
\setlength\tabcolsep{1.5pt}
\revised{
\begin{tabular}{@{}llcc@{\hspace{4\tabcolsep}}ccc@{\hspace{4\tabcolsep}}ccc@{\hspace{4\tabcolsep}}ccc@{\hspace{4\tabcolsep}}ccc@{\hspace{4\tabcolsep}}ccc@{\hspace{4\tabcolsep}}ccc@{}}
 &  &  &  & \multicolumn{3}{c}{\textbf{MSLS-Val}} & \multicolumn{3}{c}{\textbf{MSLS-Test}} & \multicolumn{3}{c}{\textbf{Pitts30k}} & \multicolumn{3}{c}{\textbf{Tokyo24/7}} & \multicolumn{3}{c}{\textbf{RobotCar Seasons v2}} & \multicolumn{3}{c}{\textbf{Extended CMU Seasons}} \\

\textbf{Method} & \textbf{Loss} & \textbf{PCA$_w$} & \textbf{Dim} & \textbf{R@1} &\textbf{ R@5} & \textbf{R@10} & \textbf{R@1} & \textbf{R@5} & \textbf{R@10} &\textbf{ R@1} & \textbf{R@5} & \textbf{R@10} & \textbf{R@1} & \textbf{R@5 }& \textbf{R@10} & \textbf{0.25m/2$\degree$} & \textbf{0.5m/5$\degree$} & \textbf{5.0m/10$\degree$} & \textbf{0.25m/2$\degree$} & \textbf{0.5m/5$\degree$} & \textbf{5.0m/10$\degree$} \\
\toprule
\multirow{4}{*}{NetVLAD} &
  CL &
  N &
  32768 &
  38.5 &
  56.6 &
  64.1 &
  24.8 &
  39.1 &
  46.1 &
  24.7 &
  48.3 &
  61.1 &
  7.0 &
  18.1 &
  24.8 &
  0.9 &
  5.2 &
  31.5 &
  1.0 &
  1.4 &
  11.1 \\
 &
  GCL &
  N &
  32768 &
  62.7 &
  75.0 &
  79.1 &
  41.0 &
  55.3 &
  61.7 &
  52.5 &
  74.1 &
  81.7 &
  20.3 &
  45.4 &
  49.5 &
  3.3 &
  14.1 &
  58.2 &
  3.0 &
  9.7 &
  52.3 \\
 &
  CL &
  Y &
  4096 &
  39.6 &
  60.3 &
  65.3 &
  26.4 &
  40.5 &
  48.2 &
  27.5 &
  51.6 &
  64.1 &
  6.7 &
  16.2 &
  25.7 &
  0.0 &
  0.0 &
  0.0 &
  1.0 &
  3.3 &
  25.4 \\
 &
  GCL &
  Y &
  4096 &
  63.2 &
  74.9 &
  78.1 &
  41.5 &
  56.2 &
  61.3 &
  53.5 &
  75.2 &
  82.9 &
  28.3 &
  41.9 &
  54.9 &
  3.4 &
  14.2 &
  58.8 &
  3.1 &
  9.7 &
  52.4 \\ \midrule
  
\multirow{4}{*}{VGG-GeM} & CL & N & 512 & 47.0 & 60.3 & 65.5 & 27.9 & 40.5 & 46.5 & 51.2 & 71.9 & 79.7 & 24.1 & 39.4 & 47.0 & 3.1 & 13.2 & 55.0 & 2.8 & 8.6 & 44.5 \\
 & GCL & N & 512 & 65.9 & 77.8 & 81.4 & 41.7 & 55.7 & 60.6 & 61.6 & 80.0 & 86.0 & 34.0 & 51.1 & 61.3 & 3.7 & 15.8 & 59.7 & 3.6 & 11.2 & 55.8 \\
 & CL & Y & 512 & 61.4 & 75.1 & 78.5 & 36.3 & 49.0 & 54.1 & 64.7 & 81.5 & 86.8 & 36.2 & 54.0 & 57.8 & 4.2 & 18.7 & 62.5 & 4.4 & 13.4 & 56.5 \\
 & GCL & Y & 512 & 72.0 & 83.1 & 85.8 & 47.0 & 60.8 & 65.5 & 73.3 & 85.9 & 89.9 & 47.6 & 61.0 & 69.2 & 5.4 & 21.9 & 69.2 & 5.7 & 17.1 & 66.3 \\ \midrule
\multirow{4}{*}{ResNet50-GeM} & CL & N & 2048 & 51.4 & 66.5 & 70.8 & 29.7 & 44.0 & 50.7 & 61.5 & 80.0 & 86.9 & 30.8 & 46.0 & 56.5 & 3.2 & 15.4 & 61.5 & 3.2 & 9.6 & 49.5 \\
 & GCL & N & 2048 & 66.2 & 78.9 & 81.9 & 43.3 & 59.1 & 65.0 & 72.3 & 87.2 & 91.3 & 44.1 & 61.0 & 66.7 & 2.9 & 14.0 & 58.8 & 3.8 & 11.8 & 61.6 \\
 & CL & Y & 1024 & 63.2 & 76.6 & 80.7 & 37.9 & 53.0 & 58.5 & 66.2 & 82.2 & 87.3 & 36.2 & 51.8 & 61.0 & 5.0 & 21.1 & 66.5 & 4.7 & 13.4 & 51.6 \\
 & GCL & Y & 1024 & 74.6 & 84.7 & 88.1 & 52.9 & 65.7 & 71.9 & 79.9 & 90.0 & 92.8 & 58.7 & 71.1 & 76.8 & 4.7 & 20.2 & 70.0 & 5.4 & 16.5 & 69.9 \\ \midrule
\multirow{4}{*}{ResNet152-GeM} & CL & N & 2048 & 58.0 & 72.7 & 76.1 & 34.1 & 50.8 & 56.8 & 66.5 & 83.8 & 89.5 & 34.6 & 57.1 & 63.5 & 3.3 & 15.2 & 64.0 & 3.2 & 9.7 & 52.2 \\
 & GCL & N & 2048 & 70.3 & 82.0 & 84.9 & 45.7 & 62.3 & 67.9 & 72.6 & 87.9 & 91.6 & 34.0 & 51.8 & 60.6 & 2.9 & 13.1 & 63.5 & 3.6 & 11.3 & 63.1 \\
 & CL & Y & 2048 & 66.9 & 80.9 & 83.8 & 44.8 & 59.2 & 64.8 & 71.2 & 85.8 & 89.8 & 54.3 & 68.9 & 75.6 & \textbf{6.1 } & \textbf{23.5} & 68.9 & 4.8 & 14.2 & 55.0 \\
 & GCL & Y & 2048 & 79.5 & 88.1 & 90.1 & 57.9 & 70.7 & 75.7 & \textbf{80.7} & \textbf{91.5} & \textbf{93.9} & \textbf{69.5} & \textbf{81.0} & \textbf{85.1} & 6.0 & 21.6 & 72.5 & 5.3 & 16.1 & 66.4 \\ \midrule
\multirow{4}{*}{ResNeXt-GeM} & CL & N & 2048 & 62.6 & 76.4 & 79.9 & 40.8 & 56.5 & 62.1 & 56.0 & 77.5 & 85.0 & 37.8 & 54.9 & 62.5 & 1.9 & 10.4 & 54.8 & 2.9 & 9.0 & 52.6 \\
 & GCL & N & 2048 & 75.5 & 86.1 & 88.5 & 56.0 & 70.8 & 75.1 & 64.0 & 81.2 & 86.6 & 37.8 & 53.6 & 62.9 & 2.7 & 13.4 & 65.2 & 3.5 & 10.5 & 58.8 \\
 & CL & Y & 1024 & 74.3 & 87.0 & 89.6 & 49.9 & 63.8 & 69.4 & 70.9 & 85.7 & 90.2 & 50.8 & 67.6 & 74.3 & 3.8 & 17.2 & 68.2 & 4.9 & 14.4 & 61.7 \\
 & GCL & Y & 1024 & \textbf{80.9} & \textbf{90.7} & \textbf{92.6} & 62.3 & \textbf{76.2} & \textbf{81.1} & 79.2 & 90.4 & 93.2 & 58.1 & 74.3 & 78.1 & 4.7 & 21.0 & \textbf{74.7} & \textbf{6.1} & \textbf{18.2} & \textbf{74.9} 
 \\\bottomrule
\end{tabular}
}}
\end{table*}

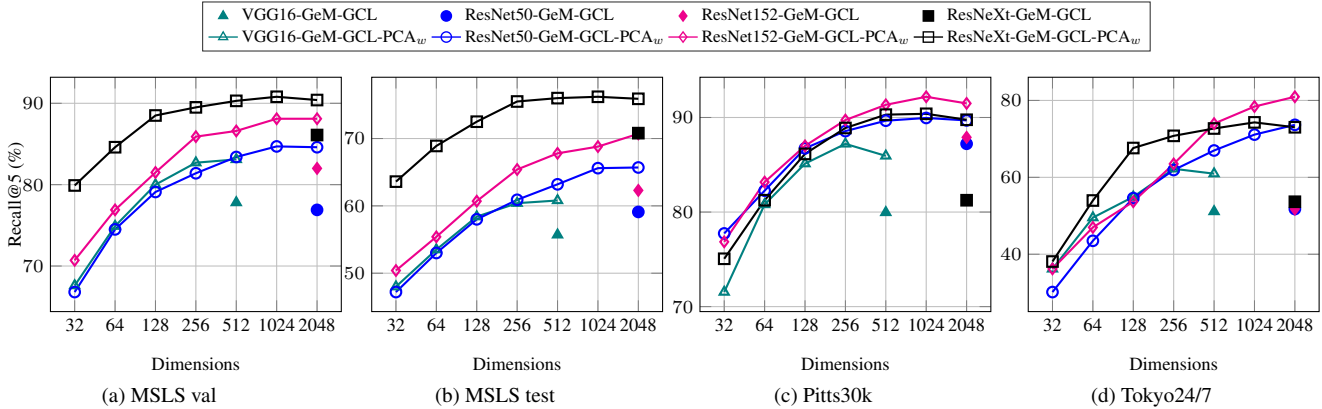
\begin{figure*}[!t]
\centering
\begin{minipage}[c]{\textwidth}
\begin{subfigure}{\textwidth}
\centering
\hspace{0.5cm}
\begin{tikzpicture}[thick, scale=.65, every node/.style={scale=1}] 
    \begin{axis}[%
    hide axis,
    xmin=20,
    xmax=50,
    ymin=0,
    ymax=0.4,
    legend style={draw=white!15!black,legend cell align=left,legend columns=4}, height=4.75cm, width=.61\textwidth
    ]
\addlegendimage{teal,thick, mark=triangle*, mark size=3pt, only marks}
\addlegendentry{VGG16-GeM-GCL};
\addlegendimage{blue,thick, mark=*, mark size=3pt, only marks}
\addlegendentry{ResNet50-GeM-GCL};
\addlegendimage{magenta,thick, mark=diamond*, mark size=3pt, only marks}
\addlegendentry{ResNet152-GeM-GCL};
\addlegendimage{black,thick, mark=square*, mark size=3pt, only marks}
\addlegendentry{ResNeXt-GeM-GCL};

\addlegendimage{teal,thick, mark=triangle, mark size=3pt}
\addlegendentry{VGG16-GeM-GCL-PCA$_w$};

\addlegendimage{blue,thick, mark=o, mark size=3pt}
\addlegendentry{ResNet50-GeM-GCL-PCA$_w$};
\addlegendimage{magenta,thick, mark=diamond, mark size=3pt}
\addlegendentry{ResNet152-GeM-GCL-PCA$_w$};
\addlegendimage{black,thick, mark=square, mark size=3pt}
\addlegendentry{ResNeXt-GeM-GCL-PCA$_w$};

    \end{axis}
    \end{tikzpicture} 
\end{subfigure}
\end{minipage}

\begin{minipage}[c]{\textwidth}
\vspace{-30pt}
\setcounter{subfigure}{0}

\scriptsize
\begin{subfigure}{0.24\textwidth}
\begin{tikzpicture}[thick]
\begin{axis}
[xlabel=Dimensions,
ylabel=Recall@5 (\%), ylabel style={at={(axis description cs:0.2,0.5)}},grid, xmode=log,log basis x={2}, xtick={32,64,128,256, 512,1024,2048}, xticklabels={32,64,128,256, 512,1024,2048}, width=1.30\textwidth
]
\addplot[teal,thick, mark=triangle*, mark size=2pt] table [x=d, y=r, col sep=comma] {plot_data/whitening/MSLS/MSLS_soft_vgg16_GeM_L2_0-1_2_nowhiten_toplot};
\addplot[teal, mark=triangle, thick, mark size=2pt] table [x=d, y=r, col sep=comma] {plot_data/whitening/MSLS/MSLS_soft_vgg16_GeM_L2_0-1_2_whiten_toplot};
\addplot[blue,thick, mark=*, mark size=2pt] table [x=d, y=r, col sep=comma] {plot_data/whitening/MSLS/MSLS_soft_ResNet50_GeM_L2_0-1_2_nowhiten_toplot};
\addplot[blue, mark=o, thick, mark size=2pt] table [x=d, y=r, col sep=comma] {plot_data/whitening/MSLS/MSLS_soft_ResNet50_GeM_L2_0-1_2_whiten_toplot};
\addplot[magenta,thick, mark=diamond*, mark size=2pt] table [x=d, y=r, col sep=comma] {plot_data/whitening/MSLS/MSLS_soft_ResNet152_GeM_L2_0-1_2_nowhiten_toplot};
\addplot[magenta, mark=diamond, thick, mark size=2pt] table [x=d, y=r, col sep=comma] {plot_data/whitening/MSLS/MSLS_soft_ResNet152_GeM_L2_0-1_2_whiten_toplot};
\addplot[black,thick, mark=square*, mark size=2pt] table [x=d, y=r, col sep=comma] {plot_data/whitening/MSLS/MSLS_soft_resnext_GeM_L2_0-1_2_nowhiten_toplot};
\addplot[black, mark=square, thick, mark size=2pt] table [x=d, y=r, col sep=comma] {plot_data/whitening/MSLS/MSLS_soft_resnext_GeM_L2_0-1_2_whiten_toplot};
\end{axis}
 \label{fig:whitening_msls_val}
\end{tikzpicture}
\captionof{figure}{MSLS val}
\end{subfigure} \hspace{1.5mm}
\begin{subfigure}{0.24\textwidth}
\begin{tikzpicture}[thick]
\begin{axis}
[xlabel=Dimensions, ,grid, xmode=log,log basis x={2}, xtick={32,64,128,256, 512,1024,2048}, xticklabels={32,64,128,256, 512,1024,2048}, width=1.30\textwidth
]
\addplot[teal,thick, mark=triangle*, mark size=2pt] table [x=d, y=r, col sep=comma] {plot_data/whitening/MSLS_test/MSLS_soft_vgg16_GeM_L2_0-1_2_nowhiten_toplot.txt};
\addplot[teal, mark=triangle, thick, mark size=2pt] table [x=d, y=r, col sep=comma] {plot_data/whitening/MSLS_test/MSLS_soft_vgg16_GeM_L2_0-1_2_whiten_toplot.txt};
\addplot[blue,thick, mark=*, mark size=2pt] table [x=d, y=r, col sep=comma] {plot_data/whitening/MSLS_test/MSLS_soft_ResNet50_GeM_L2_0-1_2_nowhiten_toplot.txt};
\addplot[blue, mark=o, thick, mark size=2pt] table [x=d, y=r, col sep=comma] {plot_data/whitening/MSLS_test/MSLS_soft_ResNet50_GeM_L2_0-1_2_whiten_toplot.txt};
\addplot[magenta,thick, mark=diamond*, mark size=2pt] table [x=d, y=r, col sep=comma] {plot_data/whitening/MSLS_test/MSLS_soft_ResNet152_GeM_L2_0-1_2_nowhiten_toplot.txt};
\addplot[magenta, mark=diamond, thick, mark size=2pt] table [x=d, y=r, col sep=comma] {plot_data/whitening/MSLS_test/MSLS_soft_ResNet152_GeM_L2_0-1_2_whiten_toplot.txt};
\addplot[black,thick, mark=square*, mark size=2pt] table [x=d, y=r, col sep=comma] {plot_data/whitening/MSLS_test/MSLS_soft_resnext_GeM_L2_0-1_2_nowhiten_toplot.txt};
\addplot[black, mark=square, thick, mark size=2pt] table [x=d, y=r, col sep=comma] {plot_data/whitening/MSLS_test/MSLS_soft_resnext_GeM_L2_0-1_2_whiten_toplot.txt};
\end{axis}
 \label{fig:whitening_msls_test}
\end{tikzpicture}
\captionof{figure}{MSLS test}
\end{subfigure} \hspace{0.5mm}
\begin{subfigure}{0.24\textwidth}
\begin{tikzpicture}[thick]
\begin{axis}
[xlabel=Dimensions,,grid, xmode=log,log basis x={2}, xtick={32,64,128,256, 512,1024,2048}, xticklabels={32,64,128,256, 512,1024,2048}, width=1.30\textwidth
]
\addplot[teal,thick, mark=triangle*, mark size=2pt] table [x=d, y=r, col sep=comma] {plot_data/whitening/Pitts30k/MSLS_soft_vgg16_GeM_L2_0-1_2_nowhiten_toplot};
\addplot[teal, mark=triangle, thick, mark size=2pt] table [x=d, y=r, col sep=comma] {plot_data/whitening/Pitts30k/MSLS_soft_vgg16_GeM_L2_0-1_2_whiten_toplot};
\addplot[blue,thick, mark=*, mark size=2pt] table [x=d, y=r, col sep=comma] {plot_data/whitening/Pitts30k/MSLS_soft_ResNet50_GeM_L2_0-1_2_nowhiten_toplot};
\addplot[blue, mark=o, thick, mark size=2pt] table [x=d, y=r, col sep=comma] {plot_data/whitening/Pitts30k/MSLS_soft_ResNet50_GeM_L2_0-1_2_whiten_toplot};
\addplot[magenta,thick, mark=diamond*, mark size=2pt] table [x=d, y=r, col sep=comma] {plot_data/whitening/Pitts30k/MSLS_soft_ResNet152_GeM_L2_0-1_2_nowhiten_toplot};
\addplot[magenta, mark=diamond, thick, mark size=2pt] table [x=d, y=r, col sep=comma] {plot_data/whitening/Pitts30k/MSLS_soft_ResNet152_GeM_L2_0-1_2_whiten_toplot};
\addplot[black,thick, mark=square*, mark size=2pt] table [x=d, y=r, col sep=comma] {plot_data/whitening/Pitts30k/MSLS_soft_resnext_GeM_L2_0-1_2_nowhiten_toplot};
\addplot[black, mark=square, thick, mark size=2pt] table [x=d, y=r, col sep=comma] {plot_data/whitening/Pitts30k/MSLS_soft_resnext_GeM_L2_0-1_2_whiten_toplot};
\end{axis}
 \label{fig:whitening_pitts30k}
\end{tikzpicture}
\captionof{figure}{Pitts30k}
\end{subfigure} \hspace{0.5mm}
\begin{subfigure}{0.24\textwidth}
\begin{tikzpicture}[thick]
\begin{axis}
[xlabel=Dimensions, ylabel style={at={(axis description cs:-0.08,0.5)}},grid, xmode=log,log basis x={2}, xtick={32,64,128,256, 512,1024,2048}, xticklabels={32,64,128,256, 512,1024,2048}, width=1.30\textwidth
]
\addplot[teal,thick, mark=triangle*, mark size=2pt] table [x=d, y=r, col sep=comma] {plot_data/whitening/Tokyo247/MSLS_soft_vgg16_GeM_L2_0-1_2_nowhiten_toplot};
\addplot[teal, mark=triangle, thick, mark size=2pt] table [x=d, y=r, col sep=comma] {plot_data/whitening/Tokyo247/MSLS_soft_vgg16_GeM_L2_0-1_2_whiten_toplot};
\addplot[blue,thick, mark=*, mark size=2pt] table [x=d, y=r, col sep=comma] {plot_data/whitening/Tokyo247/MSLS_soft_ResNet50_GeM_L2_0-1_2_nowhiten_toplot};
\addplot[blue, mark=o, thick, mark size=2pt] table [x=d, y=r, col sep=comma] {plot_data/whitening/Tokyo247/MSLS_soft_ResNet50_GeM_L2_0-1_2_whiten_toplot};
\addplot[magenta,thick, mark=diamond*, mark size=2pt] table [x=d, y=r, col sep=comma] {plot_data/whitening/Tokyo247/MSLS_soft_ResNet152_GeM_L2_0-1_2_nowhiten_toplot};
\addplot[black,thick, mark=square*, mark size=2pt] table [x=d, y=r, col sep=comma] {plot_data/whitening/Tokyo247/MSLS_soft_resnext_GeM_L2_0-1_2_nowhiten_toplot};

\addplot[magenta, mark=diamond, thick, mark size=2pt] table [x=d, y=r, col sep=comma] {plot_data/whitening/Tokyo247/MSLS_soft_ResNet152_GeM_L2_0-1_2_whiten_toplot};

\addplot[black, mark=square, thick, mark size=2pt] table [x=d, y=r, col sep=comma] {plot_data/whitening/Tokyo247/MSLS_soft_resnext_GeM_L2_0-1_2_whiten_toplot};
\end{axis}
\label{fig:whitening_tokyo247}
\end{tikzpicture}
\captionof{figure}{Tokyo24/7}
\end{subfigure}
\end{minipage}

\vspace{-2mm}    \caption{Ablation results on the MSLS validation, MSLS test, Pittsburgh30k and Tokyo 24/7 datasets, with different PCA dimensions.}
    \label{fig:whitening}
\end{figure*}

\noindent\textbf{Generalization to other datasets. } In Table~\ref{tab:msls-sota} and Table~\ref{tab:msls-ablation}, we also report the results of generalization to Pittsburgh30k, Tokyo 24/7, RobotCar Seasons v2 and Extended CMU Seasons (plus Pittsburgh250k and TokyoTM in the supplementary materials). The models trained with the GCL function generalize well to unseen datasets, in many cases better than existing methods that retrieve the $k$-nearest neighbours based on  descriptor distance only. 
%
%
%
Our models also generalize well to urban localization datasets like RobotCar Seasons V2 and Extended CMU Seasons, achieving up to 21.9\% and 19\% of correctly localized queries within 0.5m and 5$\degree$, respectively. The results of GCL-based networks are higher than those obtained by NetVLAD, and especially higher than those of TransVPR (with no re-ranking) that uses a transformer as backbone. Note that we do not perform 6DOF pose estimation, but estimate the pose of a query image by inheriting that of the best retrieved match, and thus not compare with methods that perform refined pose estimation. This is inline with the experiments in~\cite{hausler2021patch}. 

Our models are outperformed only by methods that include a re-ranking strategies to refine the list of retrieved images, on the Pittsburgh30k, RobotCar Seasons v2 and Extended CMU Seasons datasets. However, these methods perform extra heavy computations (e.g. up to $6s$ per query in PatchNetVLAD~\cite{hausler2021patch}) to re-rank the list of retrieved images, not focusing on the representation capabilities of the learned descriptors themselves. Thus, we find a direct comparison with these methods not fair. On the contrary, these results demonstrate the fact the VPR descriptors learned used the proposed labels and GCL have better representation capabilities than those produced by other methods, achieving higher results in out-of-distribution experiments as well.


\noindent\textbf{Ablation study: backbone and contrastive loss. } We carried out ablation experiments using four backbones, namely VGG16~\cite{simonyan2015very}, ResNet50, ResNet152~\cite{he2016deep}, and ResNeXt101-32x8d (hereinafter ResNeXt)~\cite{xie2017aggregated}, and the GeM~\cite{radenovic2018fine} global pooling layer, and an additional NetVLAD-GCL model. Extra ablation experiments with an average global pooling layer are included in the supplementary material. For each backbone, we train with the binary Contrastive Loss (CL) and our Generalized Contrastive Loss (GCL). We report the results in Table~\ref{tab:msls-ablation}. The models trained with the GCL consistently outperform their counterpart trained with the CL, showing better generalization to other datasets. Moreover, we demonstrate that a VGG16-GeM architecture outperforms a more complex NetVLAD when trained with our GCL function.  We also perform whitening and PCA on the descriptors, which further boost the performance.

\noindent\textbf{Ablation study: PCA and whitening. } We also study the effect of whitening and PCA dimensionality reduction from 32 to 2048 dimensions. Figure~\ref{fig:whitening} shows the results on the MSLS, Pittsburgh30k and Tokyo 24/7 datasets. In general, the larger the size of the descriptors, the better the results. However, our models maintain comparably high results when the descriptors are whitened and reduced to 256 dimensions, still outperforming the full-size descriptors without whitening. We observed up to a $28.2\%$ improvement in the case of Tokyo 24/7 and $12.8\%$ on the MSLS validation set. When comparing our VGG16-GeM-GCL model reduced to 256 dimensions with NetVLAD ($16\times$ less descriptor size), we still achieve higher results (R@5 of 82.7\% vs 80.8\% on MSLS validation, 60.4\% vs 58.9\% on MSLS test, 87.2\% vs 84.7\% on Pittsburg30k and 62.2\% vs 47.6\% on Tokyo 24/7).
It is to highlight that the contributions of the PCA/whitening and the GCL are complementary, meaning that they can be used together to optimize retrieval performance.


\newcommand{\dashrule}[1][black]{%
  \color{#1}\rule[\dimexpr.5ex-.2pt]{4pt}{.4pt}\xleaders\hbox{\rule{4pt}{0pt}\rule[\dimexpr.5ex-.2pt]{4pt}{.4pt}}\hfill\kern0pt%
}

\begin{table*}[!t]
\renewcommand{\arraystretch}{1.05}
\setlength\tabcolsep{1.5pt}
\centering
\caption{Comparison of localization results (on CMU Seasons) of VGG16 backbones trained with several metric loss functions. Methods in the upper part deploy a NetVLAD pooling layer.}
\label{tab:comparisonwiththoma}
\resizebox{\textwidth}{!}{%
\footnotesize

\begin{tabular} {l@{\hspace{16\tabcolsep}}ccc@{\hspace{20\tabcolsep}}ccc@{\hspace{14\tabcolsep}}ccc@{\hspace{14\tabcolsep}}ccc}
\toprule
 & \multicolumn{3}{c}{\textbf{All}} & \multicolumn{3}{c}{\textbf{Urban}} & \multicolumn{3}{c}{\textbf{Suburban}} & \multicolumn{3}{c}{\textbf{Park}} \\ 
\textbf{Loss function} & \textbf{0.25m/2$\degree$} & \textbf{0.5m/5$\degree$} & \textbf{5m/10$\degree$} & \textbf{0.25m/2$\degree$} & \textbf{0.5m/5$\degree$} & \textbf{5m/10$\degree$} & \textbf{0.25m/2$\degree$} & \textbf{0.5m/5$\degree$} & \textbf{5m/10$\degree$} & \textbf{0.25m/2$\degree$} & \textbf{0.5m/5$\degree$} & \textbf{5m/10$\degree$} \\\midrule
Triplet (original NetVLAD)~\cite{Arandjelovic2017} & 6.0 & 15.5 & 59.9  &  9.4 & 22.6 & 71.2 & 3.9 & 11.8 & 60.1 & 3.2 & 9.2 & 45.2    \\
Quadruplet~\cite{Chen2017beyondTriplet} & 6.9 & 17.5 & 62.3  & 10.7 & 25.2 & 73.3 & 4.4 & 13.0 & 61.4 & 3.9 & 10.8 & 47.9    \\
Lazy triplet~\cite{angelina2018pointnetvlad} & 6.4 & 16.5 & 58.6  &  9.9 & 23.5 & 69.8 & 4.1 & 11.9 & 58.2 & 3.5 & 10.1 & 42.0  \\
Lazy quadruplet~\cite{angelina2018pointnetvlad} & 7.3 & 18.5 & 61.7  & 11.4 & 26.9 & 72.7 & 4.9 & 13.9 & 64.1 & 3.7 & 10.7 & 44.1  \\
Triplet + Huber distance~\cite{Thoma2020Mappable} & 6.0 & 15.3 & 55.9 & 9.5 & 22.9 & 69.0 & 4.4 & 12.4 & 57.3 & 3.0 & 8.4 & 39.6   \\
Log-ratio~\cite{kim2019deep} & 6.7 & 17.4 & 58.8  & 10.5 & 24.9 & 71.4 & 4.6 & 13.4 & 57.4 & 3.5 & 10.2 & 42.8  \\
Multi-similarity~\cite{Wang2019multisimilarity} & 7.4 & 18.8 & 66.3 & 12.0 & 28.8 & 81.6 & 5.1 & 14.6 & 63.9 & 3.8 & 10.9 & 52.7 \\
Soft contrastive~\cite{thoma2020soft} & 8.0 & 20.5 & \bfseries 70.4 & 12.7 & 30.7 & \bfseries 84.6 & 5.1 & 14.9 & \bfseries 67.9 & 4.5 & 12.6 & \bfseries 56.8 \\ \midrule
\bfseries GCL (Ours) & \bfseries 9.2 & \bfseries 22.8 & 65.8 & \bfseries 14.7 & \bfseries 34.2 & 82.6 & \bfseries 5.7 & \bfseries 16.1 & 64.6 & \bfseries 5.1 & \bfseries 14.0 & 49.1 \\ \tabucline[0.4pt black!40 off 2pt]{~}
\textit{GCL (Ours w/ ResNeXt)} & \textit{9.9} & \textit{24.3} & \textit{75.5} & \textit{15.4} & \textit{36.0} & \textit{89.6} & \textit{6.7} & \textit{18.4} & \textit{76.8} & \textit{5.6} & \textit{14.9} & \textit{60.3} \\ 
\bottomrule
\end{tabular}%
}
\end{table*}


\noindent\textbf{Comparison with other loss functions. } We compared with other loss functions used for VPR, by following up on the experiments in~\cite{thoma2020soft}. 
The loss functions included in the comparison in~\cite{thoma2020soft} are the triplet loss~\cite{Arandjelovic2017} (also with Huber distance~\cite{Thoma2020Mappable}), quadruplet loss~\cite{Chen2017beyondTriplet}, lazy triplet and lazy quadruplet loss~\cite{angelina2018pointnetvlad}, plus functions that embed mechanisms to circumvent the use of binary labels, namely a multi-similarity loss~\cite{Wang2019multisimilarity}, log-ratio~\cite{kim2019deep} and soft contrastive loss~\cite{thoma2020soft}. 
The results reported in Table~\ref{tab:comparisonwiththoma} show that the GCL achieves higher localization accuracy especially when stricter thresholds for distance and angle are set. These results indicate that the GCL descriptors are better effective in neighbor search and retrieval, and their ranking based on distance from the query descriptor is a more reliable measure of visual place similarity. 
All methods in the upper part of the table deploy a VGG16 backbone with a NetVLAD pooling layer and make use of hard-negative pair mining. We also use a VGG16 backbone and do not perform hard-negative pair mining, substantially reducing the training time and memory requirements. This allows us to also train backbones with larger capacity, e.g. ResNeXt, on a single V100 GPU in less than a day, of which we report the results in italics for completeness.

\noindent\textbf{Processing time. } The GCL function and the graded similarity labels contribute to training effective models in a data- and computation-efficient way, largely improving on the resources and time required to train NetVLAD (see Table~\ref{tab:training-time}). Our VGG16-GeM-GCL model obtains higher results than NetVLAD while requiring $6\times$ less memory and about $100\times$ less time to converge. We point out that NetVLAD is the backbone of several other methods for VPR such as PatchNetVLAD, DELG and SuperGlue in Table~\ref{tab:msls-sota}, thus making the comparison in Table~\ref{tab:training-time} relevant from a larger perspective.
The graded similarity and GCL function contribute to an efficient use of training data. A single epoch, i.e. a model sees a certain training pair only once, is sufficient for the training of GCL-based models to converge. The low memory and time requirements also enable the training of models with larger backbones, that obtain very high results while still keeping the resource usage low.
We point out the data-efficient training that we deployed can stimulate further and faster progress in VPR, as it enables to train larger backbones, and perform extensive hyperparameter optimization or more detailed ablation studies.


\begin{table}[t!]
\centering
\revised{\caption{Training time and GPU memory utilization for a batch size of 4 images. In the epochs column, the number in parenthesis is the number of epochs until convergence. }
\label{tab:training-time}
\footnotesize
\begin{tabular}{@{}lcccc@{}}
\toprule
\textbf{Model} &\textbf{memory} & \textbf{epochs} & \textbf{t/epoch} & \textbf{t/converge} \\ \midrule
NetVLAD-16-TL & 9.67GB & 30 (22) & 24h & 22d \\
NetVLAD-64-TL & - &10 (7) & 36h & 10.5d \\
VGG16-GeM-GCL & 1.49GB & 1 & 5h & 5h \\
ResNet50-GeM-GCL & 1.65 GB & 1 & 6h & 6h \\
ResNet152-GeM-GCL & 3.78 GB & 1 & 14h & 14h \\
ResNetXt-GeM-GCL & 4.77 GB& 1 (1/2) & 28h & 14h \\ \bottomrule
\end{tabular}%
}
\end{table}




\section{Conclusions}
\label{sec:conclusions}
We extended the learning of image descriptors for visual place recognition by using measures of camera pose similarity and 3D surface overlap as proxies for graded image pair similarity to re-annotate existing VPR datasets (i.e. MSLS, 7Scenes and TB-Places). We demonstrated that the new labels can be used to effectively compose training batches without the need of hard-pair mining, decisively speeding-up training time while reducing memory requirements.
Furthermore, we reformulated the Contrastive Loss function, proposing a Generalized Contrastive Loss (GCL). The GCL exploits the graded similarity of image pairs, and contributes to learning way better performing image descriptors for VPR than those of other losses that are not designed to use graded image similarity labels (i.e. triplet, quadruplet and their variants) and that require hard-pair mining during training.  
Models trained with the GCL and new graded similarity labels obtain comparable or higher results that several existing VPR methods, including those that apply re-ranking of the retrieved images, while keeping a more efficient use of the data, training time and memory. We achieved good generalization to unseen environments, showing robustness to domain shifts on the Pittsburgh30k, Tokyo 24/7, RobotCar Seasons v2 and Extended CMU Seasons datasets.
The combination of graded similarity annotations and a loss function that can embed them in the training
paves a way to learn more effective descriptors for VPR in a data- and resource-efficient manner.

{\small
\bibliographystyle{ieee_fullname}
\bibliography{egbib}
}

\newpage
\appendix

\appendix

\section{Implementation and training details.}\label{sec:implementation}
We carried out experiments using PyTorch~\cite{paszke2017automatic} and a single Nvidia V100 GPU. For all considered backbones (pre-trained on ImageNet), we trained the two last convolutional blocks for one epoch only on the MSLS train set ($\sim$500k pairs), making an efficient use of training data. 
We use the SGD optimizer with initial learning rate of 0.1 for the runs with the GCL function and 0.01 for those with the CL function, and decrease it by a factor of $10^{-1}$ after 250k pairs. For the TB.Places and 7Scene experiments, since they are much smaller datasets we train for 30 epochs, reaching convergence around epoch 15 for both datasets. The initial learning rate is the same as we used for the MSLS experiments, and we decrease it by a factor of $10^{-1}$ after every 5 epochs. All the descriptors are L2-normalized.


\section{Relabeling with image similarity proxies}
\paragraph{2D Field of View Overlap: MSLS data set}

\begin{figure}[!t]
    \centering
\begin{subfigure}{.22\textwidth}
    \includegraphics[width=\textwidth]{figures/4_data/london_query_75_5.jpg}
    \caption{Query}
    \label{fig:msls_london_query}
    \end{subfigure}
~\begin{subfigure}{.22\textwidth}
    \includegraphics[width=\textwidth]{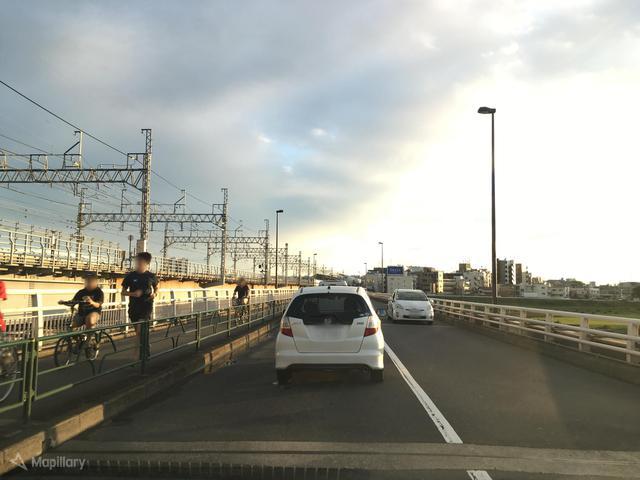}
    \caption{Query}
    \label{fig:msls_tokyo_query}
    \end{subfigure}
~\begin{subfigure}{.22\textwidth}
    \includegraphics[width=\textwidth]{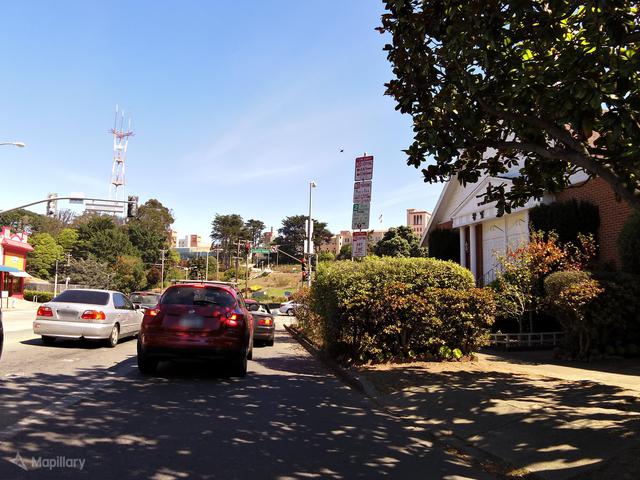}
    \caption{Query}
    \label{fig:msls_sf_query}
    \end{subfigure}
~\begin{subfigure}{.22\textwidth}
    \includegraphics[width=\textwidth]{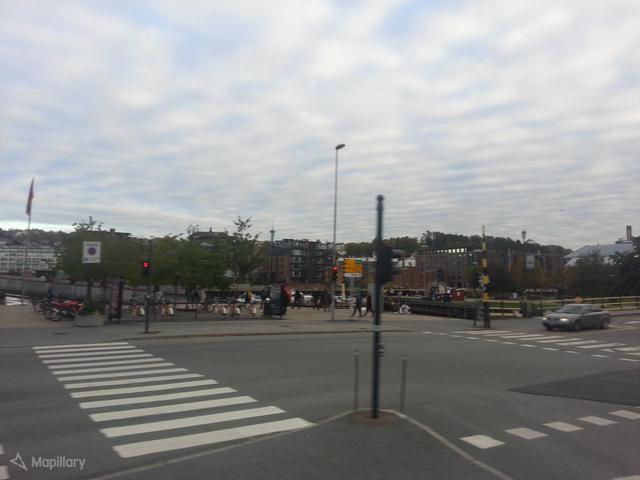}
    \caption{Query}
    \label{fig:msls_trondheim_query}
    \end{subfigure}

\begin{subfigure}{.22\textwidth}
    \includegraphics[width=\textwidth]{figures/4_data/london_db_75_5.jpg}
    \caption{Positive match}
    \label{fig:msls_london_db}
    \end{subfigure}
~\begin{subfigure}{.22\textwidth}
    \includegraphics[width=\textwidth]{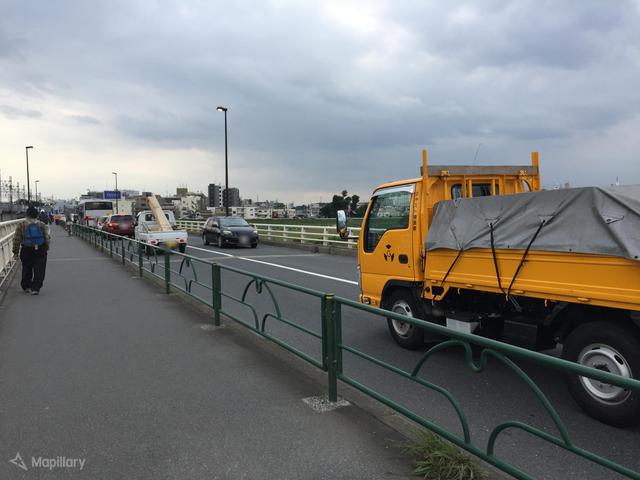}
    \caption{Positive match}
    \label{fig:msls_tokyo_db}
    \end{subfigure}
~\begin{subfigure}{.22\textwidth}
    \includegraphics[width=\textwidth]{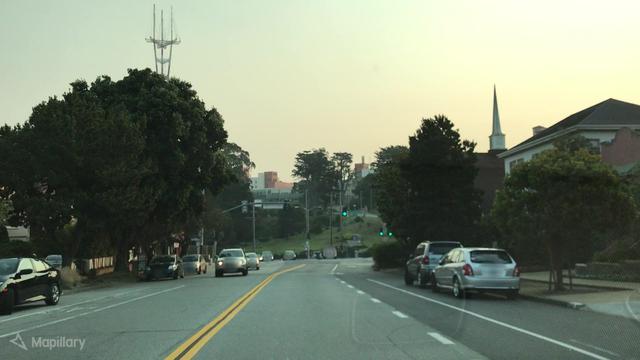}
    \caption{Soft negative match}
    \label{fig:msls_sf_db}
    \end{subfigure}
~\begin{subfigure}{.22\textwidth}
    \includegraphics[width=\textwidth]{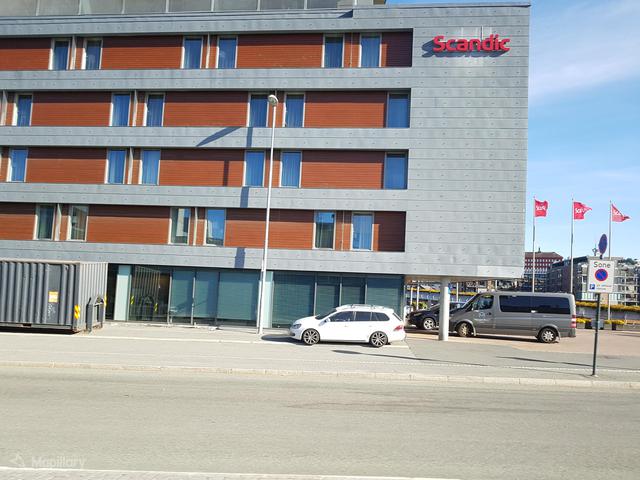}
    \caption{Hard negative match}
    \label{fig:msls_trondheim_db}
    \end{subfigure}

\begin{subfigure}{.22\textwidth}
    \includegraphics[width=\textwidth]{figures/4_data/london_fov_75_5.png}
    \caption{75.5\% FoV overlap}
    \label{fig:msls_london_fov}
    \end{subfigure}
~\begin{subfigure}{.22\textwidth}
    \includegraphics[width=\textwidth]{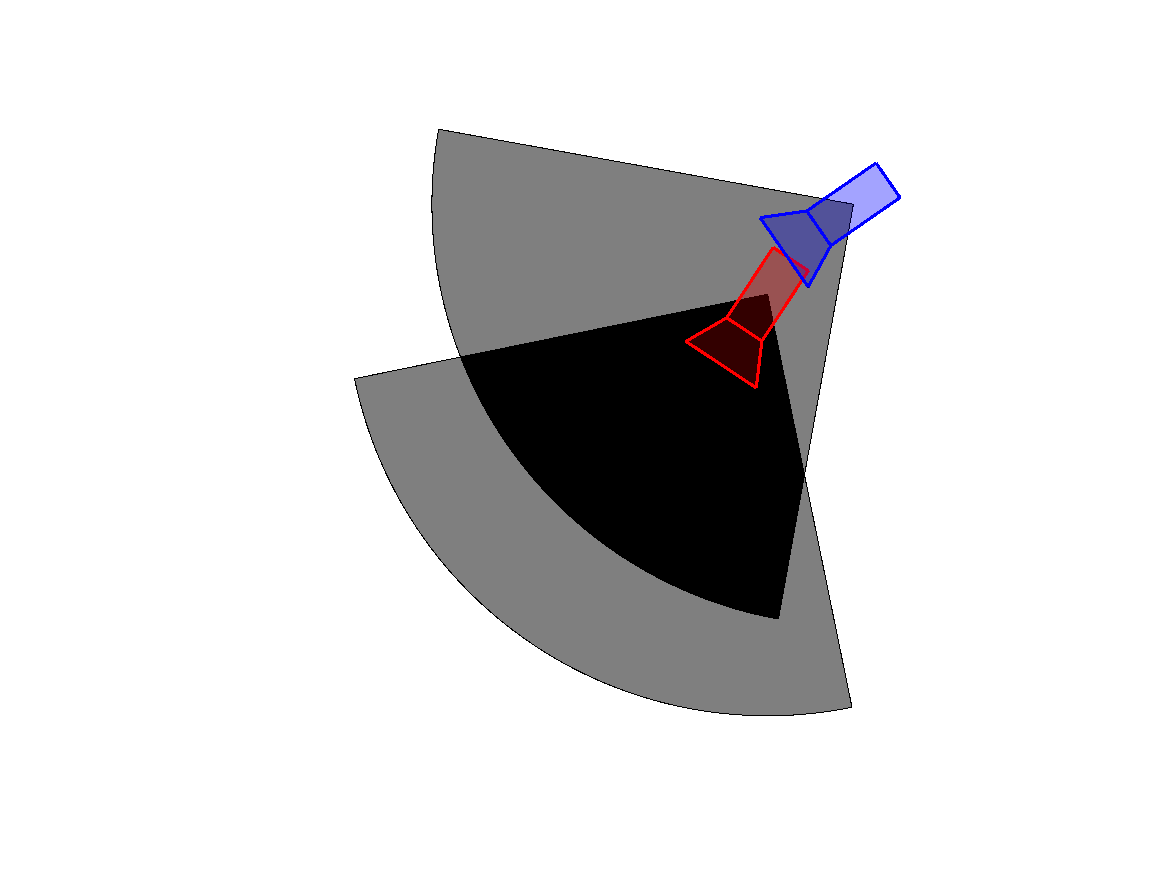}
    \caption{50.31\% FoV overlap}
    \label{fig:msls_tokyo_fov}
    \end{subfigure}
~\begin{subfigure}{.22\textwidth}
    \includegraphics[width=\textwidth]{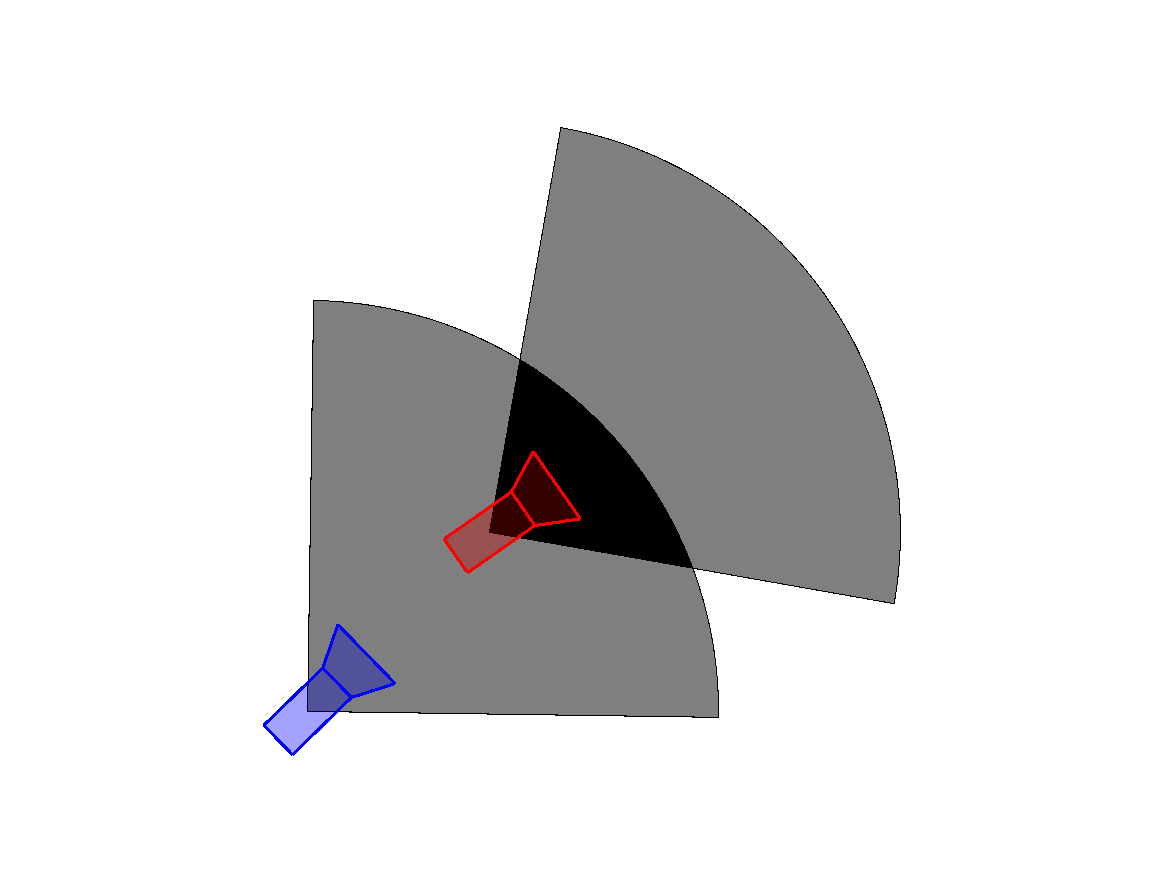}
    \caption{16.78\% FoV overlap}
    \label{fig:msls_sf_fov}
    \end{subfigure}
~\begin{subfigure}{.22\textwidth}
    \includegraphics[width=\textwidth, trim=1cm 1cm 1cm 1cm, clip]{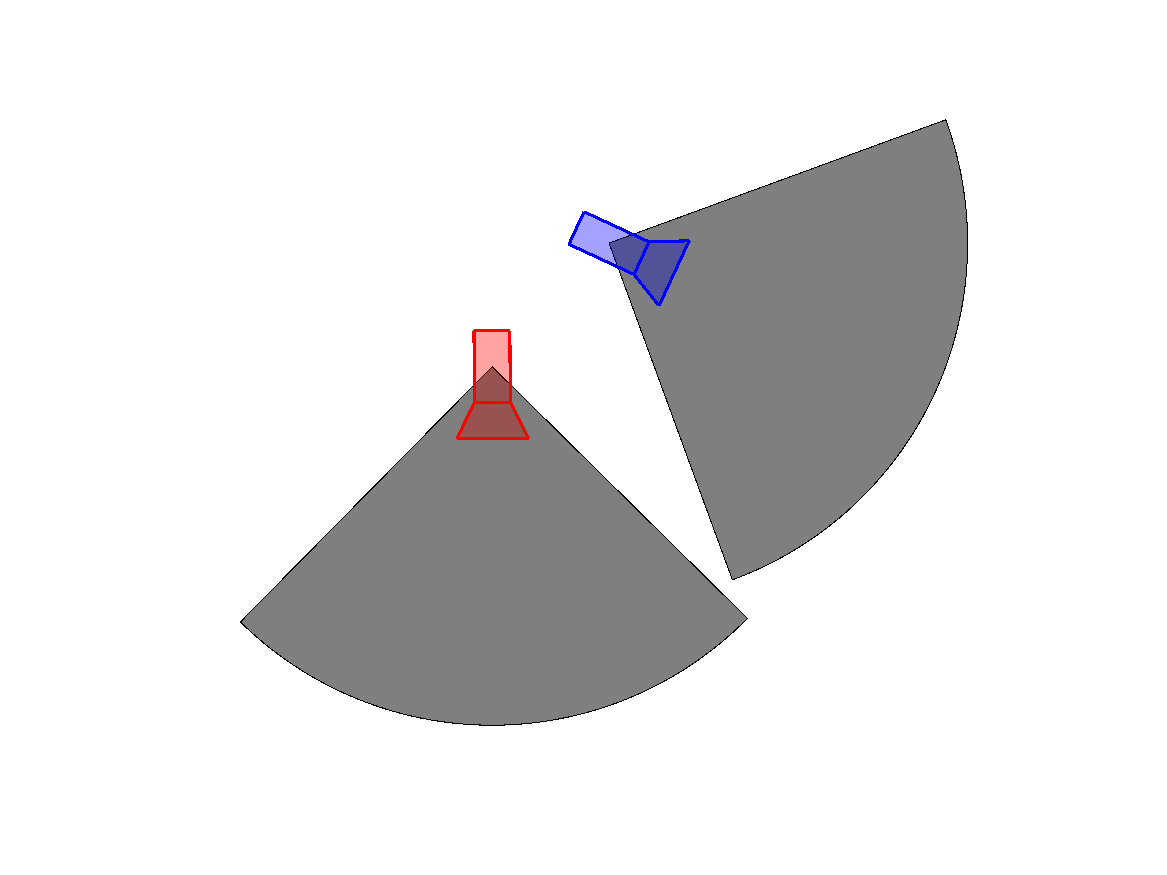}
    \caption{0\% FoV overlap}
    \label{fig:msls_trondheim_fov}
    \end{subfigure}

    \caption{Example image pairs from the MSLS dataset. The first row shows the query images, the second row shows the corresponding matches from the map set, and the third row shows the estimated 2D FoV overlap. The query image is associated with the red camera, while the map image with the blue camera. The first column shows a positive match with 75.5\% FoV overlap, and many visual features in common. The second column shows a borderline pair: the two images have 50.31\% FoV overlap and some features in common. The third column shows a soft negative match, where the two images have FoV overlap of 16.78\%. The fourth column shows a hard negative match, where the two images are taken by cameras looking in opposite directions, and the FoV overlap is 0\%.}
    \label{fig:msls_fov}
\end{figure}
The authors of MSLS defined a positive match when the retrieved map image falls within 25m and $40^{\circ}$ from the query. We define a similarity measure that satisfies those constraints. Image pairs taken at locations that are closer than 25m and with orientation differences lower than $40^{\circ}$ are expected to have a similarity higher than $50\%$, to be considered similar. Moreover, the borderline cases with distances close to 25m and/or orientation difference near to $40^{\circ}$ should have a similarity close to $50\%$. Hence, we define $r=25m\times2=50m$ and estimate a $\theta$ that gives approximately a $50\%$ FoV overlap for the borderline cases,  i.e. 0m@$40^{\circ}$, and 25m@$0^{\circ}$. For the former, the optimal $\theta$ corresponds to $80^{\circ}$, and for the second latter to $102^{\circ}$. We settle for a value in the middle and define $\theta=90^{\circ}$, which gives $55.63\%$ and $45.01\%$ FoV overlaps in the borderline cases. 
We display  examples of the similarity ground truth for MSLS in Fig.~\ref{fig:msls_fov}. We compute the similarity ground truth for each possible query-map pair, per city. The new annotations are publicly available.

\paragraph{2D Field of View Overlap with IMU and laser tracking: TB-Places dataset}

We present another computation of the 2D field of View overlap to estimate the similarity of image pairs when 6DOF camera pose information is available as metadata next to the images in the dataset. The translation vector $(t_0,t_1)$ and orientation angle $\alpha$ are extracted from the pose vector.

\begin{figure}
  \centering
  \begin{subfigure}{0.3\columnwidth}
\includegraphics[width=\columnwidth]{figures/4_data/W18_positive_q.png}
    \caption{Query}
    \label{fig:tbplaces_positive_query}
  \end{subfigure}
    \begin{subfigure}{0.3\columnwidth}
    \includegraphics[width=\columnwidth]{figures/4_data/W18_positive_db.png}
    \caption{Map}
    \label{fig:tbplaces_positive_db}
  \end{subfigure}
      \begin{subfigure}{0.3\columnwidth}
    \includegraphics[width=\columnwidth, trim=1cm 2cm 1cm 2cm, clip]{figures/4_data/w18_positive_74_21.png}
    \caption{FoV overlap (74\%)}
    \label{fig:tbplaces_positive_fov}
  \end{subfigure}
  \begin{subfigure}{0.3\columnwidth}
  \includegraphics[width=\columnwidth]{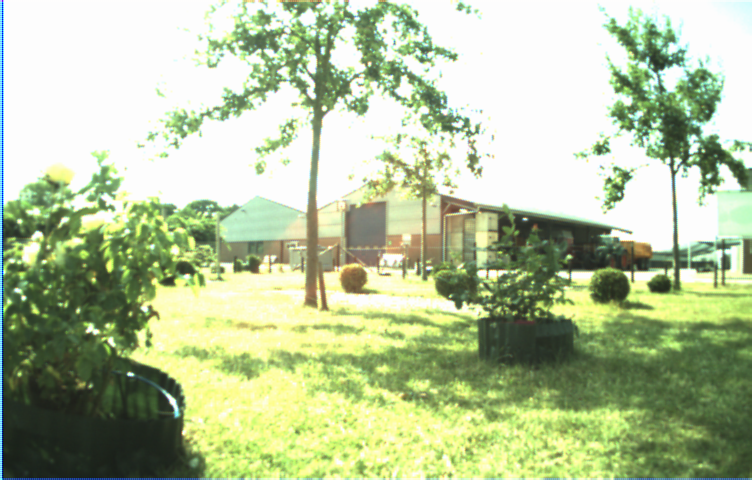}
      \caption{Query}
    \label{fig:tbplaces_negative_query}
  \end{subfigure}
    \begin{subfigure}{0.3\columnwidth}
\includegraphics[width=\columnwidth]{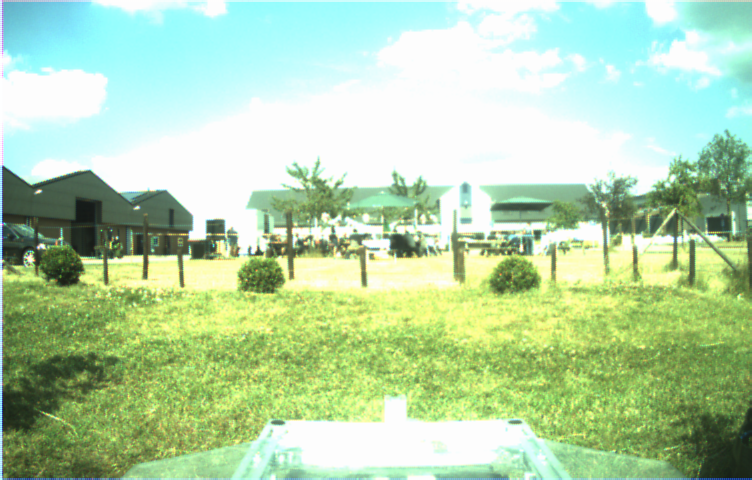}
\caption{Map}
    \label{fig:tbplaces_negative_db}
  \end{subfigure}
  \begin{subfigure}{0.3\columnwidth}
  \includegraphics[width=\columnwidth, trim=1cm 2cm 1cm 2cm, clip]{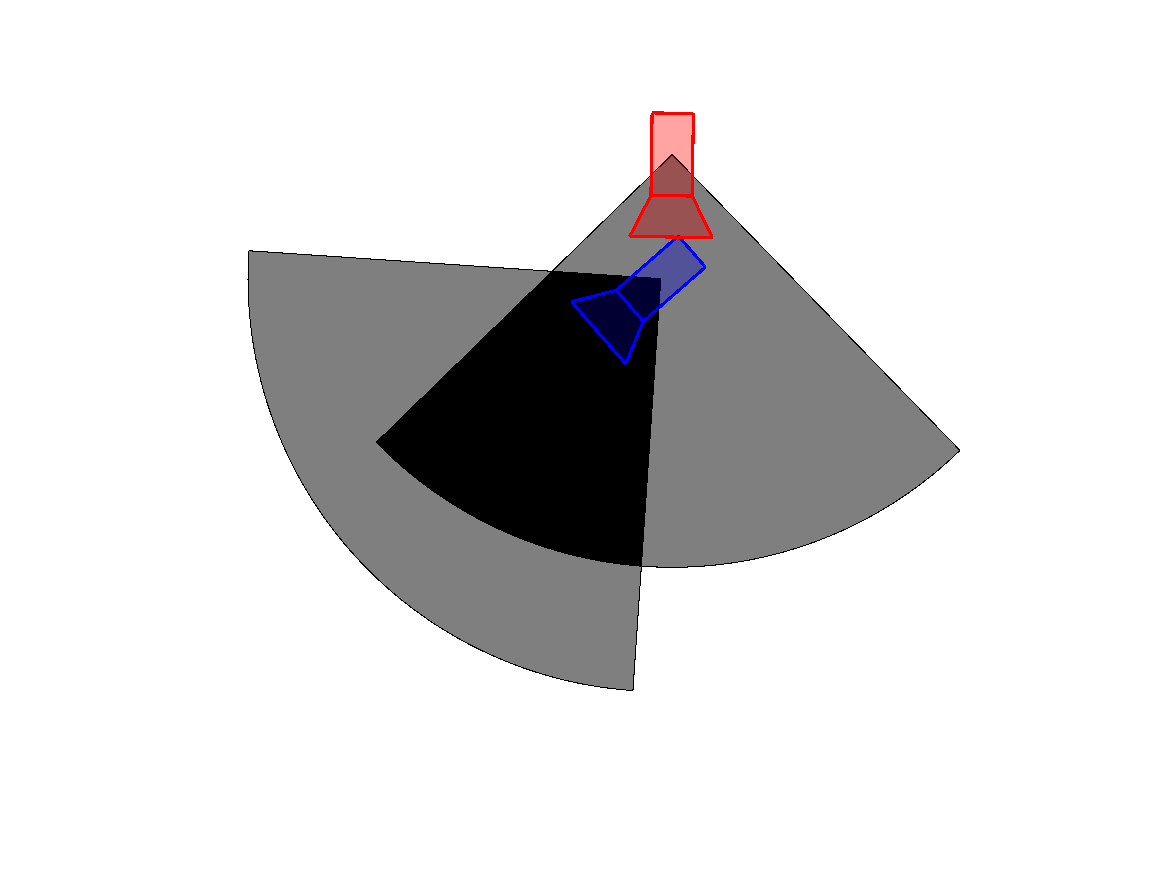}
  \caption{FoV overlap (41\%)}
    \label{fig:tbplaces_negative_fov}
  \end{subfigure}
    \caption{Example image pairs from the TB-Places dataset. The first row shows a positive pair with FoV overlap of 74\%. The second row depicts a soft negative pair with FoV overlap equal to 41\%. The red camera corresponds to the query image, while the blue one to the map image.}
    \label{fig:tbplaces_fov}
\end{figure}

This is the case of the TB-Places dataset~\cite{leyvavallina2019access}, for visual place recognition in gardens, created for the Trimbot2020 project~\cite{strisciuglio2018trimbot2020}. It contains images taken in an experimental garden over three years and includes variations in illumination, season and viewpoint. Each image comes with a 6DOF camera pose, obtained with an IMU and laser tracking, which allows us to estimate a very precise 2D FoV. According to the original paper of the TB-places dataset, we set the FoV angle of the cameras as $\theta = 90^{\circ}$ and the radius as $r=3.5m$. We thus estimate the 2D field of view overlap and use it to re-label the pairs of images contained in the dataset. We provide the similarity ground truth for all possible pairs within W17, the training set. We show some examples of image pairs and their 2D FoV overlap in Fig.~\ref{fig:tbplaces_fov}.

\paragraph{3D Surface Overlap: 7Scenes dataset}~\label{sec:3Dfov}

When the 3D reconstruction of a concerned environment is available, we estimate the degree of similarity of image pairs by computing the \emph{3D Surface overlap}. We project a given image with an associated 6DOF camera pose onto the reconstructed pointcloud of the environment. 
We select the subset of 3D points that falls within the boundaries of the image as the image 3D Surface. For an image pair, we compute their 3D Surface overlap as the intersection-over-union (IoU) of the sets of 3D points associated with the two images. \revised{This computation is similar to the \emph{maximum inliers} measure proposed in~\cite{radenovic2018fine}, where it was used as part of a pair-mining strategy that also involved the computation of a distance in the latent space.} We consider, instead, the computed 3D Surface overlap as a proxy for the degree of similarity of a pair of images. 

\begin{figure}

    \centering
    \begin{subfigure}{0.26\columnwidth}
    \includegraphics[width=\columnwidth]{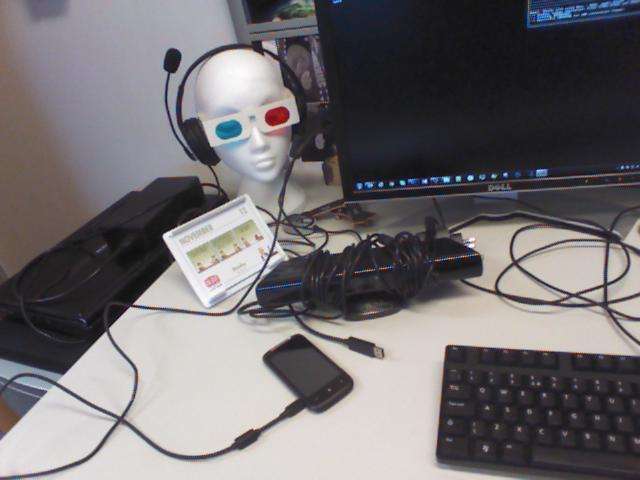}
    \caption{Query}
    \label{fig:heads_query}
    \end{subfigure}
    \begin{subfigure}{0.26\columnwidth}
    \includegraphics[width=\columnwidth]{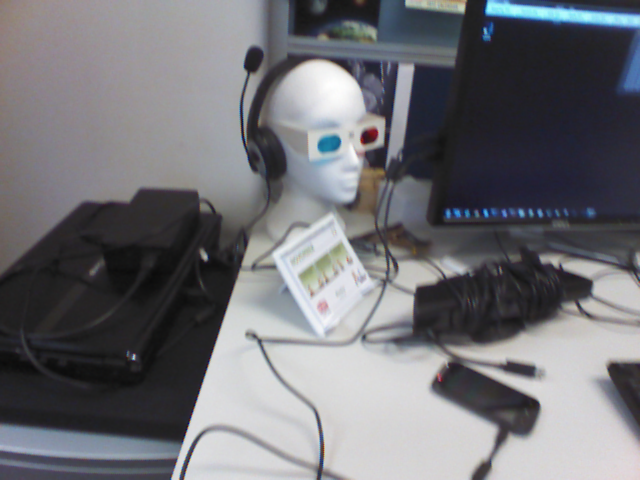}
    \caption{Map}
    \label{fig:heads_db}
    \end{subfigure}
    \begin{subfigure}{0.3\columnwidth}
    \includegraphics[width=\columnwidth,trim=0 0 0 7cm, clip]{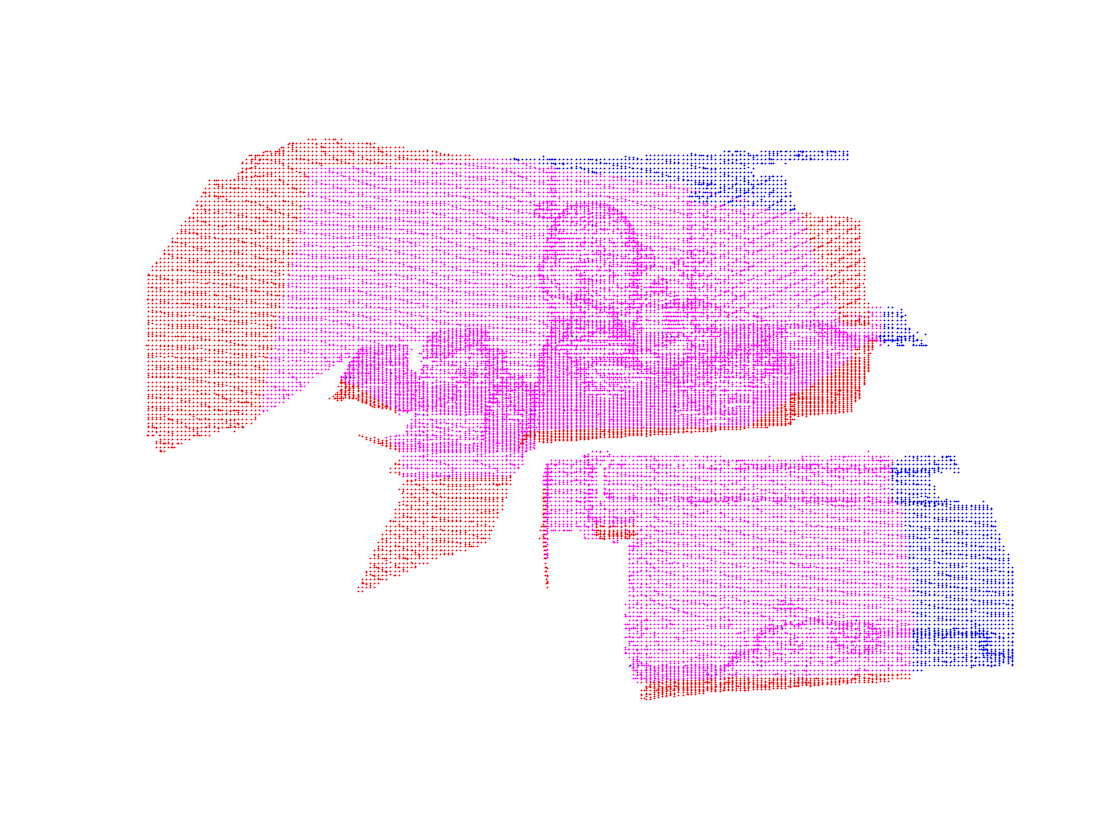}
    \caption{3D Surface overlap (75\%)}
    \label{fig:heads_fov}
    \end{subfigure}
    
    \begin{subfigure}{0.26\columnwidth}
    \includegraphics[width=\columnwidth]{figures/4_data/example_fov_pumpkin_query.png}
    \caption{Query}
    \label{fig:pumpkin_query} 
    \end{subfigure}
    \begin{subfigure}{0.26\columnwidth}
\includegraphics[width=\columnwidth]{figures/4_data/example_fov_pumpkin_map.png}
    \caption{Map}
    \label{fig:pumpkin_db}\end{subfigure}
    \begin{subfigure}{0.32\columnwidth}
    \includegraphics[width=\columnwidth]{figures/4_data/example_fov_pumpkin_noaxis.png}
    \caption{3D Surface overlap (50\%)}
    \label{fig:pumpkin_fov}
    \end{subfigure}
    
    \begin{subfigure}{0.26\columnwidth}
    \includegraphics[width=\columnwidth]{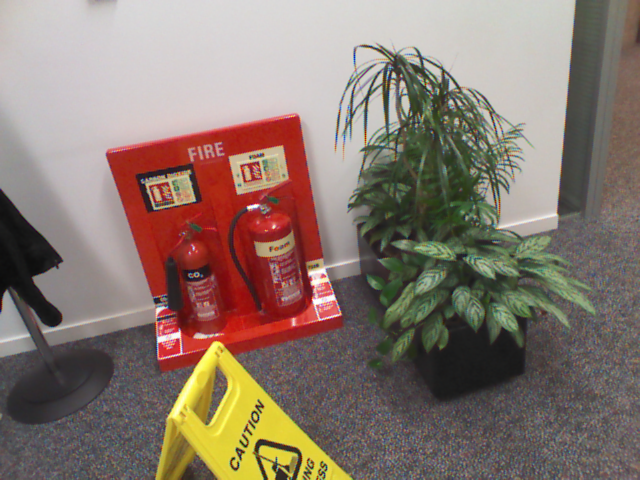}
   \caption{Query}
    \label{fig:fire_query} 
    \end{subfigure}
    \begin{subfigure}{0.26\columnwidth}
    \includegraphics[width=\columnwidth]{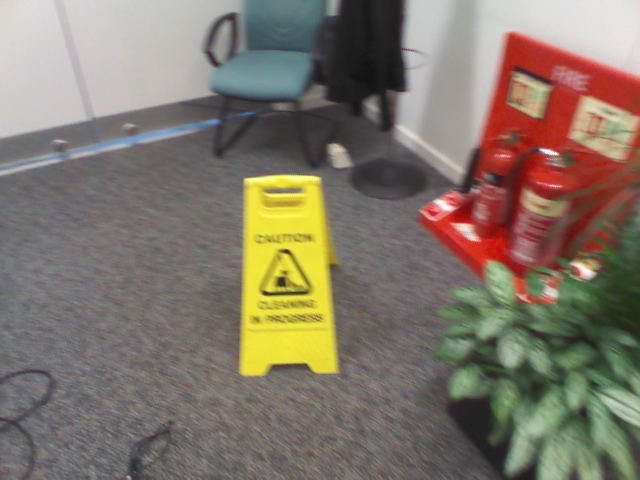}
   \caption{Map}
    \label{fig:fire_db} 
    \end{subfigure}
    \begin{subfigure}{0.3\columnwidth}
    \includegraphics[width=\columnwidth]{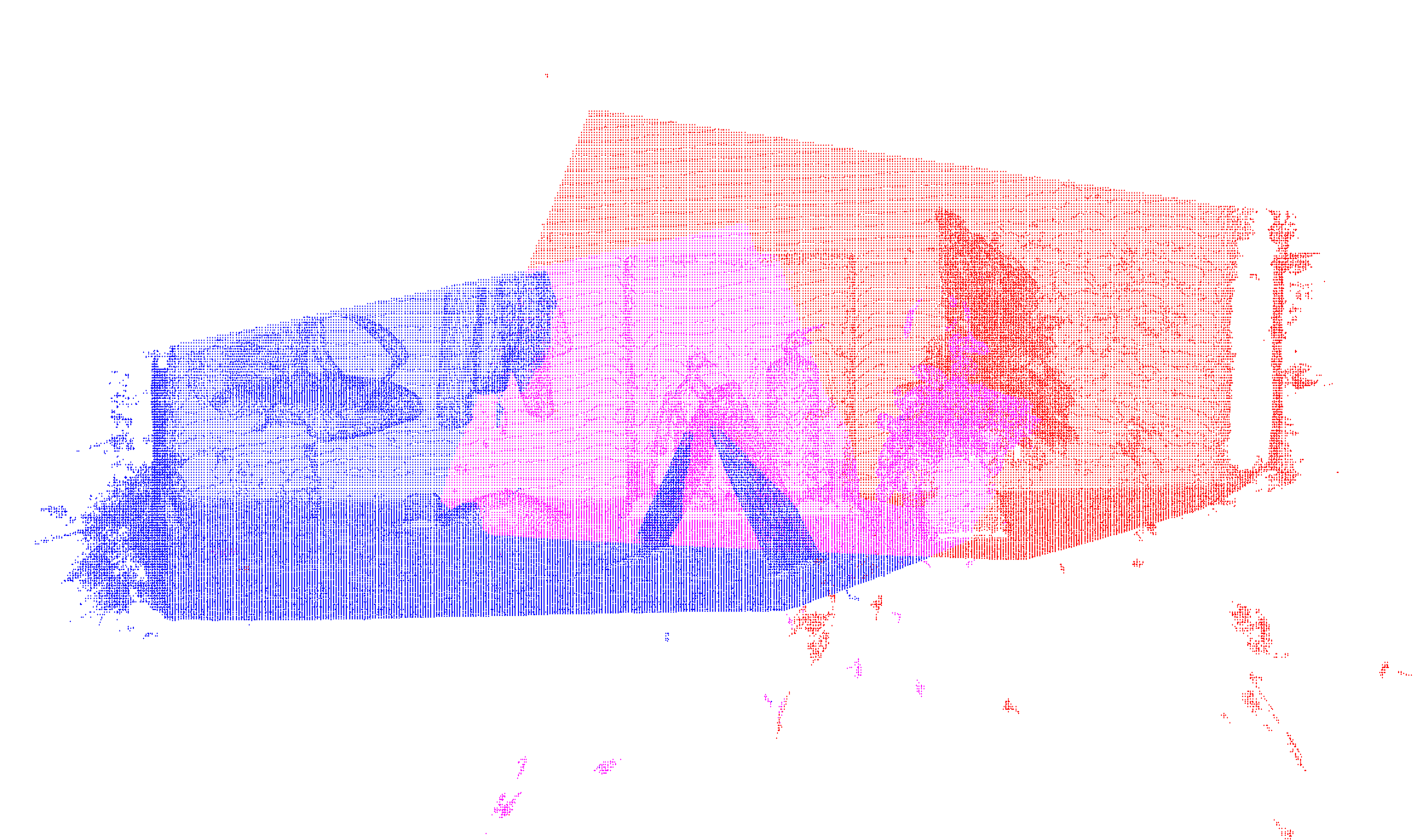}
    \caption{3D Surface overlap (25\%)}
    \label{fig:fire_fov}
    \end{subfigure}
    \caption{3D Surface overlap examples from the 7Scenes dataset. The first row shows a positive pair with 3D Surface overlap of 75\%. The second row depicts a borderline pair with 50\% 3D Surface overlap. The last row shows a soft negative image pair with 25\% 3D Surface overlap.
    The red pointcloud corresponds to the query 3D Surface, the blue one is the map, and the magenta represents the overlap between them.}
    \label{fig:7scenes_fov}
\end{figure}

We use the \emph{3D Surface overlap} to re-annotate the 7Scenes dataset~\cite{Shotton2013}, an indoor localization benchmark, that contains RGBD images taken in seven environments. Each image has an associated 6DOF pose, and a 3D reconstruction of each scene is available. We provide annotations for each possible pair within the training set per scene. 
We show some examples of the 3D Surface overlap in Fig.~\ref{fig:7scenes_fov}. We display the 3D Surface associated to the query image in red, that of the map image in blue and their overlap in magenta. 

\subsection{Field of View vs. Distance}

\begin{figure}[!t]
    \begin{subfigure}{.45\columnwidth}
    \centering
    \hspace{-15mm}\includegraphics[width=\columnwidth]{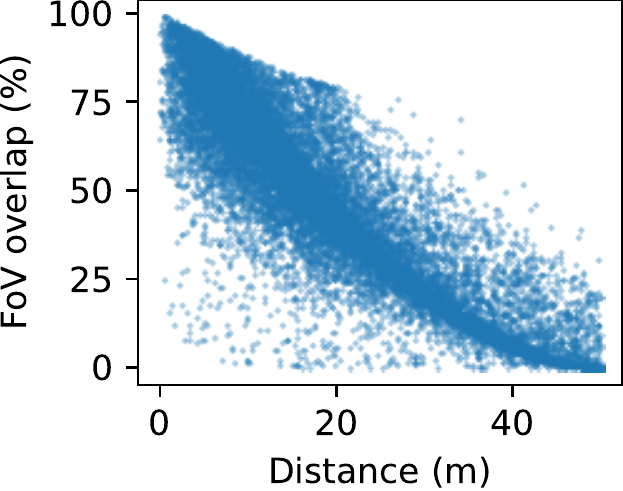}
    \caption{}
    \label{fig:supp_fov_tr}
    \end{subfigure}
    ~\begin{subfigure}{.45\columnwidth}
\centering
    \includegraphics[width=\columnwidth]{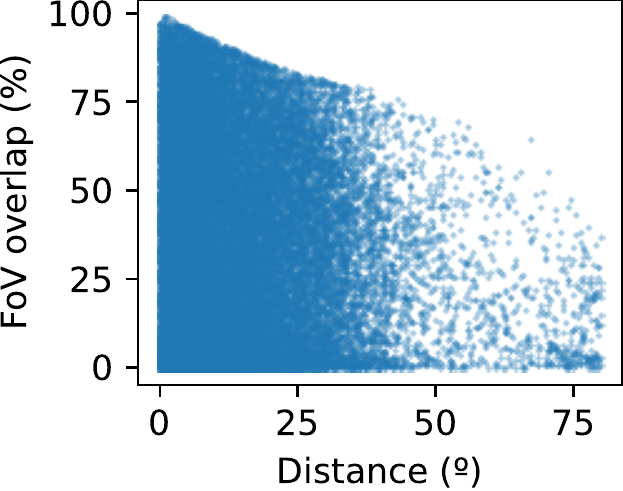}
    \caption{}
        \label{fig:supp_fov_rot}
        \end{subfigure}\\
         \begin{subfigure}{.45\columnwidth}
\centering
    \includegraphics[width=\columnwidth]{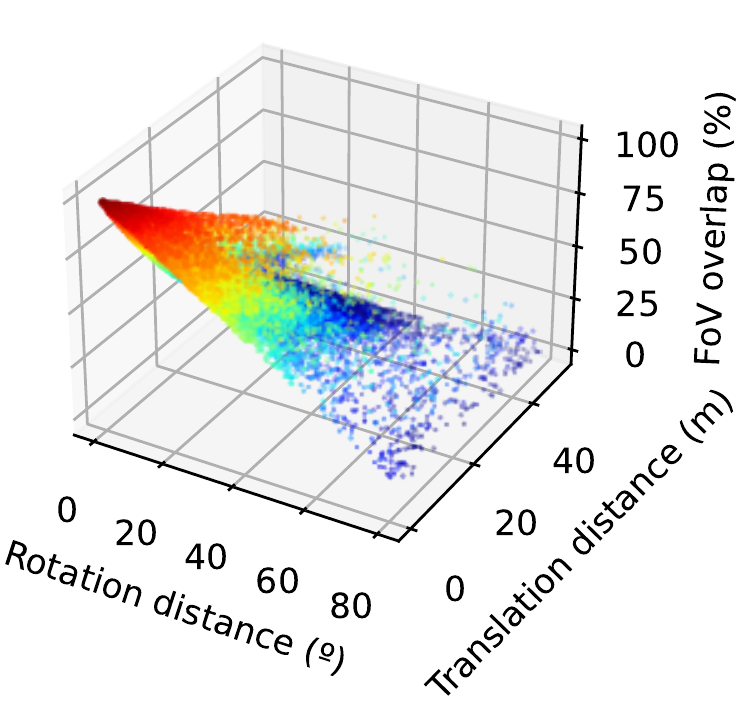}
    \caption{}
        \label{fig:supp_fov_3d}
        \end{subfigure}~         \begin{subfigure}{.45\columnwidth}
\centering
    \includegraphics[width=\columnwidth]{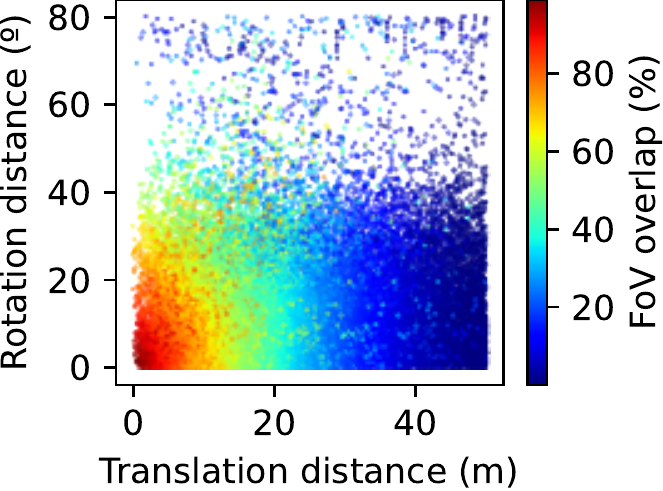}
    \caption{}
    \label{fig:supp_fov_colormap}
    \end{subfigure}
    \caption{Relation of 2D FoV overlap with translation and orientation distance (computed using a subset of MSLS validation set).} 
    \label{fig:supp_fov}
\end{figure}

We studied the relation among translation distance, rotation difference and the resulting FoV overlap measure. 
For that, we selected a subset of pairs of the MSLS training set and measured how the value of the FoV overlap varies with respect to the position and orientation difference of the image pairs. 
In Fig.~\ref{fig:supp_fov_tr}, we plot the relationship between the  translation distance and the value of the FoV overlap. We observed a somewhat linear relationship between increasing translation distance and decreasing FoV overlap. Moreover, many image pairs with FoV overlap of approximately 50\% were taken at a distance of about 25m. The variance observed in Fig.~\ref{fig:supp_fov_tr} is attributable to the orientation changes. Indeed, the relation between FoV overlap and orientation difference is less clear (see Fig.~\ref{fig:supp_fov_rot}). However, the pairs with smaller orientation distance tend to have a generally higher FoV overlap. 
We plot the relation of the FoV overlap with both translation and orientation distance in a 3-dimensional plane in Fig.~\ref{fig:supp_fov_3d}, and as a heatmap in Fig.~\ref{fig:supp_fov_colormap}. 
Rotation and translation distance jointly influence the FoV overlap: the smaller the translation and orientation distance, the higher the similarity. Furthermore, the higher one of the distance measures, the lower the computed FoV overlap, and thus the annotated ground truth similarity.

\section{CL vs GCL: comparison of  latent space }

\begin{figure}[!t]
  \centering
  \begin{subfigure}{0.48\columnwidth}
  \includegraphics[width=\columnwidth]{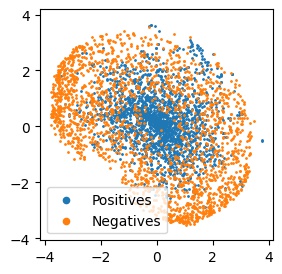}
  \caption{ResNet50-GeM-CL}
    \label{fig:tsneCL}
  \end{subfigure}
  \hfill
  \begin{subfigure}{0.48\columnwidth}
\includegraphics[width=\columnwidth]{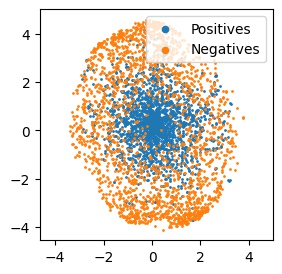}
\caption{ResNet50-GeM-GCL}
\label{fig:tsneGCL}
  \end{subfigure}
    \caption{Visualization of the learned latent space. We selected 1000 random positive pairs and 1000 random negative pairs from the MSLS Copenhagen set, computed the differences between their representations and projected them into a 2D space using t-SNE.}
    \label{fig:tsne}
\end{figure}

In Fig.~\ref{fig:tsne}, we show the 2D projection of the difference of the latent space representations of 2000 image pairs (1000 positive and 1000 negative) randomly selected from the Copenhagen set of the MSLS validation set. For each pair, we compute the difference between the map and the query image descriptors. We use this as input to t-SNE~\cite{tsne}, which projects the descriptors onto a 2D space. We visualize the vectors produced by two models with a ResNet50-GeM backbone, one trained using the CL function (Fig.~\ref{fig:tsneCL}) and the other using the GCL (Fig.~\ref{fig:tsneGCL}) function. The effect of the proposed GCL function is evident in the better regularized latent space, where the representation of similar image pairs (blue dots) are more consistently distributed towards the center of the space. The representations learned using the CL function, instead, form a more scattered and noisy distribution.

\section{Large-scale VPR: additional results}

\begin{table}[t!]
\centering
\caption{Detailed results on the Extended CMU dataset. The $^\star$ denotes the models for which PCA whitening has been applied. }
\label{tab:cmu_detailed}
\resizebox{\columnwidth}{!}{%
\setlength\tabcolsep{1.5pt}
\begin{tabular}{@{}lcccc@{}}
\toprule
\textbf{} & \textbf{Mean} & \textbf{Urban} & \textbf{Suburban} & \textbf{Park} \\ 
\textbf{Method} & \textbf{\begin{tabular}[c]{@{}c@{}}0.25/0.5/10m \\ 2/5/10$\degree$\end{tabular}} & \textbf{\begin{tabular}[c]{@{}c@{}}0.25/0.5/10m \\ 2/5/10$\degree$\end{tabular}} & \textbf{\begin{tabular}[c]{@{}c@{}}0.25/0.5/10m \\ 2/5/10$\degree$\end{tabular}} & \textbf{\begin{tabular}[c]{@{}c@{}}0.25/0.5/10m \\ 2/5/10$\degree$\end{tabular}} \\\midrule
NetVLAD-64 & 1.3 / 4.5 / 31.9 & 2.9 / 8.4 / 49.6 & 0.8 / 3.4 / 29.7 & 0.3 / 1.6 / 15.2 \\
NetVLAD-64$^\star$ & 3.9 / 12.1 / 58.4 & 7.6 / 20.3 / 77.2 & 2.6 / 9.7 / 58.9 & 1.6 / 6.3 / 37.0 \\
NetVLAD-16 & 1.7 / 5.5 / 39.1 & 3.7 / 10.0 / 57.3 & 1.1 / 4.5 / 38.6 & 0.4 / 1.9 / 19.8 \\
NetVLAD-16$^\star$ & 4.4 / 13.7 / 61.4 & 8.4 / 22.1 / 76.9 & 3.0 / 11.4 / 63.1 & 1.9 / 7.4 / 41.9 \\
\midrule
VGG-avg-CL & 0.9 / 3.0 / 22.7 & 2.0 / 5.8 / 39.4 & 0.6 / 2.3 / 21 & 0.2 / 0.7 / 6.5 \\
VGG-avg-GCL & 2.3 / 7.2 / 43.3 & 4.7 / 13.0 / 64.3 & 1.6 / 5.8 / 43.4 & 0.7 / 2.5 / 19.8 \\
VGG-avg-CL$^\star$ & 2.1 / 6.6 / 34.9 & 4.4 / 12.3 / 54.1 & 1.1 / 4.4 / 30.5 & 0.9 / 3.1 / 19.4 \\
VGG-avg-GCL$^\star$ & 3.7 / 11.2 / 52.5 & 7.2 / 19.1 / 70.8 & 2.1 / 8.2 / 50.4 & 1.8 / 6.2 / 35.1 \\
VGG-GeM-CL & 2.8 / 8.6 / 44.5 & 5.7 / 15.2 / 63.7 & 1.6 / 6.4 / 44.9 & 1.1 / 4.2 / 22.8 \\
VGG-GeM-GCL & 3.6 / 11.2 / 55.8 & 6.8 / 18.5 / 73.6 & 2.5 / 9.2 / 57.3 & 1.5 / 5.6 / 34.2 \\
VGG-GeM-CL$^\star$ & 4.4 / 13.4 / 56.5 & 8.4 / 21.5 / 72.1 & 2.6 / 10 / 55.2 & 2.3 / 8.7 / 40.8 \\
VGG-GeM-GCL$^\star$ & 5.7 / 17.1 / 66.3 & 10.4 / 26.8 / 82.2 & 3.8 / 13.8 / 67.8 & 2.8 / 10.7 / 46.7 \\\midrule
ResNet50-avg-CL & 2.6 / 7.8 / 43.4 & 5.6 / 14.9 / 64.6 & 1.4 / 5.5 / 42.6 & 0.8 / 2.9 / 21.1 \\
ResNet50-avg-GCL & 3.1 / 9.7 / 55.1 & 6 / 16.7 / 73.6 & 2.0 / 7.7 / 56.8 & 1.2 / 4.7 / 32.5 \\
ResNet50-avg-CL$^\star$ & 4.7 / 13.8 / 50.8 & 9.5 / 24.1 / 70.5 & 2.7 / 9.7 / 48.8 & 2.1 / 7.6 / 31.7 \\
ResNet50-avg-GCL$^\star$ & 5.4 / 16.2 / 66.5 & 10.2 / 26.5 / 81.8 & 3.5 / 12.5 / 69.8 & 2.6 / 9.6 / 45.3 \\
ResNet50-GeM-CL & 3.2 / 9.6 / 49.5 & 6.5 / 17.3 / 70.7 & 1.9 / 7.0 / 49.3 & 1.2 / 4.3 / 26.3 \\
ResNet50-GeM-GCL & 3.8 / 11.8 / 61.6 & 7.4 / 19.9 / 79.2 & 2.4 / 9.4 / 64.8 & 1.5 / 6.1 / 38.1 \\
ResNet50-GeM-CL$^\star$ & 4.7 / 13.4 / 51.6 & 9.5 / 23.9 / 73.5 & 2.8 / 9.6 / 49.7 & 1.9 / 6.8 / 30.0 \\
ResNet50-GeM-GCL$^\star$ & 5.4 / 16.5 / 69.9 & 10.1 / 26.3 / 84.5 & 3.5 / 13.4 / 74.1 & 2.6 / 9.8 / 48.2 \\\midrule
ResNet152-avg-CL & 3.0 / 9.2 / 49.9 & 6.2 / 16.4 / 70.4 & 1.9 / 6.9 / 48.7 & 1.0 / 4.1 / 28.9 \\
ResNet152-avg-GCL & 3.6 / 11.0 / 61.2 & 7.0 / 18.9 / 78.8 & 2.3 / 8.4 / 62.6 & 1.4 / 5.5 / 39.9 \\
ResNet152-avg-CL$^\star$ & 4.8 / 14.3 / 59.9 & 9.4 / 24 / 77.3 & 3.0 / 10.9 / 61 & 2 / 8 / 39.3 \\
ResNet152-avg-GCL$^\star$ & 5.7 / 17.0 / 66.5 & 10.8 / 27.3 / 82.0 & 3.7 / 13.5 / 68.9 & 2.6 / 10.1 / 46.3 \\
ResNet152-GeM-CL & 3.2 / 9.7 / 52.2 & 6.7 / 17.7 / 73.9 & 2.1 / 7.3 / 52.8 & 0.9 / 4.1 / 27.6 \\
ResNet152-GeM-GCL & 3.6 / 11.3 / 63.1 & 6.9 / 18.8 / 79.3 & 2.5 / 9.0 / 64.3 & 1.3 / 6.0 / 43.6 \\
ResNet152-GeM-CL$^\star$ & 4.8 / 14.2 / 55.0 & 9.6 / 24.5 / 75.2 & 3.0 / 10.5 / 54.3 & 1.9 / 7.5 / 33.6 \\
ResNet152-GeM-GCL$^\star$ & 5.3 / 16.1 / 66.4 & 9.9 / 25.8 / 81.6 & 3.5 / 12.8 / 69.3 & 2.4 / 9.7 / 46 \\\midrule
ResNext-avg-CL & 2.0 / 6.1 / 40.0 & 4.6 / 12.2 / 63.6 & 1.2 / 4.4 / 38.5 & 0.3 / 1.7 / 16.0 \\
ResNext-avg-GCL & 3.3 / 10.2 / 57.7 & 6.7 / 17.9 / 77.8 & 2.1 / 8.1 / 58.1 & 1.2 / 4.4 / 35.0 \\
ResNext-avg-CL$^\star$ & 4.6 / 13.4 / 58.6 & 9.3 / 23.2 / 78 & 2.8 / 10.1 / 59 & 1.7 / 6.7 / 36.5 \\
ResNext-avg-GCL$^\star$ & 5.6 / 16.6 / 70.7 & 10.4 / 26.6 / 85.1 & 3.7 / 13.4 / 73.2 & 2.6 / 9.8 / 51.6 \\
ResNext-GeM-CL & 2.9 / 9.0 / 52.6 & 5.9 / 15.6 / 70.2 & 1.7 / 6.8 / 53.0 & 1.2 / 4.7 / 32.6 \\
ResNext-GeM-GCL & 3.5 / 10.5 / 58.8 & 6.5 / 17.7 / 77.8 & 2.5 / 8.7 / 59.7 & 1.3 / 5.0 / 36.6 \\
ResNext-GeM-CL$^\star$ & 4.9 / 14.4 / 61.7 & 9.4 / 24 / 80.2 & 3.1 / 11.1 / 63.8 & 2.2 / 8.0 / 38.6 \\
ResNext-GeM-GCL$^\star$ & 6.1 / 18.2 / 74.9 & 11.1 / 28.7 / 87.4 & 4.2 / 14.6 / 77.4 & \textbf{3.1} / \textbf{11.1} / \textbf{57.7 }\\ \bottomrule
\end{tabular}
}
\end{table}

\subsection{RobotCar Seasons v2 and Ext. CMU Seasons}

We provide results divided by type of environment for the Extended CMU dataset in Table~\ref{tab:cmu_detailed}. We observed that all models (ours and SoTA) tend to reach a higher performance on the urban images. This is logical, as they are all trained on the MSLS dataset with images of mainly urban areas. 

We also provide detailed results for the RobotCar Seasons v2 dataset, organized by weather and illumination conditions, in Table~\ref{tab:robotcar_extended}. We observe that all methods obtain a higher successful localization rate on day conditions, and the GCL-based methods (especially with a ResNet backbone) tend to have goof performance on the night-time subsets as well. The MSLS train dataset includes images taken at night, although underrepresented, so it is interesting that GCL based models can localize images under these conditions.

\begin{table*}[!t]
\centering
\caption{Detailed results on the RobotCar Seasons v2 dataset, divided by wheather and ilumination conditions. The symbol $^\star$ denotes the models for which PCA whitening has been applied.}
\label{tab:robotcar_extended}
\renewcommand{\arraystretch}{0.95}
\resizebox{\textwidth}{!}{%
\setlength\tabcolsep{1.75pt}
\begin{tabular}{@{}l|c|c|c|c|c|c|c|c|c|c|c|c@{}}
\toprule
\textbf{} & \textbf{night rain} & \textbf{night} & \textbf{night all} & \textbf{overcast winter} & \textbf{sun} & \textbf{rain} & \textbf{snow} & \textbf{dawn} & \textbf{dusk} & \textbf{overcast summer} & \textbf{day all} & \textbf{mean} \\
\textbf{Method} & \textbf{\begin{tabular}[c]{@{}c@{}}0.25/0.5/10m \\ 2/5/10$\degree$\end{tabular}} & \textbf{\begin{tabular}[c]{@{}c@{}}0.25/0.5/10m \\ 2/5/10$\degree$\end{tabular}} & \textbf{\begin{tabular}[c]{@{}c@{}}0.25/0.5/10m \\ 2/5/10$\degree$\end{tabular}} & \textbf{\begin{tabular}[c]{@{}c@{}}0.25/0.5/10m \\ 2/5/10$\degree$\end{tabular}} & \textbf{\begin{tabular}[c]{@{}c@{}}0.25/0.5/10m \\ 2/5/10$\degree$\end{tabular}} & \textbf{\begin{tabular}[c]{@{}c@{}}0.25/0.5/10m \\ 2/5/10$\degree$\end{tabular}} & \textbf{\begin{tabular}[c]{@{}c@{}}0.25/0.5/10m \\ 2/5/10$\degree$\end{tabular}} & \textbf{\begin{tabular}[c]{@{}c@{}}0.25/0.5/10m \\ 2/5/10$\degree$\end{tabular}} & \textbf{\begin{tabular}[c]{@{}c@{}}0.25/0.5/10m \\ 2/5/10$\degree$\end{tabular}} & \textbf{\begin{tabular}[c]{@{}c@{}}0.25/0.5/10m \\ 2/5/10$\degree$\end{tabular}} & \textbf{\begin{tabular}[c]{@{}c@{}}0.25/0.5/10m \\ 2/5/10$\degree$\end{tabular}} & \textbf{\begin{tabular}[c]{@{}c@{}}0.25/0.5/10m \\ 2/5/10$\degree$\end{tabular}} \\ \midrule
NetVLAD-64 & 0.5 / 1.0 / 5.9 & 0.0 / 0.4 / 4.4 & 0.2 / 0.7 / 5.1 & 0.0 / 11.6 / 67.7 & 0.9 / 4.5 / 28.6 & 4.4 / 22.0 / 91.2 & 4.7 / 13.0 / 60.5 & 3.1 / 8.4 / 32.6 & 2.5 / 14.2 / 78.7 & 1.4 / 9.5 / 51.7 & 2.5 / 11.7 / 57.5 & 2 / 9.2 / 45.5 \\
NetVLAD-64$^\star$ & 0.0 / 1.5 / 11.8 & 0.0 / 1.3 / 12.4 & 0.0 / 1.4 / 12.1 & 0.0 / 15.2 / 87.2 & 3.1 / 12.1 / 66.5 & 8.8 / 35.6 / 99.0 & 7.0 / 28.4 / 90.2 & 9.3 / 21.6 / 77.5 & 4.1 / 25.4 / 98.0 & 5.2 / 21.3 / 77.7 & 5.5 / 22.9 / 84.7 & 4.2 / 18 / 68.1 \\
NetVLAD-16 & 0.0 / 0.0 / 3.4 & 0.0 / 0.9 / 5.3 & 0.0 / 0.5 / 4.4 & 1.2 / 10.4 / 78.0 & 0.9 / 6.2 / 33.0 & 5.9 / 25.4 / 93.7 & 3.3 / 10.7 / 62.8 & 1.8 / 6.6 / 29.1 & 1.5 / 16.2 / 86.8 & 1.4 / 8.1 / 57.8 & 2.3 / 11.8 / 61.5 & 1.8 / 9.2 / 48.4 \\
NetVLAD-16$^\star$ & 0.0 / 0.0 / 1.0 & 0.0 / 0.9 / 7.1 & 0.0 / 0.5 / 4.2 & 1.8 / 19.5 / 90.2 & 4.9 / 11.6 / 67.4 & 10.7 / 34.1 / 96.1 & 6.5 / 24.7 / 89.8 & 7.9 / 24.2 / 74.4 & 2.5 / 23.4 / 93.9 & 7.6 / 24.6 / 77.3 & 6.2 / 23.1 / 83.6 & 4.8 / 17.9 / 65.3 \\
\midrule
VGG-avg-CL & 0.0 / 0.0 / 2.5 & 0.0 / 0.0 / 1.3 & 0.0 / 0.0 / 1.9 & 0.0 / 7.9 / 54.9 & 0.0 / 2.2 / 14.3 & 4.9 / 23.9 / 85.4 & 3.3 / 8.8 / 49.8 & 0.9 / 1.3 / 14.5 & 0.0 / 12.2 / 66.5 & 0.0 / 3.3 / 31.8 & 1.3 / 8.3 / 44.0 & 1 / 6.4 / 34.4 \\
VGG-avg-GCL & 0.0 / 0.5 / 3.9 & 0.0 / 0.0 / 1.8 & 0.0 / 0.2 / 2.8 & 0.6 / 15.2 / 82.9 & 2.2 / 6.7 / 44.6 & 7.8 / 31.7 / 95.1 & 4.2 / 19.5 / 73.0 & 0.4 / 5.7 / 33.9 & 2.0 / 21.8 / 86.3 & 3.8 / 13.7 / 57.8 & 3.0 / 16.1 / 66.3 & 2.3 / 12.5 / 51.7 \\
VGG-avg-CL$^\star$ & 0.0 / 0.0 / 0.0 & 0.0 / 0.0 / 0.0 & 0.0 / 0.0 / 0.0 & 0.0 / 11.0 / 61.6 & 0.4 / 4.0 / 20.5 & 5.9 / 26.3 / 85.4 & 5.1 / 13.5 / 64.2 & 3.1 / 8.4 / 40.1 & 4.1 / 17.3 / 76.6 & 1.9 / 11.8 / 40.3 & 3.0 / 13.0 / 54.5 & 2.3 / 10 / 42 \\
VGG-avg-GCL$^\star$ & 0.0 / 0.0 / 1.0 & 0.0 / 0.0 / 0.4 & 0.0 / 0.0 / 0.7 & 0.6 / 17.1 / 79.3 & 3.1 / 11.2 / 49.1 & 9.3 / 34.6 / 97.6 & 6.5 / 17.7 / 80.0 & 3.5 / 16.3 / 56.4 & 5.6 / 25.4 / 90.9 & 3.8 / 18.0 / 69.7 & 4.7 / 19.9 / 73.9 & 3.6 / 15.3 / 57.1 \\
VGG-GeM-CL & 0.0 / 0.0 / 2.0 & 0.0 / 0.4 / 1.8 & 0.0 / 0.2 / 1.9 & 0.0 / 14.6 / 80.5 & 2.2 / 6.7 / 45.5 & 8.3 / 31.2 / 94.1 & 6.0 / 20.0 / 77.7 & 3.5 / 11.5 / 47.1 & 2.0 / 17.8 / 89.8 & 4.7 / 18.0 / 68.2 & 4.0 / 17.0 / 70.8 & 3.1 / 13.2 / 55 \\
VGG-GeM-GCL & 0.0 / 0.5 / 7.4 & 0.0 / 0.0 / 0.9 & 0.0 / 0.2 / 4.0 & 0.6 / 16.5 / 82.3 & 2.2 / 11.6 / 61.2 & 10.2 / 38.0 / 99.0 & 7.9 / 23.3 / 83.7 & 2.6 / 11.5 / 51.1 & 2.0 / 22.3 / 90.9 & 7.1 / 20.9 / 70.6 & 4.8 / 20.4 / 76.2 & 3.7 / 15.8 / 59.7 \\
VGG-GeM-CL$^\star$ & 0.0 / 0.5 / 2.0 & 0.0 / 0.0 / 0.0 & 0.0 / 0.2 / 0.9 & 0.6 / 18.9 / 84.8 & 2.7 / 13.4 / 60.3 & 8.3 / 36.6 / 98.5 & 8.4 / 27.0 / 85.1 & 6.6 / 22.5 / 70.5 & 5.6 / 29.9 / 95.9 & 4.7 / 21.3 / 74.9 & 5.4 / 24.2 / 80.8 & 4.2 / 18.7 / 62.5 \\
VGG-GeM-GCL$^\star$ & 0.5 / 1.0 / 6.4 & 0.0 / 1.3 / 6.6 & 0.2 / 1.2 / 6.5 & 1.2 / 23.2 / 90.9 & 6.7 / 21.4 / 78.1 & 11.2 / 42.4 / \textbf{100.0} & 9.3 / 32.6 / 95.3 & 7.5 / 20.7 / 76.7 & 4.6 / 28.9 / 97.5 & 6.2 / 27.0 / 79.6 & 6.9 / 28.0 / 87.9 & 5.4 / 21.9 / 69.2 \\ \midrule
ResNet50-avg-CL & 0.5 / 1.0 / 9.4 & 0.4 / 2.2 / 11.1 & 0.5 / 1.6 / 10.3 & 0.0 / 13.4 / 71.3 & 1.8 / 6.2 / 30.4 & 8.8 / 32.7 / 97.6 & 5.6 / 18.6 / 64.7 & 4.4 / 15.4 / 53.7 & 3.0 / 14.7 / 79.2 & 2.8 / 10.4 / 51.7 & 3.9 / 15.9 / 63.1 & 3.1 / 12.6 / 51 \\
ResNet50-avg-GCL & 1.0 / 2.0 / 12.8 & 0.4 / 1.3 / 8.4 & 0.7 / 1.6 / 10.5 & 0.6 / 18.3 / 80.5 & 1.3 / 6.7 / 43.8 & 8.3 / 34.1 / 96.1 & 4.7 / 15.3 / 72.6 & 2.6 / 11.0 / 51.1 & 2.0 / 16.2 / 88.8 & 4.3 / 14.2 / 64.0 & 3.5 / 16.3 / 69.9 & 2.9 / 12.9 / 56.3 \\
ResNet50-avg-CL$^\star$ & 0.0 / 0.0 / 3.0 & 0.0 / 0.4 / 1.8 & 0.0 / 0.2 / 2.3 & 2.4 / 26.2 / 89.6 & 3.6 / 14.3 / 53.1 & \textbf{14.6 / 45.4 }/ 98.0 & 8.4 / 30.7 / 84.2 & 10.1 / 31.7 / 75.3 & 5.6 / 24.9 / 93.4 & 7.6 / 33.6 / 76.8 & 7.6 / 29.5 / 80.7 & 5.9 / 22.8 / 62.7 \\
ResNet50-avg-GCL$^\star$ & 0.0 / 1.0 / 6.4 & 0.4 / 0.4 / 10.2 & 0.2 / 0.7 / 8.4 & \textbf{3.0} / 28.0 / 98.2 & 4.0 / 15.6 / 68.8 & 13.2 / 41.5 / 97.6 & 8.4 / 31.2 / 87.9 & 8.4 / 30.4 / 84.1 & 5.6 / 27.9 / 96.4 & 9.5 / 33.6 / 86.3 & 7.6 / 29.7 / 87.8 & 5.9 / 23.1 / 69.6 \\
ResNet50-GeM-CL & 0.5 / 1.5 / 11.8 & 0.4 / 1.3 / 8.0 & 0.5 / 1.4 / 9.8 & 0.0 / 16.5 / 82.9 & 0.9 / 9.8 / 61.2 & 9.8 / 33.7 / 97.1 & 5.6 / 23.7 / 81.4 & 3.1 / 15.0 / 62.6 & 1.5 / 19.8 / 93.4 & 6.2 / 18.5 / 64.9 & 4.0 / 19.5 / 76.9 & 3.2 / 15.4 / 61.5 \\
ResNet50-GeM-GCL & 0.5 / 2.0 / 9.9 & 0.0 / 1.3 / 11.9 & 0.2 / 1.6 / 11.0 & 1.8 / 21.3 / 84.8 & 1.8 / 10.3 / 47.3 & 8.8 / 33.7 / 97.1 & 5.1 / 18.1 / 80.9 & 1.8 / 8.8 / 49.8 & 2.0 / 17.8 / 91.9 & 4.7 / 16.6 / 69.7 & 3.7 / 17.7 / 73.4 & 2.9 / 14 / 58.8 \\
ResNet50-GeM-CL$^\star$ & 0.5 / 0.5 / 3.9 & 0.0 / 0.4 / 2.7 & 0.2 / 0.5 / 3.3 & \textbf{3.0} / 28.0 / 95.1 & 2.2 / 13.8 / 56.7 & 9.8 / 37.6 / 98.5 & 8.4 / 32.6 / 94.0 & 7.0 / 23.8 / 83.7 & 5.6 / 24.9 / 94.4 & 8.1 / 30.8 / 79.6 & 6.4 / 27.2 / 85.3 & 5 / 21.1 / 66.5 \\
ResNet50-GeM-GCL$^\star$ & 0.0 / 0.0 / 8.4 & 0.0 / 0.0 / 9.3 & 0.0 / 0.0 / 8.9 & \textbf{3.0} / 25.6 / 95.7 & 3.1 / 14.7 / 69.2 & 9.3 / 34.6 / 98.0 & 6.5 / 28.4 / 90.7 & 9.3 / 28.6 / 86.3 & 3.6 / 25.9 / 96.4 & 7.1 / 26.1 / 83.9 & 6.1 / 26.2 / 88.1 & 4.7 / 20.2 / 70 \\ \midrule
ResNet152-avg-CL & 0.0 / 0.0 / 8.4 & 0.0 / 0.0 / 8.0 & 0.0 / 0.0 / 8.2 & 1.2 / 17.1 / 86.0 & 1.8 / 11.2 / 51.3 & 6.8 / 34.1 / 92.7 & 4.7 / 15.8 / 74.0 & 2.6 / 11.9 / 53.3 & 3.0 / 17.3 / 82.7 & 2.4 / 13.3 / 64.0 & 3.3 / 17.0 / 71.0 & 2.5 / 13.1 / 56.6 \\
ResNet152-avg-GCL & 0.0 / 1.5 / 11.3 & 0.4 / 0.4 / 10.2 & 0.2 / 0.9 / 10.7 & 1.8 / 20.1 / 90.2 & 3.6 / 8.5 / 52.7 & 9.3 / 29.3 / 95.6 & 4.7 / 19.1 / 84.2 & 2.6 / 12.8 / 47.6 & 3.0 / 16.8 / 88.8 & 4.3 / 16.1 / 70.6 & 4.2 / 17.3 / 74.5 & 3.3 / 13.5 / 59.9 \\
ResNet152-avg-CL$^\star$ & 3.0 / 10.9 / 61.0 & 4.3 / 13.6 / 57.6 & 9.4 / 24.0 / 77.3 & 2.0 / 8.0 / 39.3 & 5.1 / 14.0 / 60.8 & 4.9 / 14.2 / 58.1 & 4.8 / 14.1 / 59.8 & 4.5 / 13.3 / 55.1 & 5.3 / 17.0 / 69.7 & 3.6 / 12.3 / 53.7 & 5.5 / 16.2 / 67.2 & 5.4 / 22.2 / 64.8 \\
ResNet152-avg-GCL$^\star$ & \textbf{3.7 / 13.5 / 68.9} & \textbf{5.3 / 16.6 / 65.0} & \textbf{10.8 / 27.3 / 82.0} & 2.6 / 10.1 / 46.3 & 6.1 / 16.7 / 68.2 & 5.7 / 16.8 / 63.1 & 5.6 / 16.4 / 65.0 & 5.7 / 16.4 / 63.9 & 6.1 / 19.6 / 74.9 & 4.3 / 14.6 / 61.6 & 6.2 / 18.5 / 73.6 & 6.2 / 23.9 / 70 \\
ResNet152-GeM-CL & 2.1 / 7.3 / 52.8 & 3.0 / 9.8 / 53.9 & 6.7 / 17.7 / 73.9 & 0.9 / 4.1 / 27.6 & 3.2 / 8.9 / 52.2 & 3.3 / 9.2 / 47.4 & 3.1 / 8.9 / 47.9 & 3.3 / 9.9 / 53.0 & 3.4 / 11.5 / 61.0 & 1.9 / 7.9 / 45.6 & 3.7 / 11.0 / 56.3 & 3.3 / 15.2 / 64 \\
ResNet152-GeM-GCL & 2.5 / 9.0 / 64.3 & 3.3 / 11.0 / 63.8 & 6.9 / 18.8 / 79.3 & 1.3 / 6.0 / 43.6 & 4.1 / 11.4 / 62.9 & 3.5 / 11.2 / 58.9 & 3.5 / 10.8 / 60.1 & 3.7 / 11.2 / 62.9 & 3.6 / 12.6 / 70.4 & 2.5 / 9.4 / 58.9 & 3.8 / 11.9 / 68.4 & 2.9 / 13.1 / 63.5 \\
ResNet152-GeM-CL$^\star$ & 3.0 / 10.5 / 54.3 & 4.4 / 13.6 / 53.2 & 9.6 / 24.5 / 75.2 & 1.9 / 7.5 / 33.6 & 4.9 / 13.4 / 53.9 & 5.0 / 14.1 / 53.5 & 4.9 / 13.7 / 54.6 & 4.4 / 12.9 / 50.0 & 5.6 / 17.5 / 65.6 & 3.9 / 12.5 / 47.1 & 5.5 / 15.9 / 62.2 & 6.1 / 23.5 / 68.9 \\
ResNet152-GeM-GCL$^\star$ & 3.5 / 12.8 / 69.3 & 4.8 / 16.0 / 65.4 & 9.9 / 25.8 / 81.6 & 2.4 / 9.7 / 46.0 & 5.6 / 15.6 / 67.0 & 5.3 / 15.5 / 62.8 & 5.2 / 15.2 / 64.7 & 5.1 / 15.6 / 63.7 & 5.7 / 19.5 / 75.8 & 4.0 / 14.8 / 62.3 & 6.1 / 17.9 / 74.0 & 6 / 21.6 / 72.5 \\ \midrule
ResNext-avg-CL & 1.0 / 3.4 / 29.1 & 0.0 / 1.3 / 15.5 & 0.5 / 2.3 / 21.9 & 1.2 / 12.8 / 86.0 & 1.3 / 4.9 / 55.8 & 6.8 / 20.5 / 97.6 & 6.5 / 16.7 / 80.0 & 2.2 / 9.7 / 44.5 & 2.5 / 17.8 / 84.8 & 2.8 / 13.3 / 66.8 & 3.4 / 13.5 / 72.6 & 2.7 / 10.9 / 61 \\
ResNext-avg-GCL & 0.0 / 0.5 / 11.8 & 0.0 / 0.9 / 15.9 & 0.0 / 0.7 / 14.0 & 1.2 / 19.5 / 91.5 & 2.7 / 8.9 / 60.3 & 6.8 / 25.4 / 96.6 & 4.2 / 15.3 / 78.1 & 2.6 / 11.9 / 63.4 & 2.5 / 16.2 / 93.9 & 4.3 / 15.6 / 66.8 & 3.5 / 15.9 / 77.7 & 2.7 / 12.4 / 63.1 \\
ResNext-avg-CL$^\star$ & 1.0 / 1.5 / 12.3 & 0.0 / 0.9 / 12.8 & 0.5 / 1.2 / 12.6 & 0.6 / 25.6 / 97.6 & 4.9 / 17.0 / 79.9 & 9.3 / 35.1 / 99.0 & 8.4 / 29.8 / 93.0 & 7.0 / 23.8 / 72.7 & 5.6 / 24.9 / 94.4 & 5.7 / 27.0 / 82.9 & 6.1 / 26.1 / 87.9 & 4.8 / 20.4 / 70.6 \\
ResNext-avg-GCL$^\star$ & 1.0 / 3.4 / 29.1 & 0.0 / 1.3 / 15.5 & 0.5 / 2.3 / 21.9 & 1.2 / 12.8 / 86.0 & 1.3 / 4.9 / 55.8 & 6.8 / 20.5 / 97.6 & 6.5 / 16.7 / 80.0 & 2.2 / 9.7 / 44.5 & 2.5 / 17.8 / 84.8 & 2.8 / 13.3 / 66.8 & 3.4 / 13.5 / 72.6 & 2.7 / 10.9 / 61 \\
ResNext-GeM-CL & 0.0 / 1.5 / 6.9 & 0.0 / 0.4 / 8.0 & 0.0 / 0.9 / 7.5 & 0.0 / 17.1 / 82.3 & 1.3 / 5.4 / 56.7 & 3.9 / 20.5 / 96.6 & 3.7 / 12.6 / 74.9 & 2.2 / 7.0 / 29.5 & 2.0 / 21.3 / 90.4 & 2.8 / 11.4 / 60.7 & 2.4 / 13.2 / 68.9 & 1.9 / 10.4 / 54.8 \\
ResNext-GeM-GCL & 0.0 / 2.5 / 22.2 & 0.4 / 1.3 / 19.9 & 0.2 / 1.9 / 21.0 & 1.2 / 15.2 / 84.8 & 0.9 / 7.6 / 68.8 & 7.3 / 28.3 / 98.5 & 5.1 / 20.0 / 84.2 & 2.2 / 11.0 / 55.5 & 4.1 / 20.8 / 90.9 & 3.3 / 16.1 / 71.6 & 3.5 / 16.8 / 78.4 & 2.7 / 13.4 / 65.2 \\
ResNext-GeM-CL$^\star$ & 0.5 / 2.5 / 10.3 & 0.0 / 0.0 / 11.9 & 0.2 / 1.2 / 11.2 & 1.8 / 20.1 / 92.1 & 4.9 / 15.6 / 79.0 & 7.8 / 31.2 / \textbf{100.0} & 7.0 / 26.0 / 93.0 & 3.5 / 15.0 / 67.0 & 4.6 / 26.9 / 95.9 & 3.8 / 19.4 / 73.0 & 4.9 / 21.9 / 85.1 & 3.8 / 17.2 / 68.2 \\
ResNext-GeM-GCL$^\star$ & 2.5 / 5.4 / 38.4 & 1.8 / 3.5 / 28.3 & 2.1 / 4.4 / 33.1 & 1.2 / 26.8 / 93.9 & 3.6 / 15.2 / 77.2 & 9.8 / 40.0 / 98.5 & 7.9 / 29.3 / 91.2 & 4.8 / 19.8 / 82.4 & 5.1 / 25.9 / 92.9 & 5.7 / 26.1 / 76.8 & 5.5 / 25.9 / 87.1 & 4.7 / 21 / 74.7 \\ \bottomrule
\end{tabular}%

}
\end{table*}

\subsection{Results on Pittsburgh250k and TokyoTM}
We evaluated the generalization of our models trained on MSLS to the Pittsburgh250k~\cite{Torii-PAMI2015} and TokyoTM~\cite{Arandjelovic2017} datasets. The test set of the former consists of 83k map and 8k query images taken over the span of several years in Pittsburgh, Pennsylvania, USA. The TokyoTM dataset consists of images collected using the Time Machine tool on Google Street View in Tokyo over several years. Its validation set is divided into a map and a query set, with 49k and 7k images.

The results (see Table~\ref{tab:ablation-pitts250k}) are in line with those reported in the main paper. Our models trained with a GCL function generalize better to unseen datasets than their counterpart trained with a binary Contrastive Loss. Their performance is further boosted if PCA whitening is applied, up to a top-5 recall of 93.7\% on Pittsburgh250k and 96.7\% on TokyoTM.

\begin{table}[t!]
\centering
\renewcommand{\arraystretch}{0.95}
\resizebox{\columnwidth}{!}{%
\begin{tabular}{@{}lcccccccc@{}}
\toprule
\textbf{} & \textbf{} & \textbf{} & \multicolumn{3}{c}{\textbf{Pittsburgh250k}} & \multicolumn{3}{c}{\textbf{TokyoTM}} \\ 
\textbf{Method} & \textbf{PCA$_w$} & \textbf{Dim} & \textbf{R@1} & \textbf{R@5} & \textbf{R@10} & \textbf{R@1} & \textbf{R@5} & \textbf{R@10} \\\midrule
 VGG-avg-CL & No & 512 & 20.2 & 38.0 & 47.2 & 38.9 & 58.5 & 67.2 \\
VGG-avg-GCL & No & 512 & 32.1 & 53.4 & 62.7 & 65.6 & 80.8 & 85.2 \\
VGG-avg-CL & Yes & 128 & 39.3 & 58.9 & 67.4 & 66.0 & 80.3 & 85.0 \\
VGG-avg-GCL & Yes & 256 & 51.3 & 71.0 & 78.0 & 78.6 & 87.7 & 90.8 \\
VGG-GeM-CL & No & 512 & 44.5 & 63.1 & 70.1 & 67.7 & 80.8 & 85.0 \\
VGG-GeM-GCL & No & 512 & 53.3 & 72.4 & 79.2 & 75.5 & 85.4 & 88.6 \\
VGG-GeM-CL & Yes & 512 & 65.1 & 81.2 & 85.8 & 83.7 & 91.1 & 93.3 \\
VGG-GeM-GCL & Yes & 512 & 73.4 & 86.4 & 89.9 & 88.2 & 93.2 & 94.7 \\\midrule
ResNet50-avg-CL & No & 2048 & 45.1 & 65.8 & 73.6 & 69.7 & 83.0 & 87.5 \\
ResNet50-avg-GCL & No & 2048 & 56.2 & 77.1 & 83.8 & 72.3 & 84.8 & 88.4 \\
ResNet50-avg-CL & Yes & 2048 & 70.6 & 85.4 & 89.5 & 90.5 & 94.8 & 96.0 \\
ResNet50-avg-GCL & Yes & 1024 & 74.6 & 87.9 & 91.6 & 91.7 & 95.7 & 96.8 \\
ResNet50-GeM-CL & No & 2048 & 54.8 & 74.2 & 80.7 & 73.3 & 85.4 & 89.2 \\
ResNet50-GeM-GCL & No & 2048 & 68.2 & 84.6 & 89.2 & 80.1 & 88.8 & 91.6 \\
ResNet50-GeM-CL & Yes & 1024 & 72.4 & 86.6 & 90.4 & 88.7 & 93.9 & 95.4 \\
ResNet50-GeM-GCL & Yes & 1024 & 80.9 & 91.4 & 94.3 & 92.2 & 95.6 & 96.8 \\\midrule
ResNet152-avg-CL & No & 2048 & 51.3 & 73.0 & 80.1 & 73.5 & 86.4 & 89.9 \\
ResNet152-avg-GCL & No & 2048 & 64.0 & 83.6 & 89.2 & 78.9 & 89.0 & 91.9 \\
ResNet152-avg-CL & Yes & 1024 & 69.6 & 85.9 & 90.6 & 90.7 & 95.1 & 96.3 \\
ResNet152-avg-GCL & Yes & 2048 & 80.9 & 92.2 & 95.3 & \textbf{94.0} & \textbf{96.7} & \textbf{97.3} \\
ResNet152-GeM-CL & No & 2048 & 60.7 & 79.0 & 85.1 & 77.8 & 88.3 & 91.2 \\
ResNet152-GeM-GCL & No & 2048 & 68.0 & 84.9 & 89.8 & 81.5 & 90.3 & 92.8 \\
ResNet152-GeM-CL & Yes & 2048 & 76.2 & 89.9 & 93.4 & 91.1 & 95.3 & 96.4 \\
ResNet152-GeM-GCL & Yes & 2048 & \textbf{83.8} & \textbf{93.7} & \textbf{96.1} & 93.1 & 96.1 & 96.8 \\\midrule
ResNeXt-avg-CL & No & 2048 & 44.2 & 65.8 & 73.9 & 69.0 & 82.3 & 86.3 \\
ResNeXt-avg-GCL & No & 2048 & 57.9 & 76.1 & 82.6 & 77.7 & 86.9 & 89.7 \\
ResNeXt-avg-CL & Yes & 1024 & 69.6 & 85.4 & 89.8 & 88.9 & 94.1 & 95.6 \\
ResNeXt-avg-GCL & Yes & 1024 & 74.7 & 88.0 & 91.7 & 89.6 & 94.6 & 95.9 \\
ResNeXt-GeM-CL & No & 2048 & 50.2 & 70.8 & 78.8 & 66.1 & 78.5 & 82.7 \\
ResNeXt-GeM-GCL & No & 2048 & 58.6 & 76.2 & 81.8 & 78.7 & 87.1 & 90.1 \\
ResNeXt-GeM-CL & Yes & 1024 & 70.3 & 85.2 & 89.6 & 87.2 & 93.1 & 94.5 \\
ResNeXt-GeM-GCL & Yes & 1024 & 78.2 & 90.1 & 93.1 & 91.8 & 95.4 & 96.6 \\ \bottomrule
\end{tabular}%
\vspace{-3mm}
\caption{Generalization results of the models trained on the MSLS dataset for the Pittsburgh250k and TokyoTM benchmarks.}
\label{tab:ablation-pitts250k}
}
\end{table}

\begin{table*}[!t]
\centering
\caption{Ablation study: all the models are trained on the MSLS train set and deploy a global average pooling layer. When PCA whitening is applied we report the descriptor size that achieves the best results on the MSLS validation set.}
\label{tab:ablation-avg}
\renewcommand{\arraystretch}{0.95}
\resizebox{\textwidth}{!}{%
\setlength\tabcolsep{1.5pt}
\begin{tabular}
{@{}lcccccccccccccccccccc@{}}
\toprule
\textbf{} & \textbf{} & \textbf{} & \multicolumn{3}{c}{\textbf{MSLS-Val}} & \multicolumn{3}{c}{\textbf{MSLS-Test}} & \multicolumn{3}{c}{\textbf{Pittsburgh30k}} & \multicolumn{3}{c}{\textbf{Tokyo24/7}} & \multicolumn{3}{c}{\textbf{RobotCar Seasons V2}} & \multicolumn{3}{l}{\textbf{Extended CMU Seasons}} \\
\textbf{Method} & \textbf{PCA$_w$} & \textbf{Dim} & \textbf{R@1} & \textbf{R@5} & \textbf{R@10} & \textbf{R@1} & \textbf{R@5} & \textbf{R@10} & \textbf{R@1} & \textbf{R@5} & \textbf{R@10} & \textbf{R@1} & \textbf{R@5} & \textbf{R@10} & \textbf{0.25m/2$\degree$} & \textbf{0.5m/5$\degree$} & \textbf{5.0m/10$\degree$} & \textbf{0.25m/2$\degree$} & \textbf{0.5m/5$\degree$} & \textbf{5.0m/10$\degree$} \\ \midrule
VGG-avg-CL & No & 512 & 28.8 & 47.0 & 53.9 & 16.9 & 30.5 & 36.4 & 28.9 & 53.7 & 64.8 & 10.5 & 24.1 & 35.6 & 1.0 & 6.4 & 34.4 & 0.9 & 3.0 & 22.7 \\
VGG-avg-GCL & No & 512 & 48.8 & 67.8 & 73.1 & 21.5 & 31.4 & 37.9 & 42.1 & 66.9 & 77.3 & 20.0 & 42.5 & 52.1 & 2.3 & 12.5 & 51.7 & 2.3 & 7.2 & 43.3 \\
VGG-avg-CL & Yes & 128 & 35.0 & 53.4 & 60.1 & 35.2 & 47.3 & 54.1 & 47.1 & 69.5 & 78.3 & 16.2 & 28.9 & 40.0 & 2.3 & 10.0 & 42.0 & 2.1 & 6.6 & 34.9 \\
VGG-avg-GCL & Yes & 256 & 54.5 & 72.6 & 78.2 & 32.9 & 49.0 & 56.5 & 56.2 & 76.7 & 83.9 & 28.6 & 45.7 & 54.9 & 3.6 & 15.3 & 57.1 & 3.7 & 11.2 & 52.5 \\ \midrule
ResNet50-avg-CL & No & 2048 & 44.3 & 60.3 & 65.9 & 24.9 & 39.0 & 44.6 & 54.0 & 75.7 & 83.1 & 20.6 & 40.0 & 50.2 & 3.1 & 12.6 & 51.0 & 2.6 & 7.8 & 43.4 \\
ResNet50-avg-GCL & No & 2048 & 59.6 & 72.3 & 76.2 & 35.8 & 52.0 & 59.0 & 62.5 & 82.7 & 88.4 & 24.1 & 44.1 & 54.6 & 2.9 & 12.9 & 56.3 & 3.1 & 9.7 & 55.1 \\
ResNet50-avg-CL & Yes & 2048 & 58.8 & 71.4 & 75.8 & 33.1 & 46.5 & 53.3 & 65.8 & 82.6 & 88.2 & 48.6 & 63.2 & 70.5 & 5.9 & 22.8 & 62.7 & 4.7 & 13.8 & 50.8 \\
ResNet50-avg-GCL & Yes & 1024 & 69.5 & 81.2 & 85.5 & 44.2 & 57.8 & 63.4 & 73.3 & 87.1 & 91.2 & 52.1 & 68.9 & 72.7 & 5.9 & 23.1 & 69.6 & 5.4 & 16.2 & 66.5 \\ \midrule
ResNet152-avg-CL & No & 2048 & 53.1 & 70.1 & 75.4 & 29.7 & 44.2 & 51.3 & 59.7 & 80.3 & 87.0 & 27.0 & 48.6 & 58.4 & 2.5 & 13.1 & 56.6 & 3.0 & 9.2 & 49.9 \\
ResNet152-avg-GCL & No & 2048 & 65.1 & 80.0 & 83.8 & 43.5 & 59.2 & 65.2 & 69.3 & 87.2 & 91.3 & 32.1 & 52.1 & 62.2 & 3.3 & 13.5 & 59.9 & 3.6 & 11.0 & 61.2 \\
ResNet152-avg-CL & Yes & 1024 & 63.0 & 77.7 & 81.5 & 37.7 & 51.6 & 56.9 & 68.8 & 85.9 & 90.4 & 49.8 & 67.3 & 74.3 & 5.4 & 22.2 & 64.8 & 4.8 & 14.3 & 59.9 \\
ResNet152-avg-GCL & Yes & 2048 & 75.8 & 87.4 & 89.7 & 52.7 & 68.1 & 74.2 & \textbf{77.9} & \textbf{90.4} & \textbf{93.5} & \textbf{64.4} & \textbf{77.8} & \textbf{83.2} & \textbf{6.2} & \textbf{23.9} & 70.0 & \textbf{5.7} & \textbf{17.0} & 66.5 \\ \midrule
ResNeXt-avg-CL & No & 2048 & 58.9 & 75.1 & 79.9 & 34.5 & 50.1 & 57.7 & 51.3 & 73.6 & 81.9 & 24.8 & 47.3 & 56.8 & 2.7 & 10.9 & 61.0 & 2.0 & 6.1 & 40.0 \\
ResNeXt-avg-GCL & No & 2048 & 72.2 & 85.1 & 87.3 & 51.5 & 66.9 & 71.7 & 62.9 & 81.0 & 87.1 & 39.4 & 58.1 & 68.9 & 2.7 & 12.4 & 63.1 & 3.3 & 10.2 & 57.7 \\
ResNeXt-avg-CL & Yes & 1024 & 71.6 & 84.7 & 88.0 & 46.5 & 62.9 & 68.9 & 69.2 & 85.3 & 89.6 & 44.8 & 63.5 & 73.6 & 4.8 & 20.4 & \textbf{70.6} & 4.6 & 13.4 & 58.6 \\
ResNeXt-avg-GCL & Yes & 1024 & \textbf{79.3} & \textbf{89.2} & \textbf{90.3} & \textbf{57.8} & \textbf{72.3} & \textbf{77.1} & 74.8 & 88.2 & 91.8 & 53.0 & 76.2 & 80.6 & 2.7 & 10.9 & 61.0 & 5.6 & 16.6 & \textbf{70.7} \\ \bottomrule

\end{tabular}%
}
\end{table*}

\section{Additional ablations}
\subsection{Average pooling and PCA}
In addition to using GeM, we trained the considered backbones (i.e. VGG16, ResNet50, ResNet152 and ResNeXt) using a Global Average Pooling layer on the MSLS dataset. We show the results in Table~\ref{tab:ablation-avg}. We observed that our method can reach good results also with a simple pooling layer, although a GeM layer usually leads to better results (see main paper). Our methods reach good results on the validation and test sets of MSLS and generalizes well to unseen datasets such as Pittsburgh30k~\cite{Arandjelovic2017}, Tokyo24/7~\cite{Arandjelovic2017}, RobotCar Seasons V2~\cite{sattler2018benchmarking} and Extended CMU Seasons~\cite{sattler2018benchmarking}. As we observed also with the GeM models, the Global Average Pooling models achieve better performance when PCA whitening is applied, up to a 72.3\% top-5 recall on the test set of MSLS.

\subsection{Composition of training batches. }

\begin{figure}[!t]
    \centering
    \begin{subfigure}{.5\textwidth}
    \includegraphics[width=\textwidth]{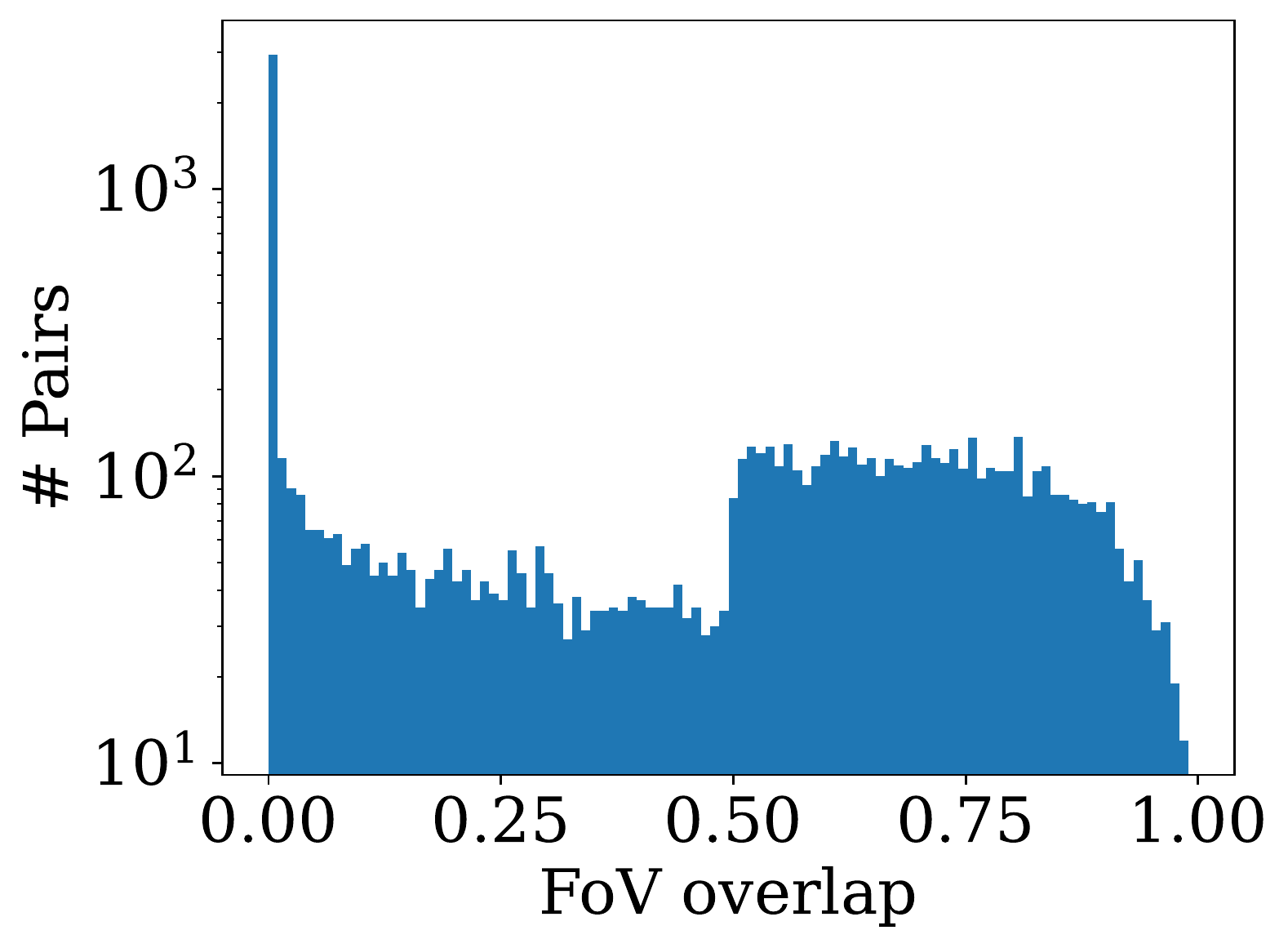}
    \label{fig:pairs_50_25_25}\vspace{-2mm} 
    \end{subfigure}\begin{subfigure}{.5\textwidth}
    \includegraphics[width=\textwidth]{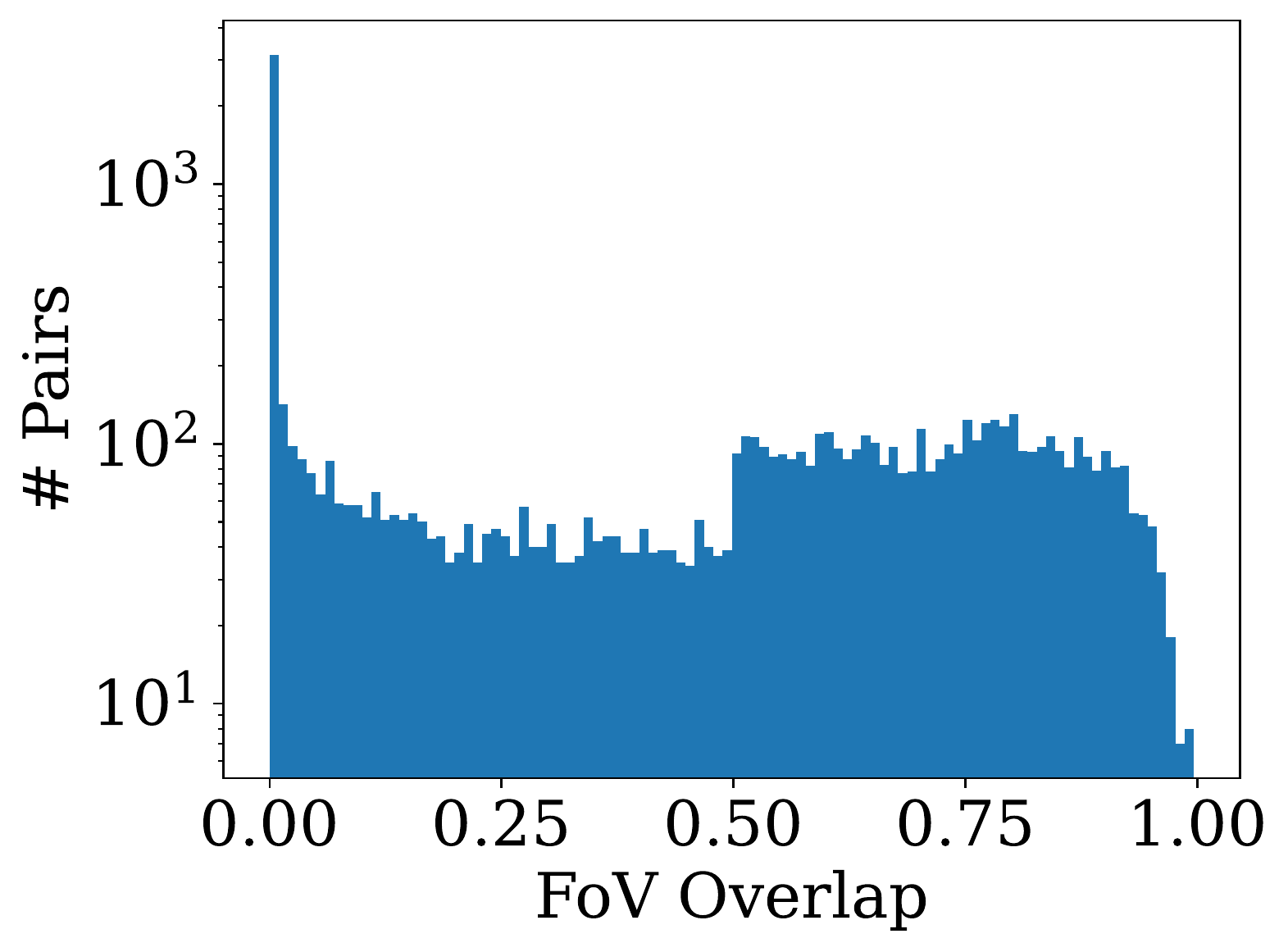}
    \label{fig:pairs_25_25_25_25}\vspace{-2mm}
    \end{subfigure}
    
        \begin{subfigure}{.5\textwidth}
    \includegraphics[width=\textwidth]{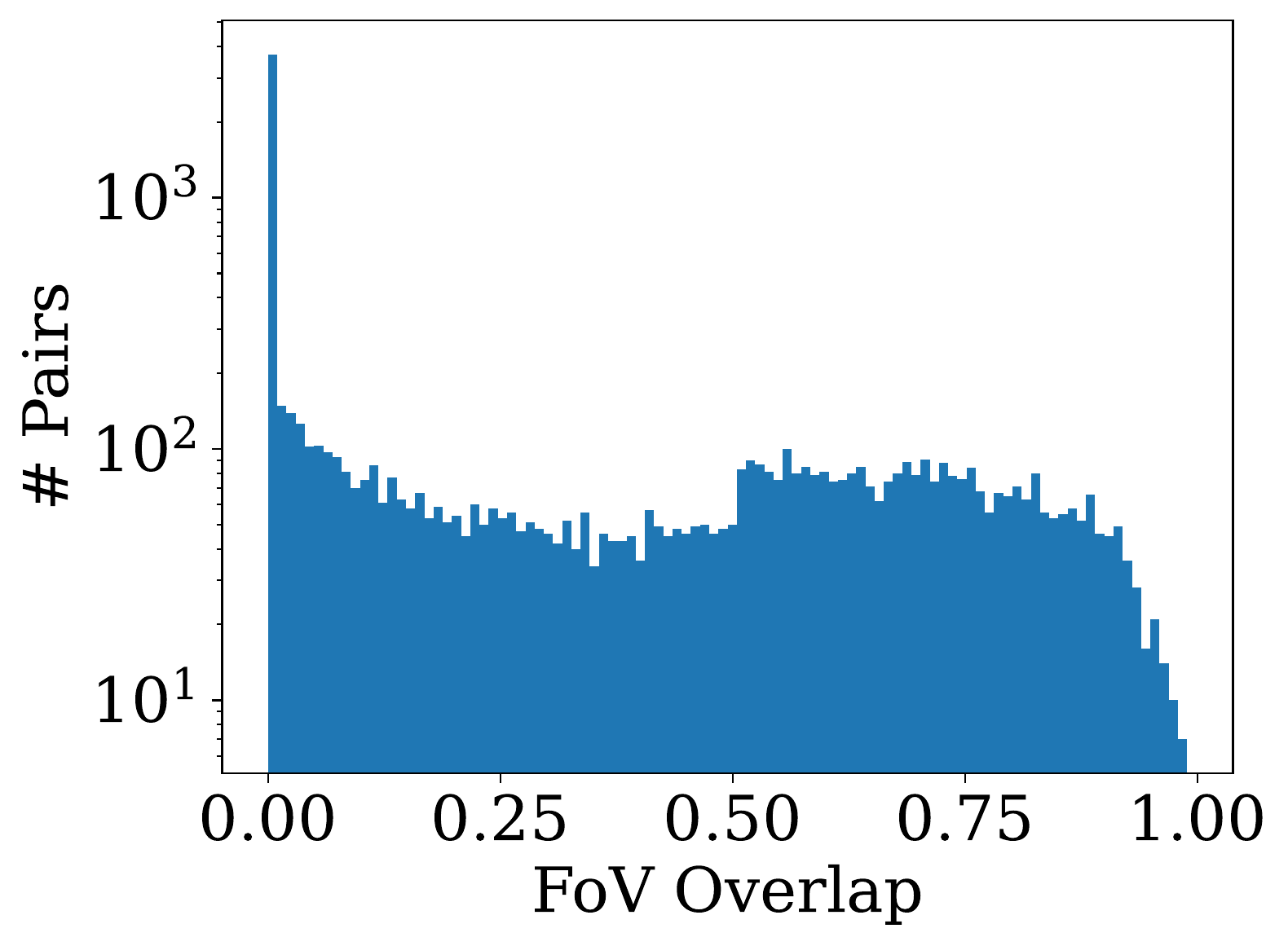}
    \label{fig:pairs_33_33_33}\vspace{-2mm}
    \end{subfigure}\begin{subfigure}{.5\textwidth}
    \includegraphics[width=\textwidth]{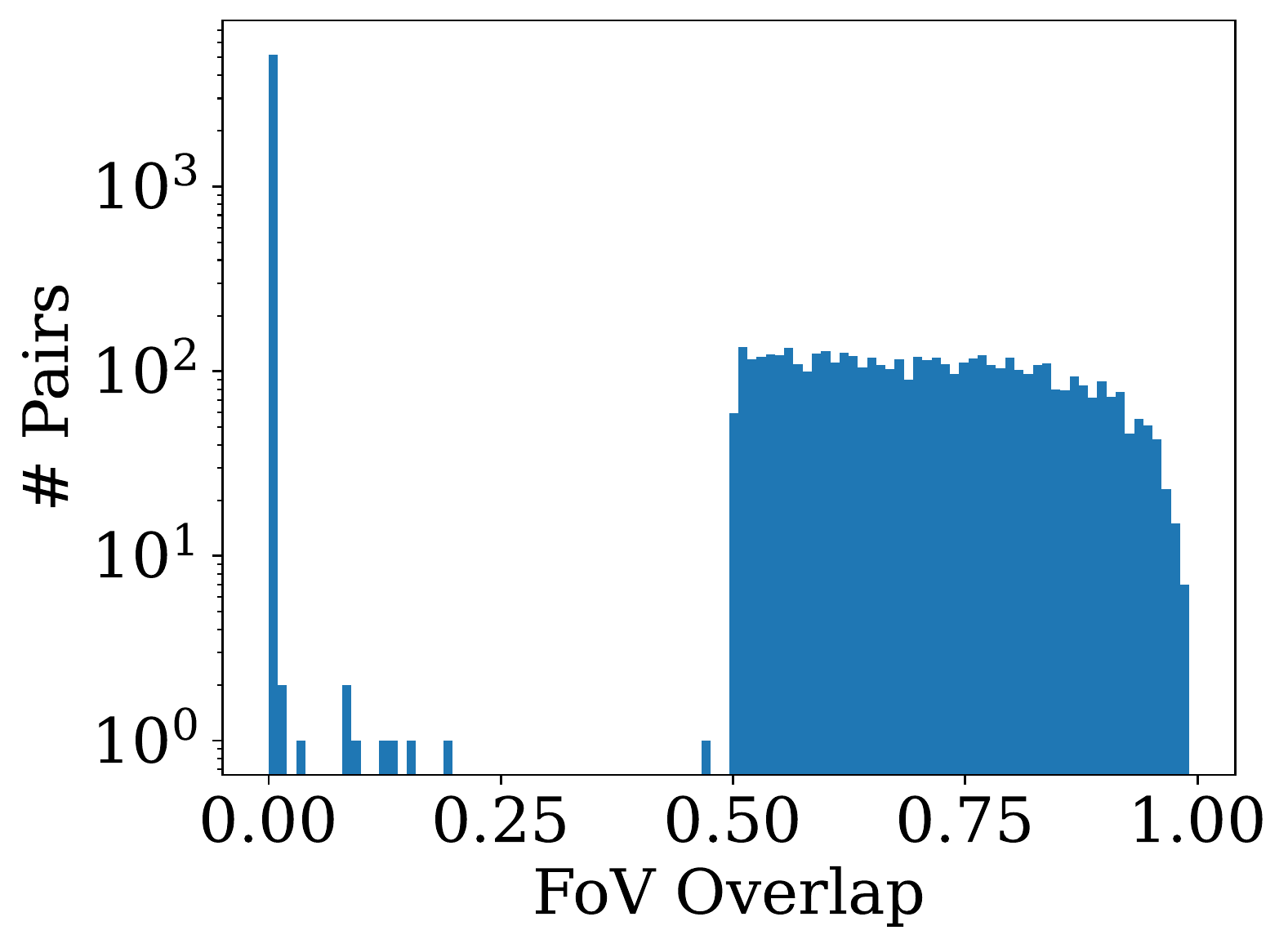}
    \label{fig:pairs_50_50}\vspace{-2mm}
    \end{subfigure}
    \vspace{-4mm}
\caption{Similarity ground truth distribution for 10000 randomly selected pairs in the MSLS train set when using different batch composition strategies. The vertical axis is in log scale.}     \label{fig:batch_composition}
     \end{figure}

Effective composition of training batches is important for the performance of VPR methods, as witnessed by the fact that existing methods deploy very expensive pair mining techniques to select samples that effectively contributes to training. Although we do not use pair mining, we studied the impact on the retrieval performance  by constructing batches with image pairs that have different degrees of similarity. We consider four strategies. Strategy A includes 50\% positive ($\psi \in [0.5, 1]$), 25\% soft-negative ($\psi \in (0, 0.5)$) and 25\% hard-negative ($\psi=0$) pairs in each batch. For Strategy B, the batches contain an uniform distribution of pairs with different similarity, namely 25\% of pairs with $\psi \in [0.75, 1]$, 25\% with $\psi \in [0.5, 0.75)$, 25\% with $\psi \in (0, 0.5)$ and 25\% with $\psi=0$.  For strategy C, we compose the batches with  33.3\% of pairs with $\psi \in [0.5, 1]$, 33.3\% with $\psi \in (0, 0.5)$ and 33.3\% with $\psi=0$. Finally, for strategy D we include 50\% of pairs with $\psi \in [0.5, 1]$ and 50\% with $\psi \in [0, 0.5)$. Histograms of the distributions of similarity in the batches are shown in Figure~\ref{fig:batch_composition}. The most important aspect of composing the training batches is to select an adequate number of soft negative pairs (see Table~\ref{tab:batch}), i.e. at least 25\% of the pairs have similarity $\psi \in (0, 0.5)$, as done for strategy A, B and C. For the main experiments, we used strategy A.

\section{Results on the TB-Places dataset}

\paragraph{The dataset} TB-Places was designed for place recognition in garden~\cite{leyvavallina2019access,leyvavallina2019caip}. It contains images taken by a rover robot in a garden at the University of Wageningen, the Netherlands, over three years. The dataset was collected for the TrimBot2020 project~\cite{strisciuglio2018trimbot2020}. It includes drastic viewpoint variations, as well as illumination changes. The garden is a very challenging small environment with repetitive textures.

\begin{figure}
    \centering
    \begin{subfigure}{.48\columnwidth}
    \includegraphics[width=\columnwidth]{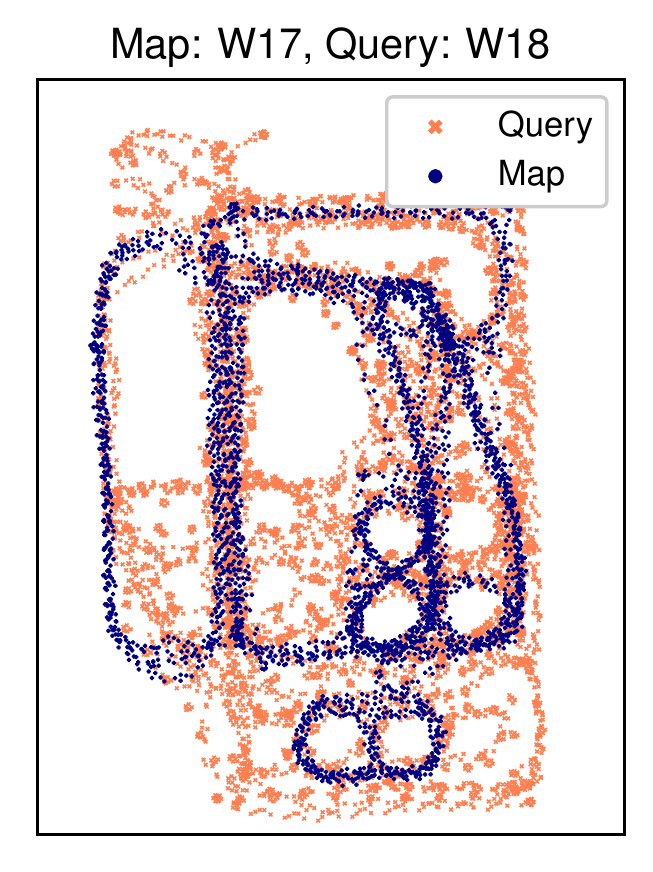}
    \label{fig:w17_18}
    \end{subfigure}
        \begin{subfigure}{.48\columnwidth}
\includegraphics[width=\columnwidth]{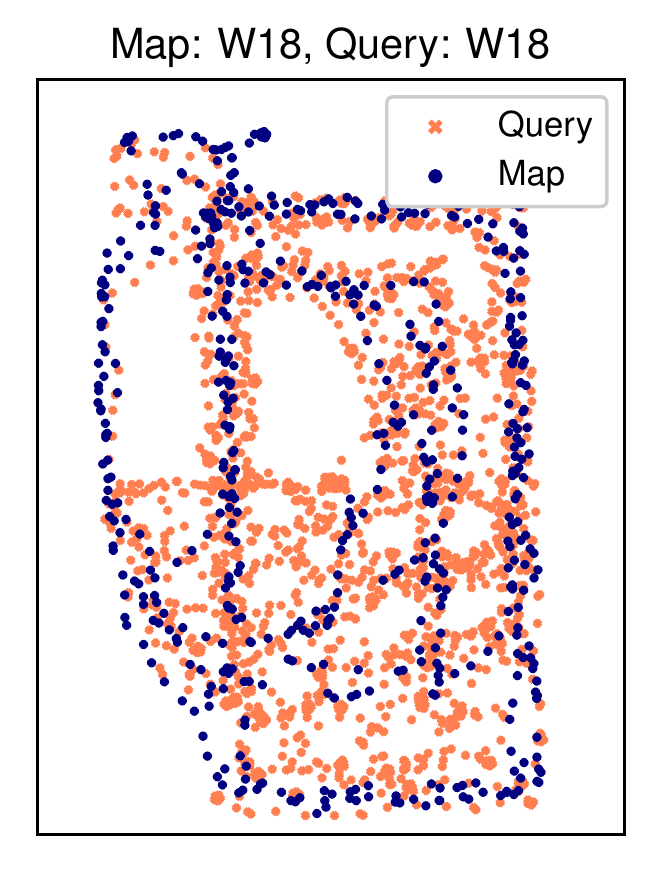}
    \label{fig:w18_w18}
    \end{subfigure}
    \caption{Configurations of the experiments on the TB-Places dataset. (a) W17 subset is the map set, and W18 is the query. (b) We divide W18 into map and query. For visualization purposes, the trajectories have been downsampled.}
    \label{fig:tb_places_test_sets}
\end{figure}

The dataset consists of three subsets, i.e. W16, with 41k images taken in 2016; W17, with 11k images taken in 2017; and W18, with 23k images taken in 2018. As in~\cite{leyvavallina2019caip} we use the W17 subset to train the models.
We design two experiments. For the first one we use W17 as map  (11k images) and W18 as query (23k images). With this configuration we test the robustness of our models w.r.t. changes between the map and the query sets. For the second experiment, we divide W18 into query (17k images) and map (6k images) to test the generalization capabilities of our models in the case both map and query sets were not used for training. In Fig.~\ref{fig:tb_places_test_sets}, we show a sketch of the trajectory that the robot covered for the recording of the reference map (blue trajectory) and query (orange trajectory) images. The query images were taken from locations not covered by map images, thus including substantial viewpoint variations.

\begin{table}[!t]
		\centering
\caption{Results (R@5) on the MSLS validation and test sets, using different training batch composition strategies. }

\label{tab:batch}
\footnotesize
\begin{tabular}{@{}lccccc@{}}
\toprule
 \multirow{2}{*}{\bfseries Model} & \multirow{2}{*}{\bfseries Set} & \multicolumn{4}{c}{\bfseries Strategy} \\ 
 ~ & ~ & \bfseries A & \bfseries B & \bfseries C & \bfseries D \\ \midrule
\multirow{2}{*}{VGG-GeM-GCL} & Validation & 77.8 &78.8 & 78.2 & 73.4  \\
 & Test  & 55.7 & 54.9  & 56 & 49.3\\
  \midrule
\multirow{2}{*}{ResNet50-GeM-GCL} & Validation & 78.9 & 76.5 & 75.7 & 75.4 \\
 & Test & 59.1 & 54.5 & 52.4 & 54.7\\\midrule
\multirow{2}{*}{ResNet152-GeM-GCL} & Validation & 82 & 80.4  & 78.6 & 78.5\\
 & Test & 62.3 & 59.7 & 58.4 &  64.8\\ \midrule
\multirow{2}{*}{ResNeXt-GeM-GCL} & Validation & 86.1 & 86.2 & 87.3  & 81.9 \\
 & Test & 70.8 & 70.7 & 72.4 &  64.8\\ \bottomrule
\end{tabular}%

	
\end{table}


\begin{figure}[!t]
    \centering
    \begin{subfigure}{\columnwidth}
    \begin{minipage}[t]{\columnwidth}

\hspace{0.5cm}
\begin{tikzpicture}[thick, scale=1, every node/.style={scale=.8}] 
    \begin{axis}[%
    hide axis,
    xmin=10,
    xmax=50,
    ymin=0,
    ymax=0.4, height=3.5cm, width=\columnwidth,
    legend style={draw=white!15!black,legend columns=2,legend style={minimum width=.41\columnwidth}},
    legend cell align={left}
    ]
    
\addlegendimage{blue, thick, mark size=2pt, mark=square};
\addlegendentry{ResNet18-CL};

\addlegendimage{red, thick, mark size=2pt, mark=square};
\addlegendentry{ResNet18-GCL (ours)};
\addlegendimage{blue, thick, mark size=2pt, mark=o};
\addlegendentry{ResNet34-CL};

\addlegendimage{red, thick, mark size=2pt, mark=o};
\addlegendentry{ResNet34-GCL (ours)};

\addlegendimage{blue, thick, mark size=2pt, mark=diamond};
\addlegendentry{DenseNet161-CL};

\addlegendimage{red, thick, mark size=2pt, mark=diamond};
\addlegendentry{DenseNet161-GCL (ours)};
\addlegendimage{black, thick, mark size=2pt, mark=triangle};
\addlegendentry{NetVLAD off-the-shelf};

\end{axis}
\end{tikzpicture}
\end{minipage}

    \end{subfigure}
  
    \setcounter{subfigure}{0}
\begin{subfigure}{\columnwidth}
\begin{tikzpicture}[thick, scale=1, every node/.style={scale=.75}]
\begin{axis}
[xlabel=K,
ylabel=Recall@K (\%), ylabel style={at={(axis description cs:0.05,0.5)}},grid, height=6cm, width=\columnwidth]
\addplot[black, thick, mark size=2pt, mark=triangle] table [x=k, y=r, col sep=comma] {plot_data/TB_Places/TB_Places_W18_W17_NetVLAD_recalls.csv};
\addplot[red, thick, mark size=2pt, mark=square] table [x=k, y=r, col sep=comma] {plot_data/TB_Places/TB_Places_ResNet18_avg_2last_soft_siamese_w18_w17_recall_toplot.txt};
\addplot[blue, thick, mark size=2pt, mark=square] table [x=k, y=r, col sep=comma] {plot_data/TB_Places/TB_Places_ResNet18_avg_2last_binary_siamese_w18_w17_recall_toplot.txt};

\addplot[red, thick, mark size=2pt, mark=o] table [x=k, y=r, col sep=comma] {plot_data/TB_Places/TB_Places_ResNet34_avg_2last_soft_siamese_w18_w17_recall_toplot.txt};
\addplot[blue, thick, mark size=2pt, mark=o] table [x=k, y=r, col sep=comma] {plot_data/TB_Places/TB_Places_ResNet34_avg_2last_binary_siamese_w18_w17_recall_toplot.txt};

\addplot[blue, thick, mark size=2pt, mark=diamond] table [x=k, y=r, col sep=comma] {plot_data/TB_Places/TB_Places_W18_W17_CAIP_recalls.csv};
\addplot[red, thick, mark size=2pt, mark=diamond] table [x=k, y=r, col sep=comma] {plot_data/TB_Places/TB_Places_W18_W17_Soft_recalls.csv};
\end{axis}
\end{tikzpicture}
\caption{Map: W17, Query: W18}
    \label{fig:w17_w18_recall}
\end{subfigure}
\begin{subfigure}{\columnwidth}
\begin{tikzpicture}[thick, scale=1, every node/.style={scale=.75}]
\begin{axis}
[xlabel=K,
ylabel=Recall@K (\%), ylabel style={at={(axis description cs:0.05,0.5)}},
legend pos =south east,grid, height=6cm, width=\columnwidth]
\addplot[black, thick, mark size=2pt, mark=triangle] table [x=k, y=r, col sep=comma] {plot_data/TB_Places/TB_Places_map_query_NetVLAD_recalls.csv};
\addplot[red, thick, mark size=2pt, mark=square] table [x=k, y=r, col sep=comma] {plot_data/TB_Places/TB_Places_ResNet18_avg_2last_soft_siamese_w18_map_query_recall_toplot.txt};
\addplot[blue, thick, mark size=2pt, mark=square] table [x=k, y=r, col sep=comma] {plot_data/TB_Places/TB_Places_ResNet18_avg_2last_binary_siamese_w18_map_query_recall_toplot.txt};

\addplot[red, thick, mark size=2pt, mark=o] table [x=k, y=r, col sep=comma] {plot_data/TB_Places/TB_Places_ResNet34_avg_2last_soft_siamese_w18_map_query_recall_toplot.txt};
\addplot[blue, thick, mark size=2pt, mark=o] table [x=k, y=r, col sep=comma] {plot_data/TB_Places/TB_Places_ResNet34_avg_2last_binary_siamese_w18_map_query_recall_toplot.txt};

\addplot[blue, thick, mark size=2pt, mark=diamond] table [x=k, y=r, col sep=comma] {plot_data/TB_Places/TB_Places_map_query_CAIP_recalls.csv};
\addplot[red, thick, mark size=2pt, mark=diamond] table [x=k, y=r, col sep=comma] {plot_data/TB_Places/TB_Places_map_query_Soft_recalls.csv};
\end{axis}
\end{tikzpicture}
\caption{Map: W18, Query: W18}
    \label{fig:w18_map_query_recall}
\end{subfigure}
   \caption{Results on the TB-Places dataset. In (a) the results when using W17 as map and W18 as query set. In (b) the top-k recall achieved when dividing W18 into map and query. }
       \label{fig:tb_places_recall}
\end{figure}
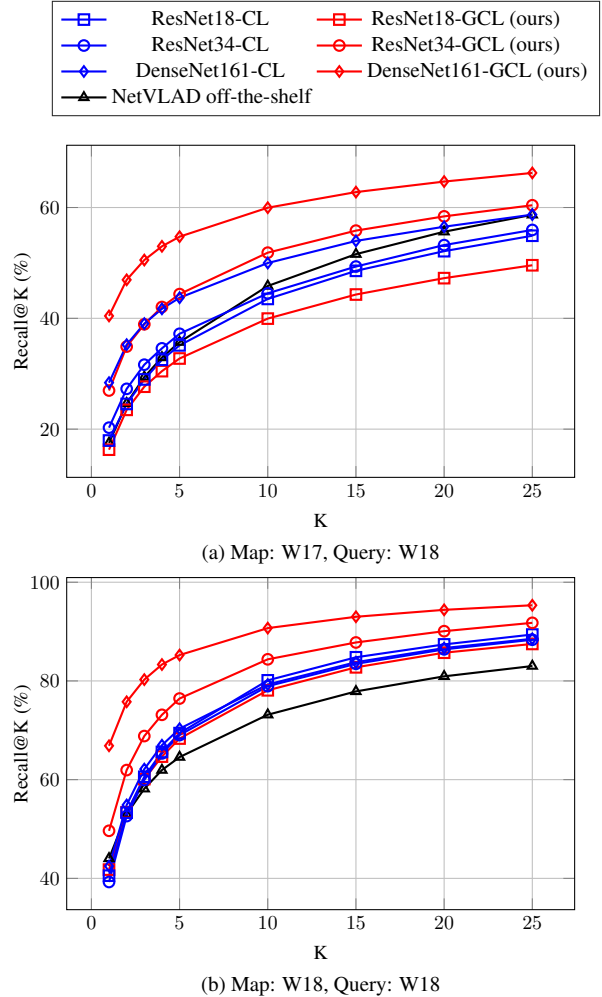

\paragraph{Experiments and results}
We trained our method on the TB-Places dataset and show the results in Fig.~\ref{fig:tb_places_recall}. We considered three  backbones, namely ResNet18, ResNet34, and DenseNet161, which learn descriptors of size 512, 512 and 2208, respectively. We trained them with the binary Contrastive Loss function and with our Generalized Contrastive Loss function. Furthermore, we compare them with the NetVLAD off-the-shelf model. For these experiments we use only a Global Average Pooling.

\begin{figure*}[!t]
\begin{subfigure}{.24\columnwidth}
\begin{tikzpicture}[thick, scale=1, every node/.style={scale=.8}]
\begin{axis}
[xlabel=K,
ylabel=Recall@K (\%), ylabel style={at={(axis description cs:0.1,0.5)}},grid, height=4.75cm, width=\columnwidth,xtick distance=5, ytick distance=5]
\addplot[black, thick, mark size=2pt, mark=triangle] table [x=k, y=r, col sep=comma] {plot_data/7Scenes/PR_7scenes_recallatk_50FOV_heads_NetVLAD_recalls.csv};
\addplot[blue, thick, mark size=2pt, mark=o] table [x=k, y=r, col sep=comma] {plot_data/7Scenes/PR_7scenes_recallatk_50FOV_heads_ResNet18_crisp_recalls.csv};
\addplot[red, thick, mark size=2pt, mark=o] table [x=k, y=r, col sep=comma] {plot_data/7Scenes/PR_7scenes_recallatk_50FOV_heads_ResNet18_soft_recalls.csv};

\addplot[blue, thick, mark size=2pt, mark=square] table [x=k, y=r, col sep=comma] {plot_data/7Scenes/PR_7scenes_recallatk_50FOV_heads_ResNet34_crisp_recalls.csv};
\addplot[red, thick, mark size=2pt, mark=square] table [x=k, y=r, col sep=comma] {plot_data/7Scenes/PR_7scenes_recallatk_50FOV_heads_ResNet34_soft_recalls.csv};
\end{axis}
\end{tikzpicture}
    \caption{Heads}
    \label{fig:7Scenes_heads_recall}
\end{subfigure}~
\begin{subfigure}{.24\columnwidth}
\begin{tikzpicture}[thick, scale=1, every node/.style={scale=.8}]
\begin{axis}
[xlabel=K,
ylabel=Recall@K (\%), ylabel style={at={(axis description cs:0.1,0.5)}},grid, height=4.75cm, width=\columnwidth,xtick distance=5, ytick distance=5]
\addplot[black, thick, mark size=2pt, mark=triangle] table [x=k, y=r, col sep=comma] {plot_data/7Scenes/PR_7scenes_recallatk_50FOV_stairs_NetVLAD_recalls.csv};
\addplot[blue, thick, mark size=2pt, mark=o] table [x=k, y=r, col sep=comma] {plot_data/7Scenes/PR_7scenes_recallatk_50FOV_stairs_ResNet18_crisp_recalls.csv};
\addplot[red, thick, mark size=2pt, mark=o] table [x=k, y=r, col sep=comma] {plot_data/7Scenes/PR_7scenes_recallatk_50FOV_stairs_ResNet18_soft_recalls.csv};

\addplot[blue, thick, mark size=2pt, mark=square] table [x=k, y=r, col sep=comma] {plot_data/7Scenes/PR_7scenes_recallatk_50FOV_stairs_ResNet34_crisp_recalls.csv};
\addplot[red, thick, mark size=2pt, mark=square] table [x=k, y=r, col sep=comma] {plot_data/7Scenes/PR_7scenes_recallatk_50FOV_stairs_ResNet34_soft_recalls.csv};
\end{axis}
\end{tikzpicture}
    \caption{Stairs}
    \label{fig:7Scenes_stairs_recall}
\end{subfigure}~
\begin{subfigure}{.24\columnwidth}
\begin{tikzpicture}[thick, scale=1, every node/.style={scale=.8}]
\begin{axis}
[xlabel=K,
ylabel=Recall@K (\%), ylabel style={at={(axis description cs:0.1,0.5)}},grid, height=4.75cm, width=\columnwidth,xtick distance=5, ytick distance=5]
\addplot[black, thick, mark size=2pt, mark=triangle] table [x=k, y=r, col sep=comma] {plot_data/7Scenes/PR_7scenes_recallatk_50FOV_pumpkin_NetVLAD_recalls.csv};
\addplot[blue, thick, mark size=2pt, mark=o] table [x=k, y=r, col sep=comma] {plot_data/7Scenes/PR_7scenes_recallatk_50FOV_pumpkin_ResNet18_crisp_recalls.csv};
\addplot[red, thick, mark size=2pt, mark=o] table [x=k, y=r, col sep=comma] {plot_data/7Scenes/PR_7scenes_recallatk_50FOV_pumpkin_ResNet18_soft_recalls.csv};

\addplot[blue, thick, mark size=2pt, mark=square] table [x=k, y=r, col sep=comma] {plot_data/7Scenes/PR_7scenes_recallatk_50FOV_pumpkin_ResNet34_crisp_recalls.csv};
\addplot[red, thick, mark size=2pt, mark=square] table [x=k, y=r, col sep=comma] {plot_data/7Scenes/PR_7scenes_recallatk_50FOV_pumpkin_ResNet34_soft_recalls.csv};
\end{axis}
\end{tikzpicture}
    \caption{Pumpkin}
    \label{fig:7Scenes_pumpkin_recall}
\end{subfigure}~
\begin{subfigure}{.4\columnwidth}
\begin{minipage}[c][2cm]{\columnwidth}
\vspace{10pt}
\begin{tikzpicture}[thick, scale=1, every node/.style={scale=.8}] 
    \begin{axis}[%
    hide axis,
    xmin=10,
    xmax=70,
    ymin=0,
    ymax=0.4,
    legend pos =north west,
    legend style={draw=white!20!black,legend cell align=left}, height=4.75cm, width=.61\columnwidth
    ]
    \addlegendimage{black, thick, mark size=2pt, mark=triangle}
    \addlegendentry{NetVLAD off-the-shelf};
    \addlegendimage{blue, thick, mark size=2pt, mark=o}
    \addlegendentry{ResNet18-CL};
    \addlegendimage{red, thick, mark size=2pt, mark=o}
    \addlegendentry{ResNet18-GCL (ours)};
    \addlegendimage{blue, thick, mark size=2pt, mark=square}
    \addlegendentry{ResNet34-CL};
    \addlegendimage{red, thick, mark size=2pt, mark=square}
    \addlegendentry{ResNet34-GCL (ours)};
    \end{axis}
    \end{tikzpicture} 
    \end{minipage}

\end{subfigure}

\setcounter{subfigure}{3}

\begin{subfigure}{.24\columnwidth}
\begin{tikzpicture}[thick, scale=1, every node/.style={scale=.8}]
\begin{axis}
[xlabel=K,
ylabel=Recall@K (\%), ylabel style={at={(axis description cs:0.1,0.5)}},grid, height=4.75cm, width=\columnwidth,xtick distance=5, ytick distance=5]
\addplot[black, thick, mark size=2pt, mark=triangle] table [x=k, y=r, col sep=comma] {plot_data/7Scenes/PR_7scenes_recallatk_50FOV_fire_NetVLAD_recalls.csv};
\addplot[blue, thick, mark size=2pt, mark=o] table [x=k, y=r, col sep=comma] {plot_data/7Scenes/PR_7scenes_recallatk_50FOV_fire_ResNet18_crisp_recalls.csv};
\addplot[red, thick, mark size=2pt, mark=o] table [x=k, y=r, col sep=comma] {plot_data/7Scenes/PR_7scenes_recallatk_50FOV_fire_ResNet18_soft_recalls.csv};

\addplot[blue, thick, mark size=2pt, mark=square] table [x=k, y=r, col sep=comma] {plot_data/7Scenes/PR_7scenes_recallatk_50FOV_fire_ResNet34_crisp_recalls.csv};
\addplot[red, thick, mark size=2pt, mark=square] table [x=k, y=r, col sep=comma] {plot_data/7Scenes/PR_7scenes_recallatk_50FOV_fire_ResNet34_soft_recalls.csv};
\end{axis}
\end{tikzpicture}
    \label{fig:7Scenes_fire_recall}
\caption{Fire}
\end{subfigure}~
\begin{subfigure}{.24\columnwidth}
\begin{tikzpicture}[thick, scale=1, every node/.style={scale=.8}]
\begin{axis}
[xlabel=K,
ylabel=Recall@K (\%), ylabel style={at={(axis description cs:0.1,0.5)}},grid, height=4.75cm, width=\columnwidth,xtick distance=5, ytick distance=5]
\addplot[black, thick, mark size=2pt, mark=triangle] table [x=k, y=r, col sep=comma] {plot_data/7Scenes/PR_7scenes_recallatk_50FOV_redkitchen_NetVLAD_recalls.csv};
\addplot[blue, thick, mark size=2pt, mark=o] table [x=k, y=r, col sep=comma] {plot_data/7Scenes/PR_7scenes_recallatk_50FOV_redkitchen_ResNet18_crisp_recalls.csv};
\addplot[red, thick, mark size=2pt, mark=o] table [x=k, y=r, col sep=comma] {plot_data/7Scenes/PR_7scenes_recallatk_50FOV_redkitchen_ResNet18_soft_recalls.csv};

\addplot[blue, thick, mark size=2pt, mark=square] table [x=k, y=r, col sep=comma] {plot_data/7Scenes/PR_7scenes_recallatk_50FOV_redkitchen_ResNet34_crisp_recalls.csv};
\addplot[red, thick, mark size=2pt, mark=square] table [x=k, y=r, col sep=comma] {plot_data/7Scenes/PR_7scenes_recallatk_50FOV_redkitchen_ResNet34_soft_recalls.csv};
\end{axis}
\end{tikzpicture}
    \caption{Redkitchen}
    \label{fig:7Scenes_redkitchen_recall}
\end{subfigure}~
\begin{subfigure}{.24\columnwidth}
\begin{tikzpicture}[thick, scale=1, every node/.style={scale=.8}]
\begin{axis}
[xlabel=K,
ylabel=Recall@K (\%), ylabel style={at={(axis description cs:0.1,0.5)}},grid, height=4.75cm, width=\columnwidth,xtick distance=5, ytick distance=5]
\addplot[black, thick, mark size=2pt, mark=triangle] table [x=k, y=r, col sep=comma] {plot_data/7Scenes/PR_7scenes_recallatk_50FOV_chess_NetVLAD_recalls.csv};
\addplot[blue, thick, mark size=2pt, mark=o] table [x=k, y=r, col sep=comma] {plot_data/7Scenes/PR_7scenes_recallatk_50FOV_chess_ResNet18_crisp_recalls.csv};
\addplot[red, thick, mark size=2pt, mark=o] table [x=k, y=r, col sep=comma] {plot_data/7Scenes/PR_7scenes_recallatk_50FOV_chess_ResNet18_soft_recalls.csv};

\addplot[blue, thick, mark size=2pt, mark=square] table [x=k, y=r, col sep=comma] {plot_data/7Scenes/PR_7scenes_recallatk_50FOV_chess_ResNet34_crisp_recalls.csv};
\addplot[red, thick, mark size=2pt, mark=square] table [x=k, y=r, col sep=comma] {plot_data/7Scenes/PR_7scenes_recallatk_50FOV_chess_ResNet34_soft_recalls.csv};
\end{axis}
\end{tikzpicture}
\caption{Chess}
    \label{fig:7Scenes_chess_recall}
\end{subfigure}~
\begin{subfigure}{.24\columnwidth}
\begin{tikzpicture}[thick, scale=1, every node/.style={scale=.8}]
\begin{axis}
[xlabel=K,
ylabel=Recall@K (\%), ylabel style={at={(axis description cs:0.1,0.5)}},grid, height=4.75cm, width=\columnwidth,xtick distance=5, ytick distance=5]
\addplot[black, thick, mark size=2pt, mark=triangle] table [x=k, y=r, col sep=comma] {plot_data/7Scenes/PR_7scenes_recallatk_50FOV_office_NetVLAD_recalls.csv};
\addplot[blue, thick, mark size=2pt, mark=o] table [x=k, y=r, col sep=comma] {plot_data/7Scenes/PR_7scenes_recallatk_50FOV_office_ResNet18_crisp_recalls.csv};
\addplot[red, thick, mark size=2pt, mark=o] table [x=k, y=r, col sep=comma] {plot_data/7Scenes/PR_7scenes_recallatk_50FOV_office_ResNet18_soft_recalls.csv};

\addplot[blue, thick, mark size=2pt, mark=square] table [x=k, y=r, col sep=comma] {plot_data/7Scenes/PR_7scenes_recallatk_50FOV_office_ResNet34_crisp_recalls.csv};
\addplot[red, thick, mark size=2pt, mark=square] table [x=k, y=r, col sep=comma] {plot_data/7Scenes/PR_7scenes_recallatk_50FOV_office_ResNet34_soft_recalls.csv};
\end{axis}
\end{tikzpicture}
    \caption{Office}
    \label{fig:7Scenes_office_recall}
\end{subfigure}
\caption{Recall@K results achieved on the 7 Scenes dataset. The results of the models trained with our Generalized Contrastive Loss are shown in red, while those of the models trained with the binary Contrastive Loss are shown in blue. The Recall@K achieved by NetVLAD off-the-shelf is plotted in black.}
\label{fig:recallat_k_7scenes}

\end{figure*}
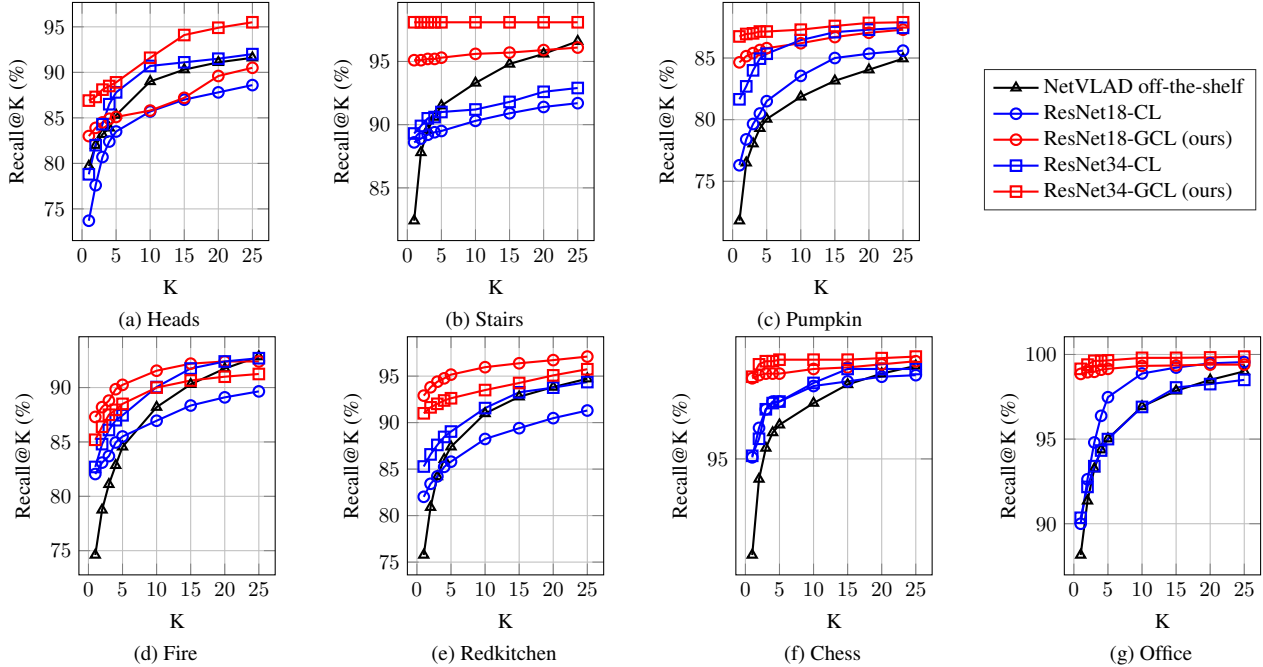

We report results of two experiments. For the first one, we used the training set W17 as map, and the W18 set as query. With this experiment we tested the strength of the descriptors against significant variations between the query and the map set. We show the results in Fig.~\ref{fig:w17_w18_recall}. For the second experiment, we divided the W18 set into map and query, to test the generalization capabilities when both map and query sets are unknown to the place recognition model (i.e. not seen during training). The results are displayed in Fig.~\ref{fig:w18_map_query_recall}.
For both experiments the models trained with GCL and the proposed similarity ground truth consistently achieve better recall than the ones trained with the binary Contrastive Loss, with the exception of the ResNet18. 

\section{Results on the 7Scenes dataset}

\paragraph{The dataset}
A benchmark dataset for indoor camera localization~\cite{Shotton2013}. It includes 26k training images and 17k test images, taken in seven environments. Each image has an associated ground truth 6DOF camera pose. Additionally, a 3D reconstruction of each scene is available. We use it to test our VPR models in indoor environments. For evaluation purposes, we define an image pair as a positive match if their annotated degree of similarity is higher than $50\%$. We use the training set as map, and the test set as query.

\paragraph{Experiments and results}
\label{sec:results7scenes}
We report the results in Fig.~\ref{fig:recallat_k_7scenes}. We used the ResNet18 and ResNet34 architectures as backbones with a Global Average Pooling layer and we compare them to NetVLAD off-the-shelf.  We achieved generally higher Recall@K results for the models trained using the GCL function, for all scenes. The cases of the \emph{stairs}  (Fig.~\ref{fig:7Scenes_stairs_recall}), \emph{chess}  (Fig.~\ref{fig:7Scenes_chess_recall}) and \emph{office} (Fig.~\ref{fig:7Scenes_office_recall}) scenes are particularly interesting, since with the GCL descriptors we are able to retrieve positive matches for nearly all the query images, with a top-5 recall for \mbox{ResNet34-GCL} of 98.1\%, 98.2\%  and 99.7\%, respectively.

Furthermore, we report the average precision results in Table~\ref{tab:ap-7scenes}. 
The models trained with the GCL function achieved higher AP than their corresponding models trained with the binary CL function. We achieved an average AP equal to 0.89 using the ResNet34-GCL model. 



\begin{table}[t!]
\centering
\caption{Average Precision results obtained by the networks trained with the proposed Generalized Contrastive loss function on the 7Scenes  dataset, compared with those achieved by the same network architectures trained using the binary Contrastive loss function and by the NetVLAD off-the-shelf model. }
\label{tab:ap-7scenes}

\renewcommand{\arraystretch}{1.0}
\resizebox{\textwidth}{!}{%
\begin{tabular}{@{\extracolsep{4pt}}lccccc@{}}
\toprule
\textbf{}   & \textbf{NetVLAD}       & \multicolumn{2}{c}{\textbf{ResNet18}} & \multicolumn{2}{c}{\textbf{ResNet34}} \\ 
\cmidrule{2-2}
\cmidrule{3-4}
\cmidrule{5-6}
\textbf{Scene}       & \textbf{off-the-shelf} & \textbf{CL}       & \textbf{GCL}               & \textbf{CL}       & \textbf{GCL}             \\ \midrule 
Heads      & 0.587        & 0.739   & 0.807           & 0.759   & \textbf{0.853}   \\
Stairs     & 0.533        & 0.855   & 0.883            & 0.884   & \textbf{0.944}   \\
Pumpkin    & 0.491        & 0.768   & 0.849            & 0.782   & \textbf{0.914}   \\
Fire       & 0.539        & 0.786   & \textbf{0.811}   & 0.796   & 0.803            \\
Redkitchen & 0.439        & 0.790   & 0.876            & 0.782   & \textbf{0.902}   \\
Chess      & 0.645        & 0.943   & 0.964 & 0.945   & \textbf{0.974}   \\
Office     & 0.399        & 0.794   & 0.890            & 0.802   & \textbf{0.896}   \\ \midrule
\textit{Mean }      & 0.519        & 0.811   & 0.868            & 0.821   & \textbf{0.898}   \\ \bottomrule
\end{tabular}}
\end{table}

\begin{figure*}[!t]
    \centering
    \includegraphics[width=0.9\textwidth]{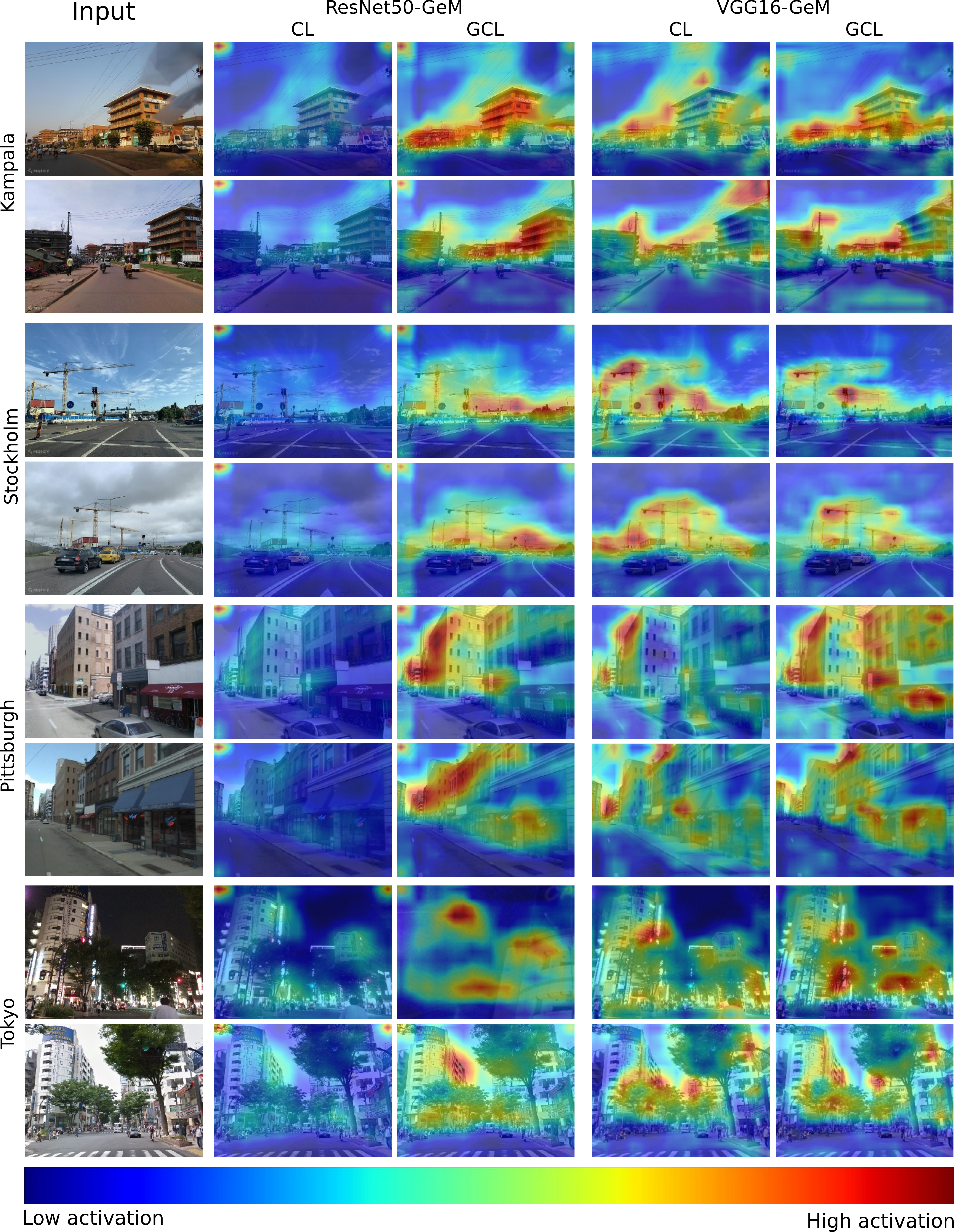}
    \caption{CNN activations for the ResNet50-GeM and the VGG16-GeM models with CL and GCL for several input image pairs. The first two pairs, corresponding to the first four columns, are part of the MSLS test set. The third and fourth belong to the Pittsburgh30k and Tokyo 24/7 test set, respectively. We show the activations for the last layer of the backbone overlapped with the input images.}
    \label{fig:activations}
\end{figure*}

\section{GCL vs CL: Activation maps}
In Fig.~\ref{fig:activations}, we show the activation maps of the last convolutional layer of our models with a VGG16-GeM and a ResNet50-GeM backbone, both trained using the Contrastive Loss function (CL) and the proposed Generalized Contrastive Loss (GCL) function.  We selected two example image pairs from the MSLS test set~\cite{msls} (Kampala and Stockholm), one from the Pittsburgh30k test set~\cite{Arandjelovic2017}, and one from the Tokyo 24/7 dataset~\cite{Torii-CVPR2013}. For all cases, we observed that the model trained with the GCL function produces higher activation for the common visual features of the images, and lower for the irrelevant parts (i.e. the road or the sky), in contrast to the model trained with the binary CL, which focuses less in the concerned areas of the pictures. In the example from Stockholm we can observe that our model does not respond to the cars (which vary from picture to picture), while it does respond strongly to the cranes (which are a longer-term cue). The example from Tokyo 24/7 is also particularly interesting: our model trained with the GCL function has high responses on the common parts of the images even under big changes of illumination. We observed that ResNet architectures tend to produce a peak of activation on the top left corner of the images. This does not seem to occur with the VGG architecture, so our intuition is that this artifact is due to the residual computation.

\section{Gradient of the Generalized Contrastive Loss}
Let us consider two input images $x_i$ and $x_j$, their latent representations $\hat{f}(x_i)$ and $\hat{f}(x_j)$, and define $ d(x_i, x_j)$ the Euclidean distance between the representations, such as:
\begin{equation}
\nonumber
 d(x_i, x_j) = \left \| \hat{f}(x_i)-\hat{f}(x_j) \right \|_2
\end{equation}
For simplicity of notation, hereinafter we refer to $d(x_i, x_j)$ as $d$.

\subsection{Contrastive loss}

\noindent The Contrastive Loss function is defined as:
\begin{equation}
\nonumber
\mathcal{L}_{CL}=\begin{cases}
\frac{1}{2} d ^2 ,& \text{if } y= 1\\
\frac{1}{2}\max(\tau- d ,0)^2,& \text{if } y=0
\end{cases}
\end{equation}
\noindent where $y$ corresponds to the binary ground truth label and $\tau$ corresponds to the margin.

\noindent In order to compute the gradient for this function, we consider three cases, depending on the ground truth label $y$ and the value of the distance $d$. 

\noindent \textbf{Case 1) $y=1$}

\noindent The loss function becomes:
\begin{equation}
    \nonumber
    \mathcal{L}_{CL}=\frac{1}{2} d ^2
\end{equation}

\noindent and its derivative with respect to $d$ is:
\begin{align}
\nonumber
\begin{split}
    \nabla  \mathcal{L}_{CL} = \frac{\partial}{\partial d} \left( \mathcal{L}_{CL} \right) =\frac{\partial}{\partial d} \left( \frac{1}{2} d^2 \right)=  d
\end{split}
\end{align}

\noindent \textbf{Case 2) $y=0, d < \tau$}
\noindent The loss function becomes:
\begin{equation}
    \nonumber
    \mathcal{L}_{CL}=\frac{1}{2}(\tau- d)^2
\end{equation}

\noindent and its derivative with respect to $d$ is:
\begin{align*}
\nonumber
\begin{split}
\nabla {} \mathcal{L}_{CL}=&
\frac{\partial}{\partial d} \left(  \mathcal{L}_{CL} \right) =\frac{\partial}{\partial d} \left[ \frac{1}{2}(\tau- d)^2 \right]=(\tau-d)(-1)= d-\tau
\end{split}
\end{align*}

\noindent  \textbf{Case 3) $y=0, d \geq \tau$}
\begin{equation}
    \nonumber
    \mathcal{L}_{CL} = 0
\end{equation}
and the gradient $\mathcal{L}_{CL} = 0$ as well. Thus, case 2 and case 3, for $y=0$, can be grouped as: 

\begin{equation}
    \nonumber
     \nabla  \mathcal{L}_{CL}=\begin{cases}
     d-\tau ,& \text{if } d < \tau\\
    0 ,& \text{if } d \geq \tau
\end{cases}
\end{equation}

\noindent and simplified as:

\begin{equation}
    \nonumber
     \nabla  \mathcal{L}_{CL}=\min( d-\tau,0)
\end{equation}

\noindent Finally, the \textbf{gradient of the Contrastive Loss function} is:

\begin{equation}
\nonumber
     \nabla  \mathcal{L}_{CL}=\begin{cases}
     d ,& \text{if } y= 1\\
    \min( d-\tau,0) ,& \text{if } y=0
\end{cases}
\end{equation}

\subsection{Generalized Contrastive loss}
We defined the Generalized Contrastive Loss function as:
\begin{equation}
\nonumber
 \mathcal{L}_{GCL}= \psi_{i,j}\cdot \frac{1}{2}d^2 + (1-\psi_{i,j}) \cdot \frac{1}{2}\max(\tau-d,0)^2
\end{equation}
\noindent where $\psi_{i,j}$ is the ground truth degree of similarity between the input images $x_i$ and $x_j$, and its values are in the interval $[0,1]$.
To compute the gradient of the GCL function we consider two cases, namely when 1) the distance $d$ between the representations is lower than the margin $\tau$ and 2) the alternative case when $d$ is larger than $\tau$. 

\noindent \textbf{Case 1) $d < \tau$}
\noindent The Generalized Contrastive loss function becomes:

\begin{equation}
\nonumber
 \mathcal{L}_{GCL}= \psi_{i,j}\cdot \frac{1}{2}d^2 + (1-\psi_{i,j}) \cdot \frac{1}{2}(\tau-d)^2 
\end{equation}

\noindent and its derivative with respect to $d$ is:
\begin{align*} 
\nonumber
\begin{split}
\nabla  \mathcal{L}_{GCL}=\frac{\partial}{\partial d} \left[ \psi_{i,j} \cdot \frac{1}{2} d^2  + (1 - \psi_{i,j}) \cdot \frac{1}{2}(\tau - d)^2 \right]=\\
=\psi_{i,j} \cdot d + (1-\psi_{i,j}) (\tau-d)(-1)=
\\=\psi_{i,j} \cdot d + d - \tau- \psi_{i,j} \cdot d + \psi_{i,j} \cdot \tau =
\\=d+\tau(\psi_{i,j} -1)
\end{split}
\end{align*}

\noindent \textbf{Case 2) $d \geq \tau$}

\noindent The Generalized Contrastive loss function becomes:
\begin{equation}
\nonumber
 \mathcal{L}_{GCL}= \psi_{i,j}\cdot \frac{1}{2}d^2 + (1-\psi_{i,j}) \cdot \frac{1}{2}(0)^2 = \psi_{i,j}\cdot \frac{1}{2}d^2
\end{equation}

\noindent and its derivative with respect to $d$ is:
\begin{equation}
    \nonumber
\nabla \mathcal{L}_{GCL}=
\frac{\partial}{\partial d}\mathcal{L}_{GCL}=
\frac{\partial}{\partial d} \left[ \psi_{i,j}\cdot \frac{1}{2} d^2  \right] = 
d \cdot \psi_{i,j}
\end{equation}

\noindent Finally, the \textbf{gradient of the Generalized Contrastive Loss function} is:
\begin{equation}
     \nabla  \mathcal{L}_{GCL}=\begin{cases}d+\tau(\psi_{i,j}-1),& \text{if } d<\tau\\
     d \cdot \psi_{i,j},& \text{if } d \geq \tau\\
     \end{cases}
\end{equation}


\end{document}